\documentclass{article}[twocolumn]
\usepackage[T1]{fontenc}
\usepackage[utf8]{inputenc}
\usepackage{graphicx}
\usepackage{amsmath}
\usepackage{multicol}
\usepackage{csquotes}
\usepackage{booktabs}
\usepackage{xcolor}
\usepackage{hyperref}
\usepackage{subcaption}
\usepackage{amsfonts}
\usepackage{algpseudocode}
\usepackage{algorithm}
\usepackage{amsthm}
\usepackage{tikz}
\usepackage[inline]{enumitem}
\usepackage{natbib}
\usepackage{arxiv}

\setcitestyle{authoryear,round,citesep={;},aysep={,},yysep={;}}

\theoremstyle{definition}

\newtheorem{property}{Property}

\algdef{SE}[DOWHILE]{Do}{doWhile}{\algorithmicdo}[1]{\algorithmicwhile\ #1}%

\title{Reducing Aleatoric and Epistemic Uncertainty through Multi-modal Data Acquisition}

\author{
 Arthur Hoarau \\
  Université de Lorraine, CentraleSupélec\\
  CNRS, LORIA, Metz, France \\
  \texttt{arthur.hoarau@centralesupelec.fr} \\
   \And
 Benjamin Quost \\
  Universit\'e de technologie de Compi\`egne\\
  UMR CNRS 7253 Heudiasyc, France \\
  \texttt{benjamin.quost@utc.fr} \\
  \And
 S\'ebastien Destercke \\
  CNRS, Universit\'e de technologie de Compi\`egne\\
  UMR CNRS 7253 Heudiasyc, France \\
  \texttt{sebastien.destercke@utc.fr} \\
  \And
 Willem Waegeman \\
  University of Ghent\\
  Ghent, Belgium \\
  \texttt{willem.waegeman@UGent.be} \\
}

\begin{document}

\maketitle


\begin{abstract}
To generate accurate and reliable predictions, modern AI systems need to combine data from multiple modalities, such as text, images, audio, spreadsheets, and time series. However, collecting training and test data for many modalities is challenging and time-consuming, creating a need for cost-efficient multi-modal data acquisition. In this paper we advocate that this can be realized by disentangling epistemic and aleatoric uncertainty. It is commonly assumed in the machine learning community that epistemic uncertainty can be reduced by collecting more data, while aleatoric uncertainty is irreducible. We claim that this assumption can be challenged in modern multi-modal AI systems, and we introduces an innovative data acquisition framework where uncertainty disentanglement leads to actionable decisions, allowing cost-efficient sampling in two directions: sample size and data modality. The main hypothesis is that aleatoric uncertainty decreases as the number of modalities increases, while epistemic uncertainty decreases by collecting more observations. We provide a theoretical analysis and proof-of-concept implementations on various multi-modal datasets to prove the usefulness of our framework, which combines ideas from active learning, active feature acquisition and uncertainty quantification.
\end{abstract}


\section{Introduction}\label{section:intro}

Machine learning models are increasingly used to make decisions affecting people's lives, such as in medical diagnosis, law enforcement, police investigation, insurance and loan approval, or environmental permit deliverance. Because incorrect decisions can have significant consequences, these systems must not only be accurate, but also provide a credible measure of their own uncertainty, warranting whenever model outputs should be trusted. Therefore, the problem of uncertainty quantification is currently the focus of many research works, a number of which are dedicated to distinguish between two specific types of uncertainty, known as aleatoric and epistemic~\citep{Hora1996,kendall2017,eyke2019,Bengs2023,Lakshmi2024}. 

Aleatoric uncertainty is due to inherent randomness, and is generally assumed to be irreducible. Epistemic uncertainty arises from a lack of knowledge about the true data-generating distribution, and can be reduced by collecting more data. Typically, a machine learning model uses an estimate of the (unknown) true probability distribution of the data based on a training set. Epistemic uncertainty reflects the difference between this estimate and the actual distribution, and methods estimating this type of uncertainty usually construct a (second-order) distribution over probability vectors (see Section~\ref{section:related} for a short review of these methods). 

To generate accurate and reliable predictions, modern AI systems often need to combine data from multiple modalities, such as text, images, audio, spreadsheets, and time series. This requirement has led to the emergence of a research field known as multi-modal machine learning~\citep{Baltru2018, rudovic2019, Liang2024, He24}, which aims to coherently integrate multiple modalities. This is typically achieved through multi-modal fusion strategies~\citep{Tongxin2021, Zhao2024, MENG2020115}. Such multi-modal acquisition introduces new opportunities and challenges for disentangling uncertainty. More specifically, the irreducibility assumption of aleatoric uncertainty can be challenged when information is obtained from different modalities. Future-proof general AI systems should be able to ``search for evidence'' in a cost-efficient manner to make complex decisions with high reliability. For example, in medical diagnostics, an AI assistant for patient examination could begin with simple low-cost procedures, such as blood or urine tests, and potentially follow up with more complicated (and expensive) investigations, such as MRI, CT, or genetics-based scans that uncover additional modalities. One then expects the aleatoric uncertainty about the diagnosis to decrease as more informative (yet more expensive) modalities are queried \citep{Acosta2022}. Similar AI assistants can also be useful in other application domains, such as police investigations, customs checks, and failure monitoring in industrial processes.

Motivated by potential use cases in such domains, this paper introduces a novel data acquisition framework for multi-modal data by disentangling aleatoric and epistemic uncertainties, which we believe is crucial for multi-modal sampling problems. This original approach involves reducing both types of uncertainties through querying additional instances and modalities, which will be realized with a two-stage backtracking protocol that ``searches for evidence''. Several data acquisition approaches, such as active learning~\citep{Settles2009} and active feature acquisition~\citep{Li2020ActiveFA}, have demonstrated that iteratively gathering more knowledge can enhance performance in machine learning. 
However, knowledge acquisition (whether through new observations or additional modalities) can be costly, making an effective data collection strategy essential.

Our new protocol combines active learning and active feature acquisition for multi-modal sampling. On a high level, this protocol proceeds as follows. 
For a test instance, we first assess by means of conformal prediction \citep{sadinle2019least} whether a new incoming sample should be rejected based on the uncertainty of the prediction, which is measured via the size of the conformal set. If the prediction is not rejected, our framework has statistical guarantees about the predictive performance. If the prediction is rejected, we assess in a next step whether the primary source of uncertainty is a lack of representative training data (high epistemic uncertainty) or a lack of representative features (higher aleatoric uncertainty). This uncertainty disentanglement leads to two possible actions: collect more labeled training data, or measure an additional modality for the test instance. The algorithm stops when it is sufficiently certain about the prediction, or when the two types of uncertainty cannot be further reduced. In the latter case, no prediction is made. As a practical illustration, let us consider the MIMIC-IV dataset~\citep{Johnson2023}, which is a real medical dataset that will be more thoroughly explained in the experimental section. The dataset has three modalities that can be queried to obtain reliable predictions: the online medical record of the patient, classical lab tests (such as blood sample tests) and microbiological cultures. The proposed framework can then be used to predict the disease of a new patient in a cost-efficient manner, by examining as few modalities as possible for that patient.
 
The paper is structured as follows. Section~\ref{section:related} presents related work on uncertainty disentanglement, conformal prediction, active learning and active feature acquisition. Section~\ref{section:alfa} formalizes the proposed protocol of active learning for multi-modal data acquisition, together with different methods that disentangle aleatoric and epistemic uncertainty. We also provide theoretical arguments that justify the usefulness of the protocol. Section~\ref{section:exp} presents three experiments, on a simple dataset to illustrate the behavior of the method, on the BIOSCAN-5 dataset with multiple modalities (including images and DNA sequences) and on MIMIC-IV, a real world medical dataset.  Finally, Section~\ref{section:conclusion} concludes the article.

\section{Background and related work}
\label{section:related}

\paragraph{Uncertainty disentanglement.}
To disentangle aleatoric and epistemic uncertainty in classification and regression settings, numerous methods have been proposed in recent years~\citep{huang2023}. In classification settings, most of these methods leverage the training phase to provide a prediction for a test instance in the form of a second-order distribution over first-order categorical distributions (as opposed to a classical supervised learning model, like a softmax neural network, which outputs a single probability vector that represents a single categorical distribution). For example, Bayesian methods compute a posterior distribution over the parameters of an ML model and construct a second-order distribution over class probabilities via Bayesian model averaging~\citep{kendall2017,Depeweg2018}. Conversely, evidential deep learning takes a frequentist statistics perspective and introduces specific loss functions to estimate a parametric second-order distribution over the first-order categorical distribution, without the need for specifying a prior and posterior over the parameters. For multi-class classification problems, the Dirichlet distribution is typically chosen as second-order distribution \citep{Sensoy2018,Malinin2018,Malinin2019,Ulmer2023}. Ensemble methods represent epistemic uncertainty via ensemble diversity, which can be realized by randomizing the training data via bootstrapping, randomizing a neural network architecture via techniques such as inference-time Monte Carlo Dropout, or randomizing the optimizer as in Deep Ensembles~\citep{Lakshminarayanan2017,nguyen2023learning, Mobiny2021, Shaker2020}. Some of these methods have been interpreted as providing a sample of the second-order distribution \citep{Gal2016}. Density-based and distance-based methods usually model aleatoric and epistemic uncertainty in a direct way~\citep{Nguyen2022, Hoarau2024-ml, van-amersfoort20a,sun2022,Liu2020}, but they can also construct the second-order distribution in a post-processing step~\citep{Charpentier2020,Charpentier2022}.
All methods can disentangle and measure both aleatoric and epistemic uncertainty using metrics like entropy, variance, or pseudo-counts \citep{wimmer23a,Schweighofer2023,Sale2023,Duan2024}. 

\paragraph{Conformal prediction (CP).}
CP methods focus on constructing adaptive prediction sets that reflect the uncertainty of individual predictions while maintaining rigorous, distribution-free coverage guarantees \citep{sadinle2019least}. Using nonconformity scores, these methods adapt the size of the prediction set based on aleatoric and epistemic uncertainty, but the two sources of uncertainty are not disentangled. CP only provides a marginal coverage guarantee, which implies that the desired coverage is typically not achieved in difficult regions of the feature space, while a too high coverage is observed in easier regions. Recent papers try to improve conditional coverage by introducing novel nonconformity scores such as the popular adaptive prediction sets (APS) score \citep{romano2020classification}, or by grouping instances into clusters \citep{ding2023class}. In our work, we apply APS-based conformal prediction to decide whether a reliable decision can be made for a test instance. We consider other techniques for uncertainty disentanglement, but analyzing the influence of aleatoric and epistemic uncertainty in the context of conformal prediction is an important open research question that is currently receiving a lot of attention \citep{karimi2024evidential,rossellini2024integrating,cabezas2025epistemic,marques2025quantifyingaleatoricepistemicdynamics,Stutts,correia2025informationtheoreticperspectiveconformal}.   

\paragraph{Active learning.} Given the cost of collecting labels in classification, a trade-off between performance and labeling cost has been proposed. 
Active learning (AL)~\citep{Settles2009} aims to select the most relevant observations to be labeled, thereby reducing the number of labeled data points and the associated costs~\citep{Hacohen2022}. 
The most common approach is pool-based sampling, where classes of unlabeled observations (referred to as \emph{instances} in AL) are queried one by one or in batches. Various methods have been proposed, such as random sampling, where labels are queried randomly as the labeled dataset grows. 
The method most relevant to our paper is uncertainty sampling~\citep{Lewis1994}, where the learner tries to reduce its uncertainty by querying the labels for which it would be the least certain. In our work we will use a variant called epistemic uncertainty sampling, which focuses on the regions where epistemic uncertainty can be reduced. This sampling method has been implemented in the past using various second-order uncertainty quantification techniques, such as density-based methods~\citep{Nguyen2022, Hoarau2024-ml} or Bayesian approaches~\citep{Houlsby2011, shelmanov-2021}.

\paragraph{Active feature acquisition.} Active learning focuses on selecting which observations and labels to query, while active feature acquisition concentrates on the variables. Many problems are characterized by missing variables, and feature acquisition offers potential solutions for addressing variable collection, the goal being to collect the most informative variables during training to improve the model~\citep{Zheng2002}. This should be distinguished from feature collection at test stage, whose purpose is to minimize misclassification costs (or a variant thereof) during the test phase~\citep{GREINER2002137, quost21a}. Recently, several methods for active feature acquisition have been applied to deep architectures and multi-modal data~\citep{Li2020ActiveFA, NEURIPS2018_e5841df2,kossen2023,rudovic2019}. 
Our combined active learning and active feature acquisition approach leverages uncertainty disentanglement at both training and prediction time to improve reliability, by focusing on both observations and variables from different modalities.

\section{Active learning with multi-modal data acquisition}\label{section:alfa}

\subsection{Formal description of the algorithm}\label{section:alfa_setup}

We first describe our Active Learning with Multi-modal data Acquisition (ALMA) algorithm on a high level. Given a new test observation, the framework assesses whether the model's prediction should be rejected or not based on conformal prediction. If the prediction is not rejected, which happens when the predicted set size is one, the process can be stopped with statistical coverage guarantees. If the prediction is rejected, the model's EU (Epistemic Uncertainty) and AU (Aleatoric Uncertainty) can be assessed --- see Section~\ref{section:unc_quant} for a more precise description. EU reflects the model's level of knowledge about a given problem. If this uncertainty is too high, new observations similar to the one being processed must be added to the training set to reduce the model's epistemic uncertainty. AU represents the inherent difficulty of predicting the class given the current variables. If the AU is too high, a new representation space is created by acquiring a new modality. This process is repeated till a reliable prediction can be returned (conformal set size one), or till all modalities are queried. In the former case, we make a reliable prediction with statistical guarantees. In the latter case, no reliable prediction can be returned for the test instance.  

More formally, in a multi-class classification setting, we consider a sequence of $\mu$ input spaces $\mathcal{I} = (\mathcal{X}^1, \dots, \mathcal{X}^\mu)$ associated with the same output space $\mathcal{Y} = \{y_1,\dots,y_K\}$, with $K\geq2$. For each input space we assume $\mathcal{X}^j = \mathbb{R}^{P_j}$, with $j \leq \mu$, $\mathcal{X}^j$ being a collection of $P_j$ random variables from modality $j$. We denote by $\textbf{x}^j$ an instance in space $\mathcal{X}^j$, and by $(\textbf{x}_i^j, y_i)$ a labeled training instance sampled \emph{i.i.d} from an unknown probability distribution $\mathbb{P}^j$. 
We assume that the collection $\mathcal{I}$ has a fixed order and is uncovered iteratively via Algorithm~\ref{algo:method}, starting from $\mathcal{X}^1$ and stopping at $\mathcal{X}^\mu$ when the algorithm ends. We assume that modalities are ordered by acquisition cost, with a higher cost being compensated by the decisiveness of the underlying information. Remark that this setup is different from some previous works on active feature acquisition, which assume that the order in which features are queried varies over test instances~\citep{pmlrclertant19a}.

\paragraph{Training phase.} We first describe the training phase, before predictions are made for test instances. Given $\mu$ hypothesis spaces $\mathcal{H}^j$, $j \leq \mu$, composed of hypotheses $h^j : \mathcal{X}^j \rightarrow \Delta^{K-1}$ mapping any instance $\textbf{x}^j$ to a probability vector in the probability simplex $\Delta^{K-1}$. In addition, we assume that these classifiers also provide measures of aleatoric and epistemic uncertainty, further denoted as $AU^j(\textbf{x})$ and {$EU^j(\textbf{x})$, and explained more thoroughly in Section~\ref{section:unc_quant}. 
We train models $\hat{h}^j$ based on a labeled training set $\mathcal{D_L}^j = \{(\textbf{x}_i^j, y_i)|i\in\Phi_\mathcal{L}^j\}$, for all $j \leq\mu$ (the indices $i\in\Phi_\mathcal{L}^j$ point out to the labeled instances). 
For computing the threshold on the nonconformity score in CP we also need calibration datasets $\mathcal{D}_\mathrm{Cal}^j = \{(\textbf{x}_i^j, y_i)\}_{i=1}^{n_\mathrm{cal}}$, \emph{i.i.d.}\ sampled from $\mathbb{P}^j$ for all $j \leq\mu$. 
For active learning purposes, we assume that we also have access to an unlabeled training set $\mathcal{D_U}^j = \{\textbf{x}_i^j|i\in\Phi_\mathcal{U}^j\}$ with instances defined in the same space $\mathcal{X}^j$. 

\begin{figure}[t]
\centering
\begin{minipage}{0.5\linewidth}
    \begin{algorithm}[H]
    \captionsetup{font=small}
    \caption{ALMA}
    \small
    \label{algo:method}
    \begin{algorithmic}[1]
        \Require $\textbf{x}$, $\mathcal{D_L}^j$, $\mathcal{D_U}^j$
        \State $j \gets 1$ 
        \State Train model on $\mathcal{D_L}^j$
        \State Compute $\Upsilon^j($\textbf{x}$)$, $AU^j(\textbf{x})$ and $EU^j(\textbf{x})$
        \While{$|\Upsilon^j($\textbf{x}$)| \neq 1$ and $j < \mu$}  
            \If{$EU^j(\textbf{x}) \geq AU^j(\textbf{x})$} 
                \State \textbf{$\textbf{x}_{\text{new}}$} $\gets$ Query\_strategy($\textbf{x}$, $\mathcal{D_U}^j$)
                \State $y_{\text{new}} \gets$ Query\_label(\textbf{$x_{\text{new}}$})
                \State Move$(\textbf{$\textbf{x}_{\text{new}}$}, y_{\text{new}})$ from $\mathcal{D_U}^j$ to $\mathcal{D_L}^j$
            \Else
                \State $j \gets j + 1$  \Comment{Retrieve additional modality}
            \EndIf
            \State Train model on $\mathcal{D_L}^j$
            \State Compute $\Upsilon^j($\textbf{x}$)$, $AU^j(\textbf{x})$ and $EU^j(\textbf{x})$
        \EndWhile
        \State Return $\Upsilon^j($\textbf{x}$)$ \Comment{Return prediction}
    \end{algorithmic}
    \end{algorithm}
\end{minipage}
\hfill
\begin{minipage}{0.44\linewidth}
\centering
    \includegraphics[width=\textwidth]{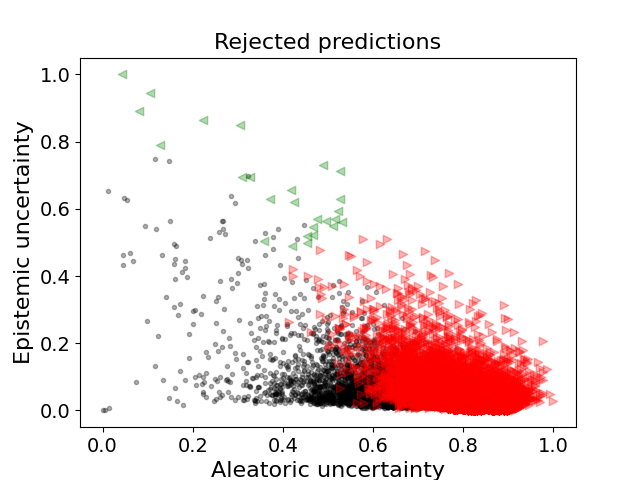}
    \captionof{figure}{Rejections at a training stage for test instances using Online Medical Record (1 modality from MIMIC-IV) to predict diseases with Deep Ensemble: black for reliable, green for EU-based rejection, red for AU-based rejection.}\label{fig:omr}
\end{minipage}
\end{figure}

\begin{figure}
    \centering
    \begin{subfigure}{0.19\linewidth}
    \includegraphics[width=\linewidth]{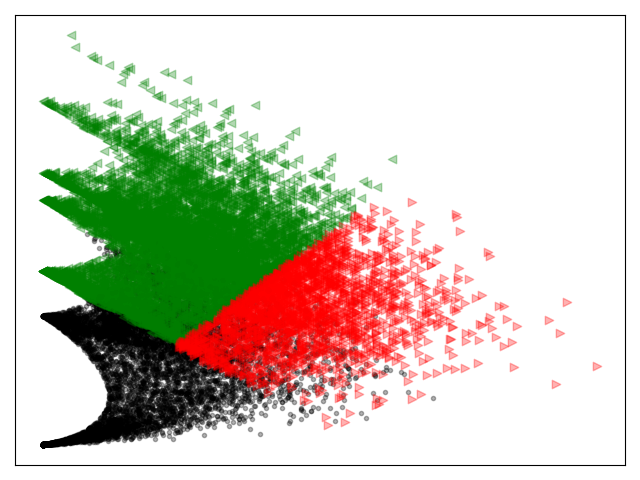}
    \caption{5k}
    \end{subfigure}
    \begin{subfigure}{0.19\linewidth}
    \includegraphics[width=\linewidth]{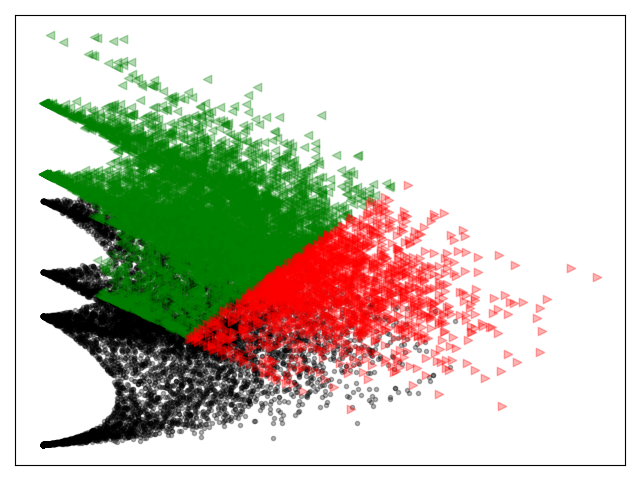}
    \caption{10k}
    \end{subfigure}
    \begin{subfigure}{0.19\linewidth}
    \includegraphics[width=\linewidth]{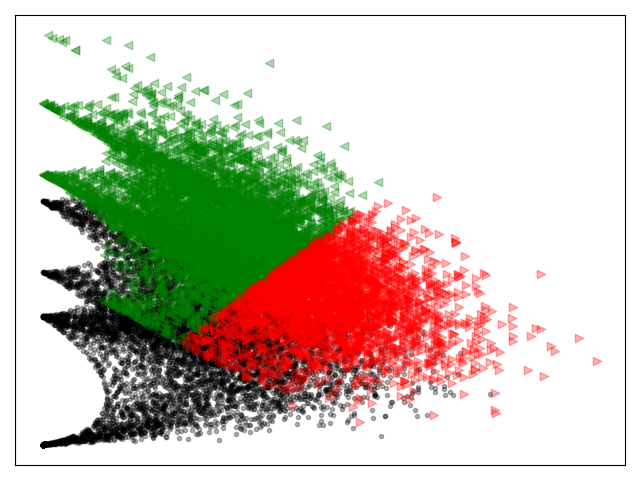}
    \caption{20k}
    \end{subfigure}
    \begin{subfigure}{0.19\linewidth}
    \includegraphics[width=\linewidth]{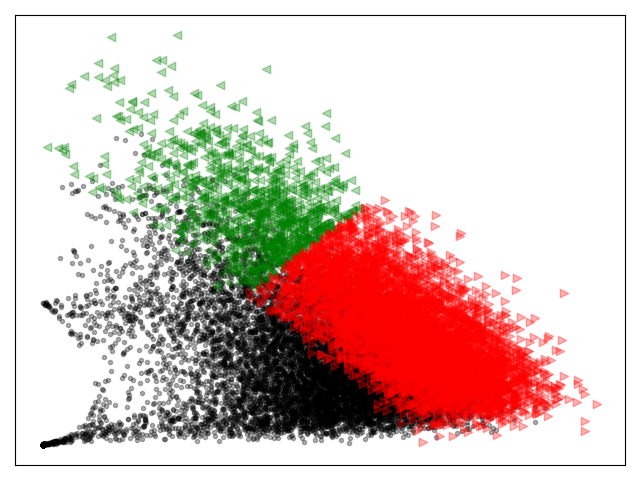}
    \caption{40k}
    \end{subfigure}
    \begin{subfigure}{0.19\linewidth}
    \includegraphics[width=\linewidth]{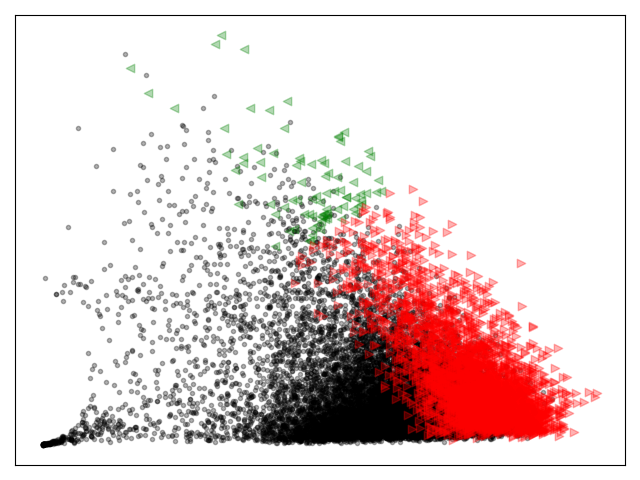}
    \caption{100k}
    \end{subfigure}
    \caption{Rejections across training stages using multiple modalities (Online Medical Records + Laboratory Tests) with Deep Ensemble: black for reliable, green for EU-based rejection, red for AU-based rejection. Increasing the training set reduces rejections based on epistemic uncertainty.}\label{figure:lab_reject}
\end{figure}

\paragraph{Inference phase.} Algorithm~\ref{algo:method} sketches our active learning with multi-modal acquisition strategy. When one needs to predict the class of a new incoming instance $\textbf{x}$, the first input space (generally composed of the cheapest variables or modalities, \textit{i.e.} the easiest to acquire) is queried such that the instance space of the new incoming instance $\mathcal{X}_\textbf{x}$ is compatible with $h^1$, $\mathcal{X}^1 \subseteq \mathcal{X}_\textbf{x}$. A prediction $\hat{y} = \hat{h}^1(\textbf{x})$ can thus be made. 
Unlike in the classical classification setting, this decision is in our procedure not necessarily final. If the conformal set $\Upsilon^1(\textbf{x})$ constructed from the APS nonconformity score only contains one class ($|\Upsilon^1(\textbf{x})| = 1$), we consider the prediction reliable and the process stops with statistical coverage guarantees. 
If the conformal set contains multiple classes ($|\Upsilon^1(\mathbf{x})| > 1$), the prediction is rejected and both AU and EU are computed. If $EU^1(\textbf{x}) \geq AU^1(\textbf{x})$), a new observation from the unlabeled set $\mathcal{D_U}^1$ is added to the labeled set $\mathcal{D_L}^1$ together with its true class in $\mathcal{Y}$. Observations can be added one by one, or by batches --- the latter is called pool-based sampling in active learning~\citep{Settles2009}. 

Otherwise, if $AU^1(\textbf{x}) \geq EU^1(\textbf{x})$, new modalities are added to $\mathcal{X}_\textbf{x}$ such that the instance space of the new incoming instance $\mathcal{X}_\textbf{x}$ is compatible with $h^2$, $\mathcal{X}^2 \subseteq \mathcal{X}_\textbf{x}$. These two steps are repeated with $EU^2$ and $AU^2$ and the process continues with $EU^j$ and $AU^J$, for $j \leq \mu$. 
Whenever the conformal set $\Upsilon^j(\textbf{x})$  for any model $\hat{h}^j$ is a singleton, then the state of knowledge is deemed sufficient for a decision to be safely made, and the process is stopped. The process is also stopped when all modalities are acquired, but then no reliable prediction can be made if the conformal set is still not a singleton. 
By simply comparing $EU$ and $AU$ values in Line 5 of Algorithm~\ref{algo:method}, we assume that they are commensurate and of equal importance, which is a reasonable default assumption. We will also see in Section~\ref{section:exp} that this heuristic yields good empirical results --- see Figures~\ref{fig:omr} and~\ref{figure:lab_reject}, which are more thoroughly discussed in Section~\ref{section:exp}, for a teaser.   

\paragraph{Conformal prediction.} The calibration datasets $\mathcal{D}^j_{\textrm{cal}}$ all remain fixed during the ALMA algorithm, which is a requirement  for the calibration dataset to be exchangeable with the future test point $(\textbf{x}^j_{n+1}, y_{n+1})$. For any nonconformity score function $s:\mathcal{Y} \times \Delta^{K-1} \to \mathbb{R}$ and any user-adjusted coverage rate $1 - \alpha$, split conformal prediction  constructs sets as follows:
\begin{equation*}
    \label{eq:conformal:set}
    \Upsilon^j(\textbf{x}^j_{n+1}) = \{ y: s(\textbf{x}^j_{n+1},y) \leq q\} \quad \text{for} \quad q := Q(1 - \alpha;\{s(\textbf{x}^j_i, y_i): (\textbf{x}^j_i, y_i) \in \mathcal{D}^j_{\textrm{cal}}\}),
\end{equation*} 
where $Q(1 - \alpha;\mathcal{A})$ is the $(1 - \alpha)\cdot(1 + \frac{1}{n+1})$ quantile of the set $\mathcal{A}$. CP guarantees that $\Pr[y_{n+1} \in \Upsilon^j(\textbf{x}^j_{n+1})] \ge 1 - \alpha$. This guarantee holds only marginally; \emph{i.e.,} in expectation over the inputs drawn from the same distribution as the calibration set. In our work we use Adaptive Prediction Sets (APS) \citep{romano2020classification} as nonconformity score, which is known to improve conditional coverage compared to more traditional nonconformity scores. The APS score includes labels in descending order with respect to their predicted probabilities until the cumulative probability exceeds a calibrated threshold. Formally, the APS score is defined as $s(\textbf{x}^j, y) := \sum_{y' \in \mathcal{Y}} \pi(y'\mid\textbf{x}^j) \cdot \mathbb{1} \{\pi(y'\mid\textbf{x}^j) \geq \pi(y\mid\textbf{x}^j) \}$. This is the cumulative probability of classes ranked above $y$, and $\pi(y\mid \textbf{x}^j)$ are the estimated class probabilities of classifier $\hat{h}^j$. A randomization term can also be incorporated into the nonconformity score to prevent over-coverage \citep{romano2020classification}, but this term is omitted in our work. A too high coverage is for us not an issue, but deterministic predictions are preferred. 

The described setup relies on two important assumptions. The first is that new relevant instances similar to $\mathbf{x}$ are available at inference time, along with sufficient computational resources. This enables a reduction in epistemic uncertainty. If a new sample is considered unknown, the model must have access to similar instances that can be added to the labeled set. Note that this is a common limitation of classical active learning. Solutions have been proposed, such as online learning, but they are not always feasible in practice.
The second assumption is that one can acquire new modalities for an incoming sample at test time as well. This is particularly relevant in clinical settings: for instance, when assisting a doctor, additional data such as an MRI or an EEG may need to be obtained for a specific patient.

\subsection{Uncertainty disentanglement}\label{section:unc_quant}

Our study investigates multiple state-of-the-art approaches to disentangle aleatoric and epistemic uncertainty. 
Among the four main families of approaches discussed in Section~\ref{section:related}, we focus on Bayesian deep learning, ensemble-based and distance-based methods. Ensemble methods have been interpreted as Bayesian estimators by generating samples from the posterior second-order distribution~\citep{Mobiny2021, Shaker2020}, but we also incorporate Bayesian methods in an explicit way. We do not consider evidential deep learning methods, since they have faced increasing criticism~\citep{Meinert2022TheUE, Bengs2023, jurgens2024} regarding EU estimation, although it could be interesting to combine ALMA with recent methods that address these criticisms. We also consider various metrics to quantify aleatoric and epistemic uncertainty, of which Appendix~\ref{app:methods} provides a formal description, while a higher-level overview is given here. 

For entropy-based uncertainty estimation, we implemented the well-known information-theoretic decomposition, which was introduced by~\citep{Depeweg2018} for Bayesian methods, and later used in tandem with many other uncertainty disentanglement methods~\citep{Malinin2018,Malinin2019,Charpentier2020,Charpentier2022}. The total uncertainty is the Shannon entropy of the mean of the second-order distribution, while the aleatoric uncertainty is the mean entropy of individual members of the second-order distribution. The epistemic uncertainty is then obtained by subtracting the aleatoric uncertainty from the total uncertainty, assuming an additive decomposition. We implemented this information-theoretic decomposition together with Random Forests, Deep Ensembles and Bayesian Deep Learning. For Random Forests we used the classical ensemble of decision trees trained with bootstrap aggregation (bagging) and random variable selection~\citep{Breiman1996}. For Deep Ensembles~\citep{Lakshminarayanan2017}, we trained multiple neural networks with random weight initialization and a randomized training set order (without bagging). We used Laplace approximation~\citep{daxberger2022} to train a Bayesian deep network and sampled predictions from this second-order distribution~\citep{Mobiny2021}.

For variance-based uncertainty estimation, we investigated the label-wise variance decomposition~\citep{Sale2024-b}. Total uncertainty is defined for each class as the variance of the outcomes, and global uncertainty is obtained by summing over all classes, while aleatoric uncertainty is defined as the expected conditional variance. This decomposition is applied to the same range of models as previously mentioned.

For distance-based methods, we examined the centroid-based approach presented in~\citep{van-amersfoort20a}. To compute the model's uncertainty, the authors proposed measuring the distance from the incoming observation to the class centroids using a radial basis function. The epistemic uncertainty is obtained by measuring the distance from the incoming observation to the nearest class centroid, while aleatoric uncertainty is obtained based on the relative distances between all centroids. The second distance-based method~\citep{Hoarau2024-ml} computes EU and AU based on the nearest neighbors of the incoming test instance. Discord among the neighbors indicates high aleatoric uncertainty, while a substantial distance from all neighbors indicates high epistemic uncertainty.

\subsection{Theoretical guarantees}
\label{section:theory}

This section provides a formal characterization of the conditions under which the acquisition of additional modalities or labels leads to a reduction in predictive uncertainty within the ALMA framework. Our analysis relies on standard information-theoretic concepts and follows the multi-modal setting introduced in Section~\ref{section:alfa_setup}. We denote by $\textbf{x}^{1:j}$ a sample on the concatenation of the first $j$ modalities and we use $H(\cdot)$ and $I(\cdot\,,\cdot)$ to represent the Shannon entropy and mutual information, respectively. 

\begin{property}[Monotonicity of aleatoric uncertainty wrt adding modalities]
\label{thm:monotone-AU}
For any $1 \le j < j' \le \mu$:
\begin{equation}
H\!\left(y\,\middle|\,\textbf{x}^{1:j'}\right) \;\le\; H\!\left(y\,\middle|\,\textbf{x}^{1:j}\right).
\end{equation}
\end{property}
\begin{proof}
The inequality is a consequence of entropy being reduced by conditioning:
\begin{multline}
        H(y\,|\,\textbf{x}^{1:j}) - H(y\,|\,\textbf{x}^{1:j'}) \;=\; I\!\left(y, \textbf{x}^{j+1:j'} \,\middle|\, \textbf{x}^{1:j}\right) 
        \\\quad\Leftrightarrow\quad
        H(y\,|\,X^{1:j}) \;=\; H(y\,|\,\textbf{x}^{1:j'}) + I\!\left(y, \textbf{x}^{j+1:j'} \,\middle|\, \textbf{x}^{1:j}\right). 
\end{multline}
The equality holds if and only if joint mutual information is null, \textit{i.e.}, the additional modalities $\textbf{x}^{j+1:j'}$ bring no information about $y$ beyond $\textbf{x}^{1:j}$. 
\end{proof}

Property~\ref{thm:monotone-AU} rigorously supports the modeling choice in ALMA: querying a richer modality space can only shrink the irreducible part of uncertainty, and it shrinks it exactly by the conditional information $I(y, \textbf{x}^{j+1:j'} \mid \textbf{x}^{1:j})$. In particular, modalities for which this quantity is (near) zero are provably unhelpful for reducing aleatoric uncertainty, a phenomenon visible in Table~\ref{tab:mimic} for some diseases. 

\begin{property}[Monotonicity of the entropy-based epistemic term]
\label{prop:epistemic-entropy-simple-en}
Let a random variable $\Theta=(\Theta_1,\dots,\Theta_K)\in\Delta^{K-1}$ represent the epistemic uncertainty over the categorical distribution $p=(p_1,\dots,p_K)$, with $\mathbb{E}[\Theta]=p$ \footnote{In practice, specific realizations of $\Theta$ can be viewed as approximate samples from the posterior distribution in ensemble learning (see Appendix~\ref{app:methods}), which itself provides an estimate of the true Bayesian posterior. In Evidential Deep Learning, by contrast, $\Theta$ is typically modeled using a Dirichlet distribution.}.
The entropy-based epistemic uncertainty term is defined as:
\begin{equation}
EU := H(p)-\mathbb{E}[H(\Theta)].
\end{equation}
Given an incoming instance $\textbf{x}$, this becomes $EU(\textbf{x}) = TU(\textbf{x}) - AU(\textbf{x})$, as mentioned in Appendix~\ref{app:methods}. Assume that as the number of observations $n$ increases, the posterior over $\Theta$ becomes more concentrated around $p$, \emph{i.e.}\ $\operatorname{Var}(\Theta_i)=O(1/n)$ for all $i$.  
Then: 
\begin{enumerate}
\item $EU\ge0$ for all $n$, with equality if and only if $\Theta$ is degenerate at $p$; 
\item $EU$ decreases as the posterior sharpens (\emph{i.e.}, as its variance decreases); 
\item the following second-order approximation holds: 
\begin{equation}
EU \;\approx\;
\tfrac12\sum_{i=1}^K \frac{\operatorname{Var}(\Theta_i)}{p_i},
\end{equation}
and in particular $EU=O(1/n)$ whenever $\operatorname{Var}(\Theta_i)=O(1/n)$. 
\end{enumerate}
\end{property}
\begin{proof}
By concavity of the entropy $H$, Jensen’s inequality gives 
$\mathbb{E}[H(\Theta)]\le H(\mathbb{E}[\Theta])=H(p)$,
and therefore $EU\ge0$. To relate $EU$ to the dispersion of $\Theta$, we expand $H$ by second-order Taylor approximation around $p$:
\begin{align*}
\nonumber 
H(\Theta) &= H(p)
+\nabla H(p)^{\!\top}(\Theta-p)
+\tfrac12(\Theta-p)^{\!\top}\nabla^2 H(p)(\Theta-p)
+R_3 , \\
 &= H(p) + \sum_{i=1}^K(-\log p_i - 1)(\Theta_i-p_i) + \frac{1}{2} \sum_{i=1}^K\left(-\frac{1}{p_i}\right)(\Theta_i-p_i)^2 + R_3 . 
\end{align*}
Taking expectations and using $\mathbb{E}[\Theta]=p$ yields:
\begin{equation*}
    \mathbb{E}[H(\Theta)] = H(p)-\tfrac12\sum_{i=1}^K\frac{\operatorname{Var}(\Theta_i)}{p_i}+R_3,
\end{equation*}
where $R_3\to 0$ as the posterior over $\Theta$ concentrates around $p$. Therefore we obtain:
\begin{equation*}
EU
\approx\tfrac12\sum_{i=1}^K\frac{\operatorname{Var}(\Theta_i)}{p_i} ; 
\end{equation*}
as the posterior concentrates (variances $\!\downarrow 0$),
the leading term decreases monotonically, yielding $EU\!\to0$.
\end{proof}

Intuitively, $EU$ quantifies the overestimation of uncertainty due to epistemic variability in $\Theta$. When the posterior is flat, the gap 
$H(p)-\mathbb{E}[H(\Theta)]\!\approx\tfrac12\sum_i \operatorname{Var}(\Theta_i)/p_i$
is significant. As data accumulate and the posterior sharpens, this gap becomes negligible, typically of order $O(1/n)$. This motivates our approach in ALMA: by monitoring the decrease in $EU$, we can identify a practical threshold beyond which the epistemic component becomes negligible and the remaining uncertainty is predominantly aleatoric, \textit{i.e.}, it corresponds to the true irreducible uncertainty in the data. 

\section{Experiments}\label{section:exp}

\begin{figure}
    \centering
    \includegraphics[width=0.55\linewidth]{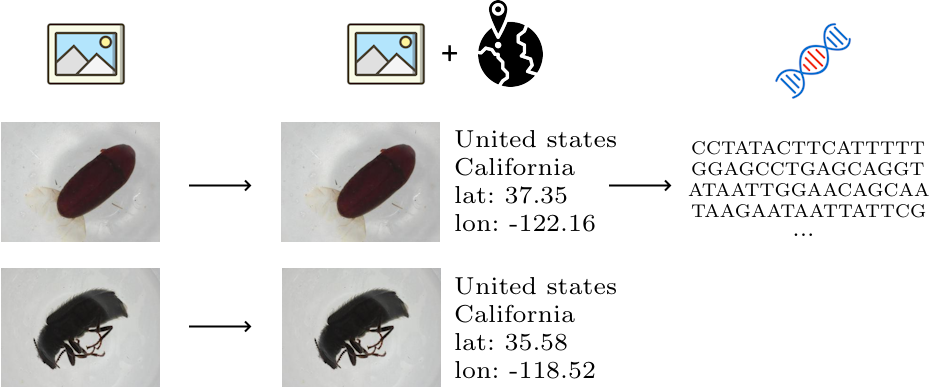}
    \caption{BIOSCAN-5M: ALMA on two test instances. For the first insect, all three modalities are needed to reduce aleatoric uncertainty, while for the second, adding geographical data to the image suffices for reliable prediction.}
    \label{figure:alfa_bioscan}
\end{figure}

\subsection{Experimental setup}

In this section, we present the results obtained using our ALMA framework on three multi modal datasets. A total of eight uncertainty disentanglement methods are under study. Deep Ensemble, Bayesian Deep Learning, and Random Forest are combined with both entropy and variance-based decomposition, while the two distance-based methods are also examined.
The tabular Wine dataset~\citep{Dua2019} is used to classify different types of wine by means of variables grouped in modalities according to an assigned fictitious cost. The four modality groups are an ignition study, a visual study, a chemical study, and an acidity study.
BIOSCAN-5M~\citep{gharaee2024bioscan5m} is an insect classification dataset, including images (photos), geographical information, and DNA sequences of various insects.
MIMIC-IV\citep{Johnson2023} is a freely accessible electronic health record dataset. Information available includes patient measurements, orders, diagnoses, procedures, treatments, and de-identified free-text clinical notes. We extracted three different modalities to detect the eight diseases presented in Table~\ref{tab:mimic}. The Online Medical Record (OMR) contains information such as gender, age, height, BMI, weight and blood pressure, which do not require expensive procedures. Classical laboratory tests (from blood samples) are also conducted and grouped into 40 variables, along with a third modality comprising results from microbiological cultures grouped into 50 additional variables.

To obtain a comprehensive understanding of the method's behavior, a model is trained on each modality group and with varying sample sizes  for each variable space, ordered by acquisition cost. In BIOSCAN-5M, we used $50k$ training instances and $5k$ test instances uniformly distributed over 10 classes. One ResNet18 is trained solely on insect normalized images. We employed a cross-entropy loss with SGD, using a learning rate of 0.1, momentum of 0.9, and a weight decay of $5 \times 10^{-4}$. Each batch had a size of $100$, and the models were trained for 10 epochs. For the second model, after training, a reduced feature space is extracted from the ultimate layer of the ResNet18. This feature space is then enriched with the geographical information: country and state are encoded using one-hot encoding, while latitude and longitude are included as numerical values. A third model is based on the individual's DNA sequence (a high-cost approach, as it requires sequencing the insect's DNA). We used $k$-mer frequency feature extraction with $k=3$ to encode the DNA sequences. Each model is trained from $1k$ sample to $20k$ according to random sampling. For the Wine and MIMIC-IV datasets, modalities are combined as they consist of variables of the same types. For MIMIC-IV we used $100k$ training instances and $4k$ test instances distributed with high imbalance over 8 diseases. Models are trained from $5k$ instances to $100k$ according to random sampling. The dataset
also contains several missing values for which we applied a mean imputation method. In the experiments, we considered a risk level of $\alpha = 0.05$, except for the experiment on MIMIC-IV, which did not result in a model capable of achieving 95\% disease detection accuracy, and where we instead required 50\% accuracy. 
Details of each experiment are provided in  Appendix~\ref{app:wine},~\ref{app:bioscan} and~\ref{app:mimic}.

\begin{table}
    \centering
    \small
    \begin{tabular}{@{}lccc@{}}
    \toprule
    Disease & OMR \;\;\;\;\;\;\;\;\;\;\;\;\;& Lab. tests \;\;\;\;\;\;\;\;\;\;& Microb. culture\\
    \midrule
    Hyperlipidemia & 77\% & 66\% & 67\%\\
    Hypertension & 21\% & 30\% & 27\%\\
    Acute kidney failure & 14\% & 45\% & 44\%\\
    Alcoholic cirrhosis of liver \;\;\;\;\;\;\;\;\;\;\;\;\;& 1\% & 59\% & 62\%\\
    End stage renal disease & 1\% & 65\% & 64\%\\
    Urinary infection & 18\% & 26\% & 35\%\\
    E. coli infection & 0\% & 2\% & 40\%\\
    Pseudomonas infection & 0\% & 1\% & 30\%\\
    \bottomrule\\
    \end{tabular}
    \caption{MIMIC-IV \emph{vs} Deep Ensemble: diseases from real patients \emph{vs} class-wise recall based on different modalities. Restrictive and costly tests are not always useful.}\label{tab:mimic}
\end{table}

\subsection{Illustration of the ALMA procedure}

\begin{figure}
    \centering
    \begin{subfigure}{0.24\linewidth}
    \includegraphics[width=\linewidth]{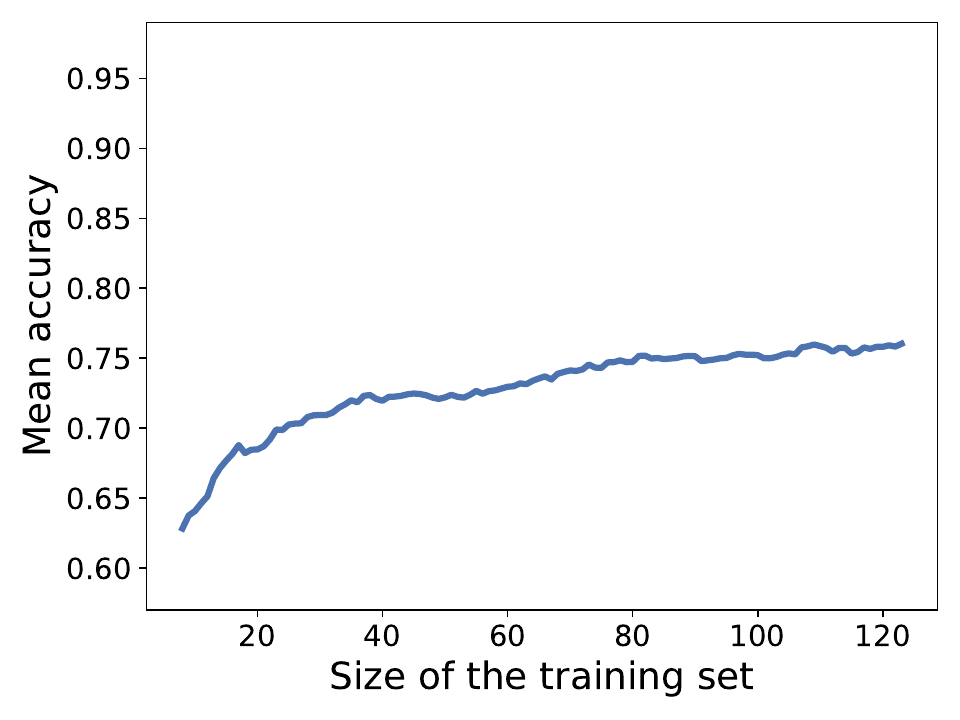}
    \caption{\footnotesize Ignition}\label{figure:winge_ig1}
    \end{subfigure}
    \begin{subfigure}{0.24\linewidth}
    \includegraphics[width=\linewidth]{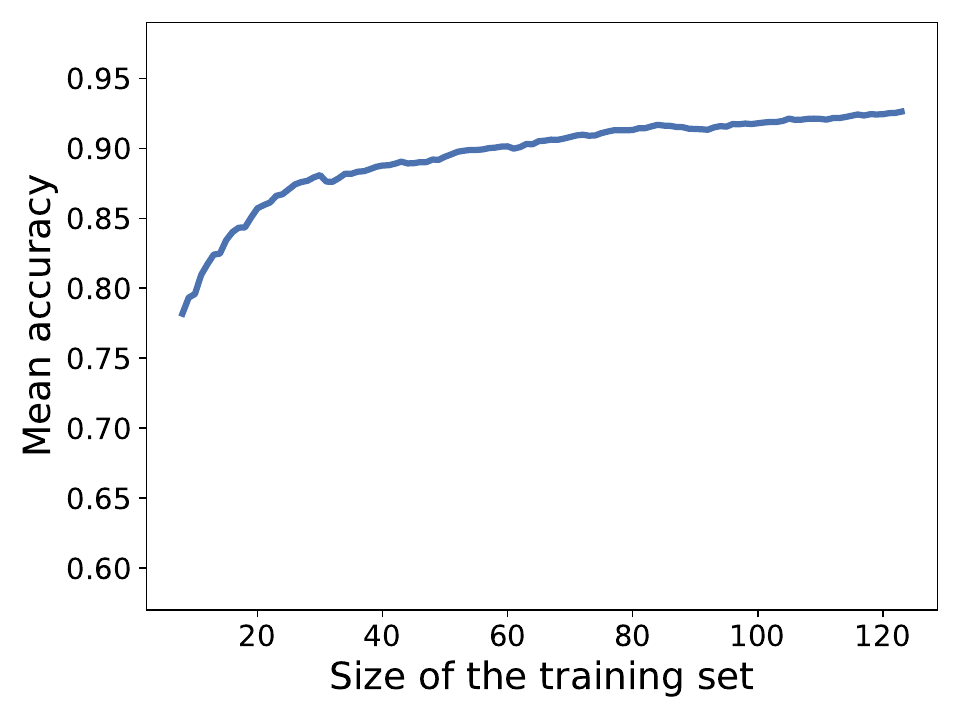}
     \caption{\footnotesize\emph{Former +} Visual}
    \end{subfigure}
    \begin{subfigure}{0.24\linewidth}
    \includegraphics[width=\linewidth]{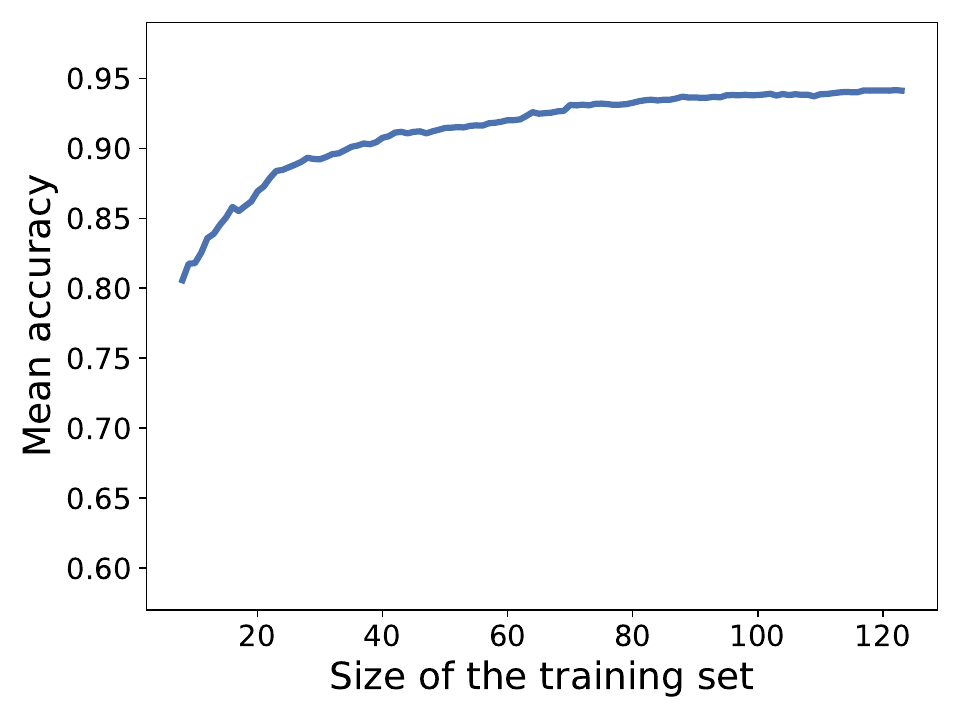}
    \caption{\footnotesize\emph{Former +} Chemical}
    \end{subfigure}
    \begin{subfigure}{0.24\linewidth}
    \includegraphics[width=\linewidth]{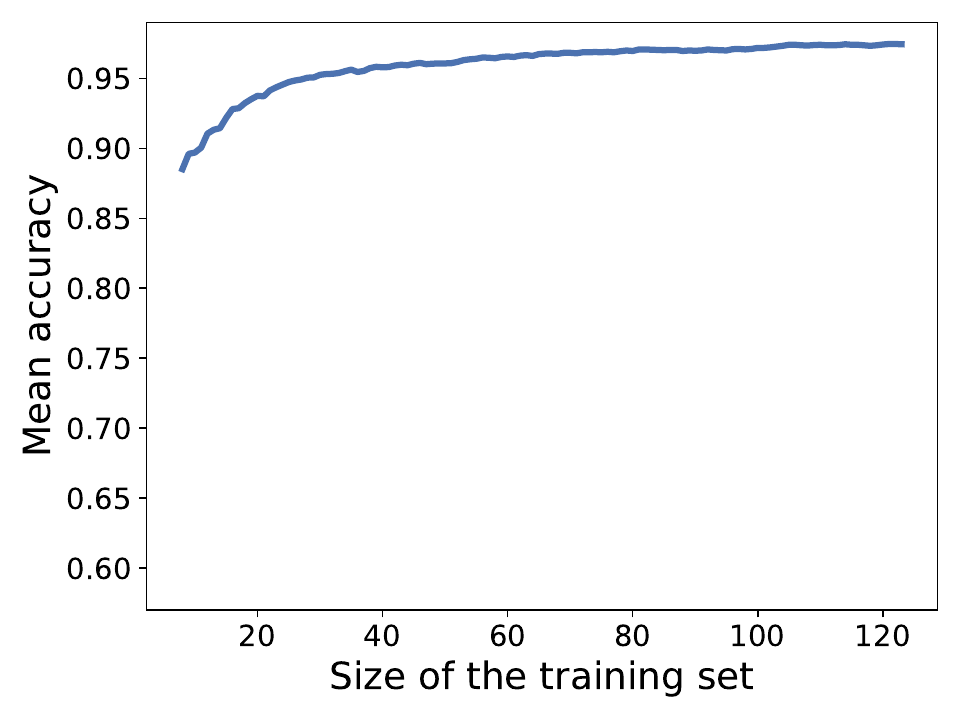}
    \caption{\footnotesize\emph{Former +} Acidity}\label{figure:winge_ac1}
    \end{subfigure}
    \begin{subfigure}{0.24\linewidth}
    \includegraphics[width=\linewidth]{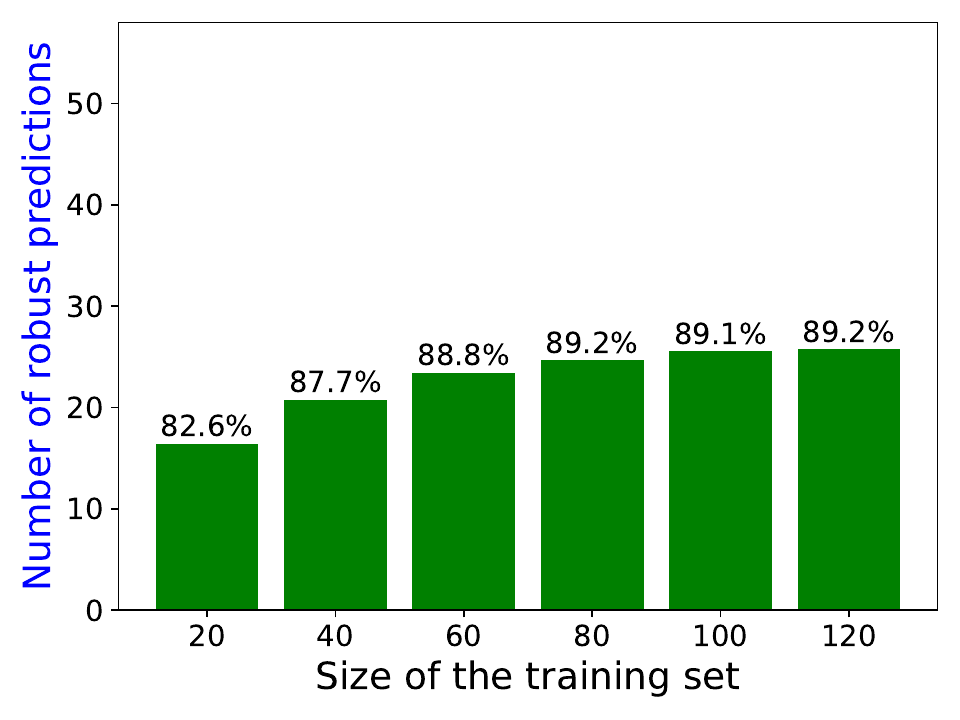}
    \caption{\footnotesize Ignition}\label{figure:winge_ig2}
    \end{subfigure}
    \begin{subfigure}{0.24\linewidth}
    \includegraphics[width=\linewidth]{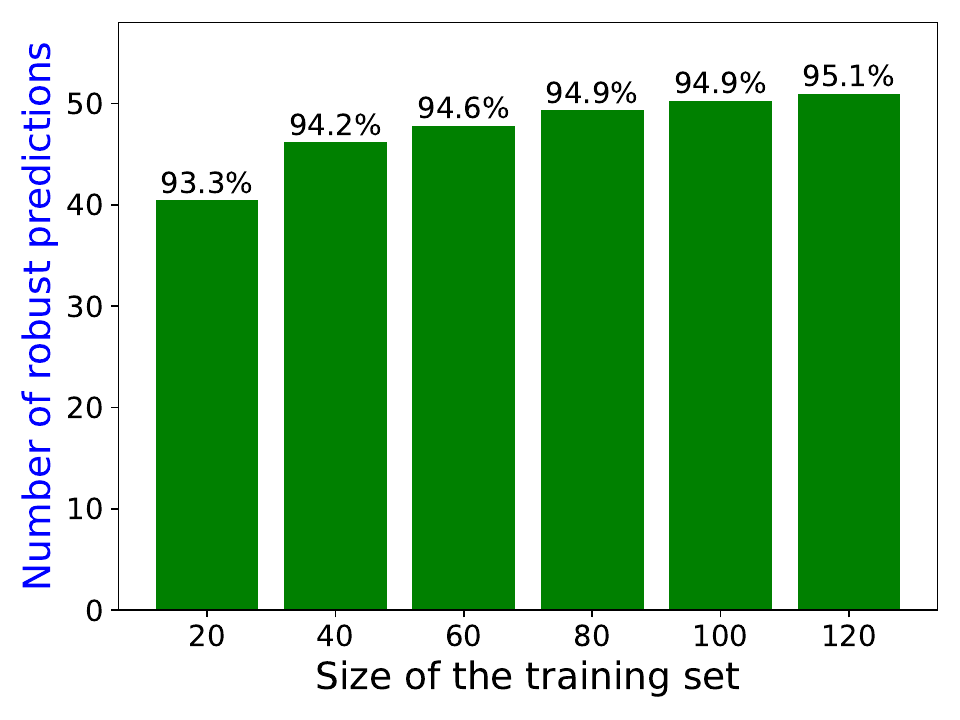}
    \caption{\footnotesize\emph{Former +} Visual}
    \end{subfigure}
    \begin{subfigure}{0.24\linewidth}
    \includegraphics[width=\linewidth]{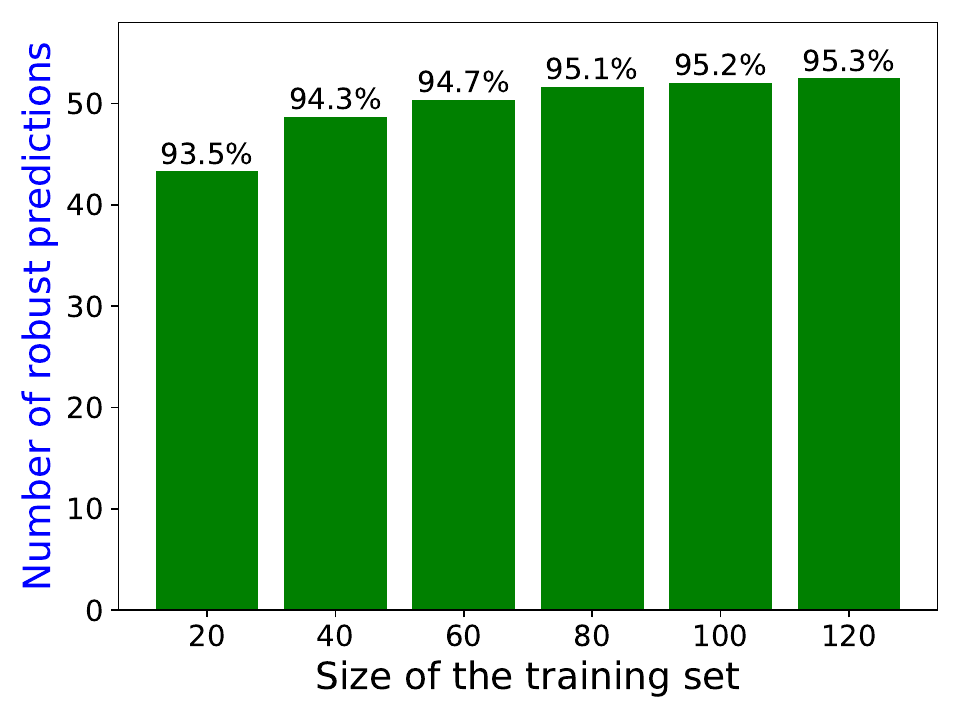}
    \caption{\footnotesize\emph{Former +} Chemical}
    \end{subfigure}
    \begin{subfigure}{0.24\linewidth}
    \includegraphics[width=\linewidth]{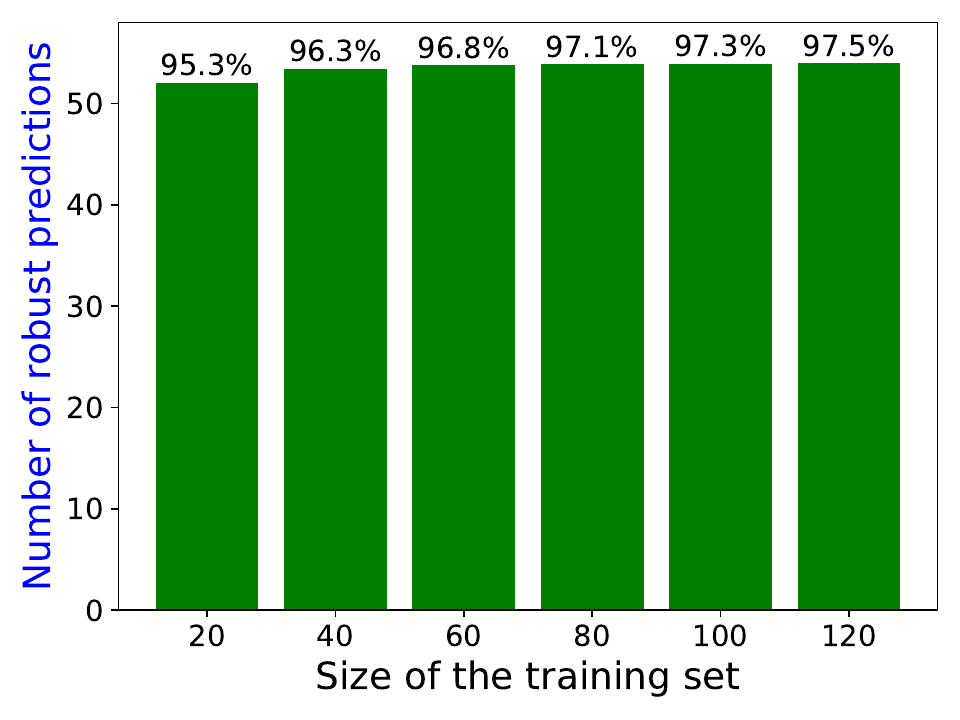}
    \caption{\footnotesize\emph{Former +} Acidity}\label{figure:winge_ac2}
    \end{subfigure}
    \caption{Global test performance (top) and number of reliable predictions with test performance for reliable predictions above each bar (bottom) for each model \emph{vs.} size of the training set on Wine dataset with Deep Ensemble.}\label{figure:evolution_wine}
\end{figure}

For each experiment, ALMA provides a protocol for assessing the reliability of a given prediction. If the EU is too high, additional labels must be queried; if the AU is too high, new modalities must be added. ALMA simply compares EU and AU values for a specific rejection, assuming that they are commensurate and equally important, which is a reasonable default assumption of the method. However, we can see in a concrete example that the functioning of this rejection scheme is indeed intuitive. 
Figure~\ref{fig:omr} shows all rejections at a training stage given only one modality (the online medical record of new incoming patients in the MIMIC-IV dataset). Such a simple modality is clearly insufficient for accurately predicting diseases. As a result, most test instances are rejected due to high aleatoric uncertainty, highlighting the need for more complex modalities.
Figure~\ref{figure:lab_reject} displays all rejections for a test set with two modalities, taking into account multiple training set sizes. As expected, the more training data the model has access to, the fewer instances are rejected due to epistemic uncertainty.

Figure~\ref{figure:alfa_bioscan} presents ALMA applied to two insects of the order \emph{Coleoptera}. Both test instances are analyzed in light of their respective ALMA process. The picture of the first \emph{Coleoptera} is initially queried as it represents the least costly form of observation. Starting from this point, the first model attempts to predict the corresponding class.
If the prediction is rejected, as long as EU remains above the threshold (EU $\geq$ AU), reflecting the model's lack of knowledge, active learning must be considered to add new observations to the dataset. 
Once the model achieves sufficiently low epistemic uncertainty, and if the prediction is not rejected, the problem is considered simple enough to make a prediction. However, this example indicates that the classification problem remained too complex when relying solely on images.
At this stage, a more expressive (and typically more expensive) representation space is required, so the model is enriched with geographical information.
Finally, by collecting the DNA data of the observation, the prediction becomes reliable, allowing the decision-making process to proceed.
For the second example, the first model can be stopped earlier. This time, with the enriched geographical information, AU was deemed sufficiently low. The prediction was therefore considered as reliable, even without the DNA sequence information. 

\subsection{ALMA improves predictive performance}

Let's now evaluate the ALMA algorithm in a more quantitative manner. As illustrated on the Wine dataset in Figure~\ref{figure:winge_ig1}, the performance of the model, trained on only the ignition study features, improves as the training set size increases. This process is repeated until all variables are included in the dataset, as depicted in Figure~\ref{figure:winge_ac1}, incorporating variables from the ignition, visual, chemical, and acidity studies. Two expected behaviors are observed: on the one hand, more labeled observations lead to better performance, and on the other hand, adding more variables also improves performance. At any given stage of the learning process, ALMA provides reliable predictions (or predictions deemed as such), as well as an acquisition protocol for non-reliable ones. Figures~\ref{figure:winge_ig2} to~\ref{figure:winge_ac2} display the number of reliable predictions alongside the model's performance on this ``reliable subset'' (represented by the value on top of the bars). As expected, the model's performance on the reliable subset is significantly higher than its overall performance at every learning stage. Intuitively, the number of reliable predictions increases with the inclusion of more variables (a more expensive model), ranging from 15-30 in Figure~\ref{figure:winge_ig2} to 50-54 in Figure~\ref{figure:winge_ac2}, out of 54 test instances. 

The number of reliable predictions can vary significantly depending on the model's certainty, and can sometimes be very low. On the BIOSCAN-5M dataset, the model trained on DNA sequences achieves $84.5\%$ accuracy with only $1k$ training instances, as depicted in Figure~\ref{figure:evolution}. Among the $4000$ test instances, $2353$ are considered reliable, achieving an accuracy of $97.8\%$ on this subset. With $20k$ training instances, all test instances are deemed reliable. 
In contrast, the model trained on images and enriched with geographical information achieves $59.0\%$ accuracy with $20k$ training instances. Among the $4000$ test instances, only $327$ are considered reliable, yet they achieve a $89.9\%$ accuracy: a significant improvement compared to the $59.0\%$ overall accuracy. 
For the first model (trained solely on images), between $79$ and $317$ predictions are considered reliable with fewer than $20k$ training instances. However, this number could be drastically increased by adding more observations in the training set (increasing costs), provided the model's uncertainty is of epistemic nature. 

The reliability of the predictions strongly depends on the quality of the underlying model. If the base learner performs poorly, the ALMA framework cannot compensate for this deficiency. This highlights the importance of selecting a robust model when implementing the ALMA approach. On MIMIC-IV, the classification task proved to be very challenging, with a maximum accuracy of around 50\% (compared to 13\% accuracy for a random classifier with 8 classes). However, the ALMA strategy still provides coherent results, even when the classifier's performance is not exceptional. These results are presented in the Appendix, following a similar analysis as in the previous experiments. Furthermore, Table~\ref{tab:mimic} shows the recall for each disease based on the number of features. This clearly highlights the irrelevance of certain modalities for specific diseases (such as microbiological cultures for detecting hypertension or alcoholic cirrhosis of the liver), while they may be essential for others (\emph{e.g.}, microbiological cultures for infection detection). ALMA thus enables not only more reliable predictions but also guides the way toward the optimal solution for handling cases that are less certain. 

\begin{figure}
    \centering
    \begin{subfigure}{0.29\linewidth}
    \includegraphics[width=\linewidth]{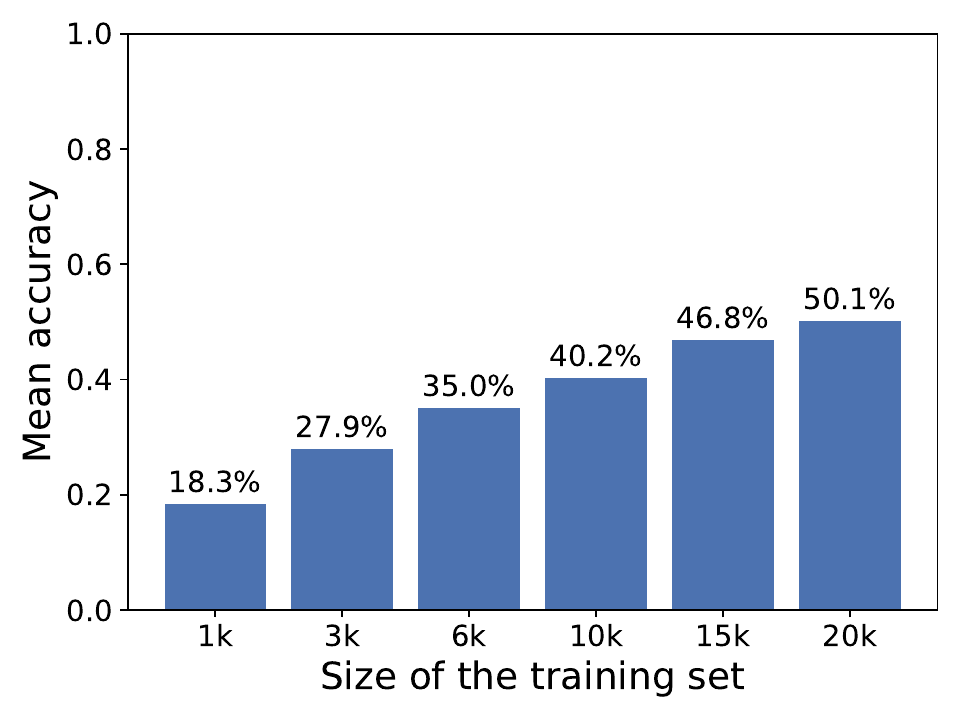}
    \caption{\footnotesize Image}
    \end{subfigure}
    \begin{subfigure}{0.29\linewidth}
    \includegraphics[width=\linewidth]{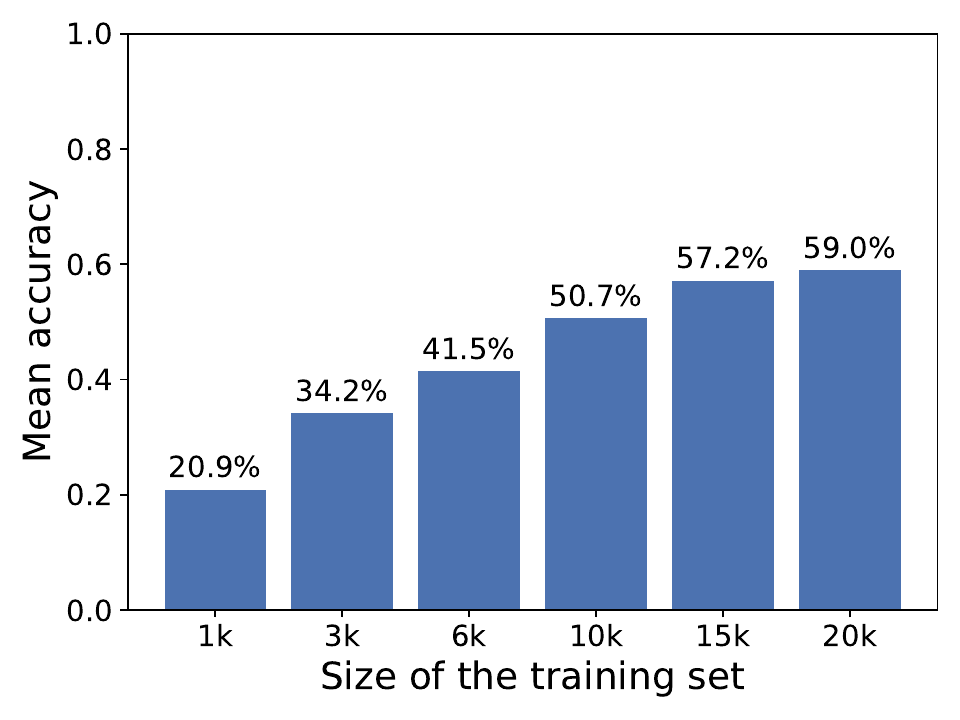}
    \caption{\footnotesize Image + Geographical}
    \end{subfigure}
    \begin{subfigure}{0.29\linewidth}
    \includegraphics[width=\linewidth]{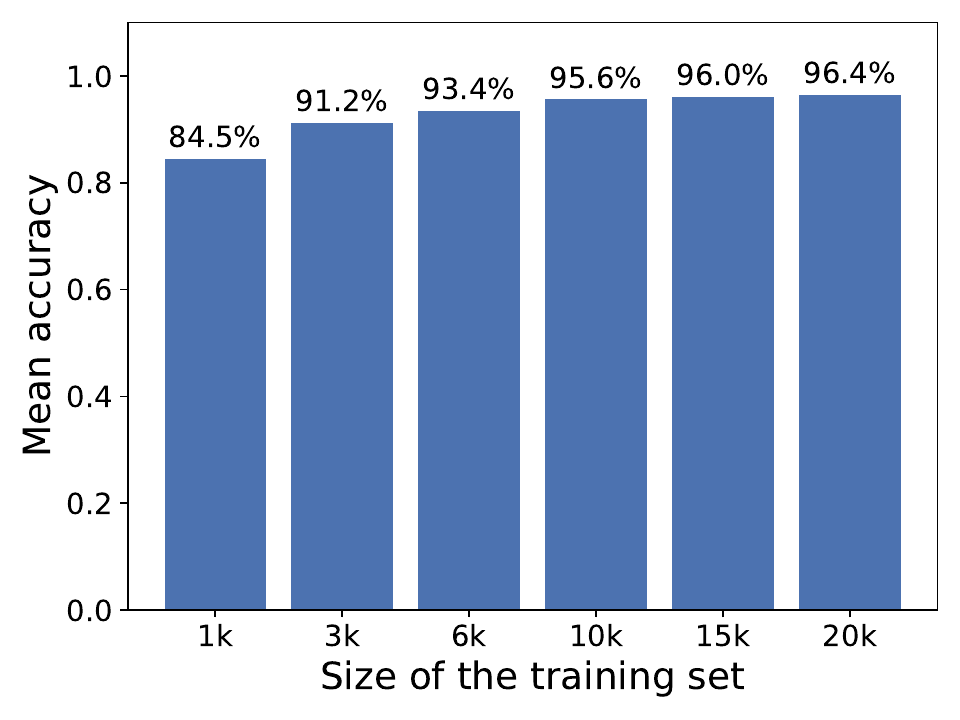}
    \caption{\footnotesize DNA Sequence}
    \end{subfigure}
    \begin{subfigure}{0.29\linewidth}
    \includegraphics[width=\linewidth]{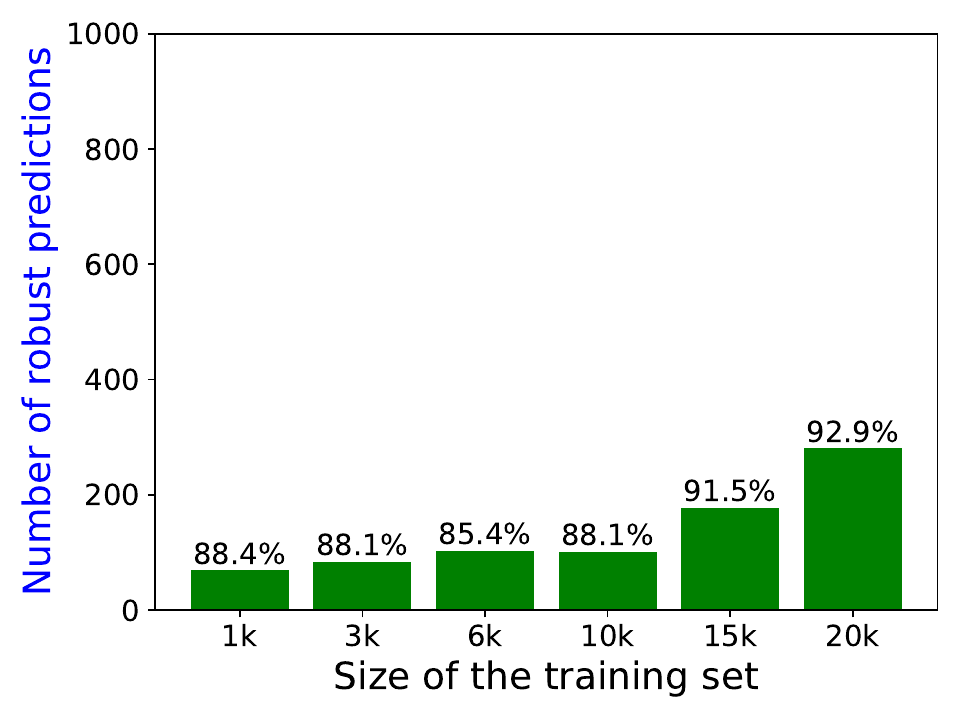}
    \caption{\footnotesize Image}\label{figure:robust1}
    \end{subfigure}
    \begin{subfigure}{0.29\linewidth}
    \includegraphics[width=\linewidth]{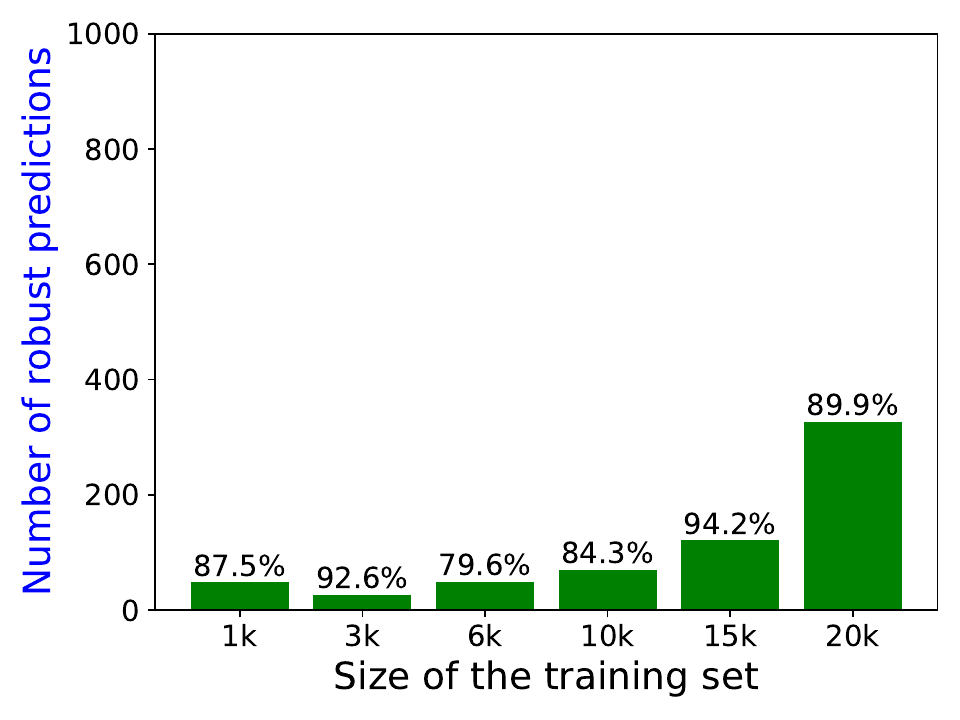}
    \caption{\footnotesize Image + Geographical}
    \end{subfigure}
    \begin{subfigure}{0.29\linewidth}
    \includegraphics[width=\linewidth]{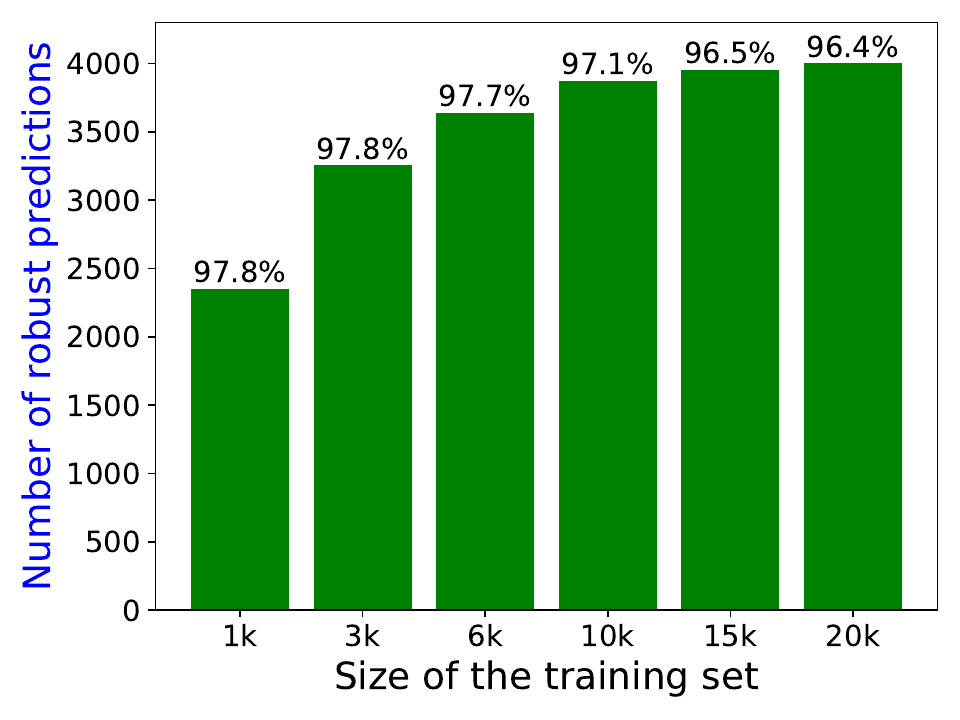}
    \caption{\footnotesize DNA Sequence}
    \end{subfigure}
    \caption{Global test performance (top) and number of reliable predictions with test performance for reliable predictions above each bar (bottom) for each model \emph{vs.} size of the training set on BIOSCAN-5M dataset with Deep Ensemble.}\label{figure:evolution}
\end{figure}

\subsection{Comparison of uncertainty disentanglement methods}

A summary of benchmarking experiments for the eight uncertainty disentanglement methods is provided in Appendix~\ref{app:summary}. The findings reveal that the proposed strategy is effective across different scenarios, though the number of reliable predictions depends heavily on the underlying model and its performance. When the base model is weak, the strategy cannot yield satisfactory results. Additionally, the experiments demonstrate that ALMA's reliability varies with the complexity of the model, and it works best when the model has sufficient training data and complexity. Furthermore, the rejection strategy, which compares epistemic uncertainty and aleatoric uncertainty, is crucial for guiding the model toward more reliable predictions.

The overall results, including the number of rejected predictions and the model performance across various uncertainty disentanglement methods, are summarized in the appendix in Tables~\ref{tab:bioscan_reject} and \ref{tab:mimic_reject}. 
These tables provide a model comparison for all methods, offering insights into the effectiveness of ALMA across different model types and dataset characteristics. The difference between the entropy-based and variance-based decompositions does not appear to have a particularly strong impact. The centroid-based approach is the only one that performs very poorly with our ALMA method, but this is due to the base learner rather than the uncertainty disentanglement. It struggled with more complex data, although it performed comparably to the other methods on the simplest dataset. Deep Ensemble appears to be overall a good fit, whether using entropy-based or variance-based decomposition. Additionally, it offers strong overall base performance. The number of rejections based on epistemic uncertainty consistently decreases during active learning, while the number of rejections based on aleatoric uncertainty decreases as more modalities are added. Bayesian Deep Learning performed similarly, with slightly worse results, followed by the other methods.

\section{Conclusions}\label{section:conclusion}

In conclusion, this paper introduces an active learning strategy that integrates multi-modal feature acquisition to enhance the reliability of machine learning predictions. By employing a reject criterion with statistical guarantees, grounded in the assessment of both epistemic and aleatoric uncertainties, the proposed approach not only delivers reliable predictions but also provides a systematic protocol to address uncertainties when predictions lack confidence. This dual-directional sampling strategy (focusing on acquiring additional observations for epistemic uncertainty and new modalities for aleatoric uncertainty) ensures a flexible and cost-effective framework for improving model performance.

The experimental evaluation across three datasets, including two multi-modal datasets combining images, geographical data, DNA sequences, patient information, laboratory tests, and microbiological cultures, demonstrates the effectiveness of the strategy. By leveraging various state-of-the-art uncertainty quantification methods (whether entropy-based, variance-based, or distance-based) the framework showcases its adaptability in mitigating uncertainties while accounting for the associated costs. This work highlights the potential of multi-modal data acquisition in active learning as a reliable and resource-efficient paradigm for uncertainty-aware machine learning.


\bibliography{ref/main}
\bibliographystyle{style}

\newpage
\appendix

\section{Uncertainty disentanglement}\label{app:methods}

\subsection{Entropy-based decomposition}

For entropy-based uncertainty estimation, we used the entropy decomposition method described in~\citep{Shaker2020}. Here, the posterior distribution is approximated by a finite ensemble of estimators. The AU can be obtained as follows:
\begin{equation}
    AU(\textbf{x}) = - \frac{1}{M}\sum_{i=1}^M\sum_{y\in\mathcal{Y}}p(y|h_i, \textbf{x})\log_2p(y|h_i, \textbf{x}),
\end{equation}
with $M$ the number of estimators.
Total Uncertainty (TU) can be approximated in terms of Shannon's entropy of the predictive posterior distribution as follows:
\begin{equation}
    TU(\textbf{x}) = -\sum_{y\in\mathcal{Y}}\Bigg\{\left(\frac{1}{M}\sum_{i=1}^M p(y|h_i, \textbf{x})\right)\log_2\left(\frac{1}{M}\sum_{i=1}^M p(y|h_i, \textbf{x})\right)\Bigg\}.
\end{equation}
Finally, EU can be derived through additive decomposition as:
\begin{equation}
    EU(\textbf{x}) = TU(\textbf{x}) - AU(\textbf{x}).
\end{equation}

\subsection{Variance-based decomposition}

For variance-based uncertainty estimation, we investigated the label-wise variance decomposition approach proposed in~\citep{Sale2024-b}. TU is defined for each class as the variance of the outcomes, and global uncertainty is obtained by summing over all classes as follows:
\begin{equation}
    TU(\textbf{x}) = \sum_{y\in\mathcal{Y}}\text{Var}(Y_y),
\end{equation}
where $Y_y$ are binary outcomes according to a $K$-dimensional random vector indicating the presence or absence of a particular label $y\in\mathcal{Y}$.
AU is defined as the expected conditional variance, as follows:
\begin{equation}
    AU(\textbf{x}) = \sum_{y\in\mathcal{Y}}\mathbb{E}\left[\Theta_y(1-\Theta_y) \right],
\end{equation}
with $\Theta_y = P(Y_y = 1)$ associated to a random vector distributed according to a second-order distribution; whereas EU corresponds to the expected reduction in squared-error loss, defined as follows: 
\begin{equation}
    EU(\textbf{x}) = \sum_{y\in\mathcal{Y}}\text{Var}(\Theta_y) . 
\end{equation}

\subsection{Deep Ensemble}

As an alternative to Bayesian Deep Learning, the authors in~\citep{Lakshminarayanan2017} proposed constructing an ensemble of deep neural networks for uncertainty quantification. For each experiment, we trained five neural networks, each time randomizing the initial weights. To ensure diversity, we also shuffled the order of the training set, as suggested by the authors.
An ensemble of $M = 5$ learners is then obtained. EU and AU are subsequently computed according to the entropy-based decomposition and the variance-based decomposition described above.
For the Wine dataset, we used a fully connected multi-layer perceptron with two hidden layers of dimensions (100, 20) and the Adam solver (with a maximum of 2000 iterations). For the other experiments, when not specified, we used two hidden layers of dimensions (200, 50).

\subsection{Bayesian Deep Learning}

Bayesian Deep Learning can also be used to disentangle aleatoric and epistemic uncertainty~\citep{kendall2017} by modeling distributions instead of simple weights.
First, we construct a fully connected neural network with architecture (100, 20) for the Wine dataset, and (200, 50) for other experiments when not explicitly specified, using the Adam solver with a maximum of 2000 iterations.
In a post-training step, we use Laplace approximations to construct the Bayesian network with Laplace Redux~\citep{daxberger2022} (default parameters). We then sample 100 predicted probabilities from the posterior distribution for each test instance, which can be considered as an ensemble of $M = 100$ estimators.
EU and AU are subsequently computed according to the entropy-based decomposition and the variance-based decomposition described above.

\subsection{Random Forest}

The experiments are also conducted using the Random Forest method presented in~\citep{Shaker2020} to estimate epistemic and aleatoric uncertainties.
For the Wine dataset, we trained a random forest with 100 estimators and a maximum depth of 4. All other parameters are set to the default values of the scikit-learn library~\citep{sklearn2011}. An ensemble of $M = 100$ estimators in the obtained.
To compute epistemic and aleatoric uncertainties, we used the entropy-based decomposition and the variance-based decomposition described above.

\subsection{Centroid-based decomposition}

For distance-based methods, we examined the centroid-based approach presented in~\citep{van-amersfoort20a}. To compute the model's uncertainty, the authors proposed measuring the distance from the incoming observation to the class centroids using a radial basis function:
\begin{equation}
    U_y(\textbf{x}, \textbf{e}_y) = \exp\left[-\frac{\frac{1}{n}||\textbf{W}_y \hat{h}(\textbf{x}) - \textbf{e}_y||^2_2}{2\sigma^2}\right],
\end{equation}
with $\hat{h}$ the model, $\textbf{e}_y$ the centroid associated with class $y$ and $\textbf{W}_y$ a weight matrix of
size $n$ (centroid size) by $d$ (feature extractor output size) in the case of a deep architecture. The length scale $\sigma$ is an hyper-parameter of the method set to 1.
The epistemic certainty $\mathcal{C}$ associated with instance $\textbf{x}$ is then obtained by measuring the distance from the incoming observation to the nearest class centroid:
\begin{equation}
    \mathcal{C}(\textbf{x}) = \max_{y\in\mathcal{Y}}U_y(\textbf{x}, \textbf{e}_y).
\end{equation}
One can then obtain $EU(\textbf{x}) = 1 / \mathcal{C}(\textbf{x})$ while $AU$ is computed based on the relative distances between all clusters (entropy of the obtained softmax distribution).

\subsection{Neighbors-based decomposition}

To locally estimate the epistemic uncertainty of the model, the authors in~\citep{Hoarau2024-ml} propose using Evidential K-NN~\citep{denoeux1995}, which is based on the mathematical theory of evidence~\citep{Dempster1967,shafer1976}. With the $K$ nearest neighbors of $\textbf{x}$ according to a distance metric $d$, the model outputs a belief function $m$~\citep{shafer1976} defined over $2^\mathcal{Y}$ as the prediction for $\textbf{x}$, such that $m(A) \in [0,1]$ and $\sum_{A \subseteq \mathcal{Y}}m(A)=1$. Given the $K$ nearest neighbors of $\textbf{x}$, one can assign a mass function to each neighbors $x_j, 1 \leq j \leq K$ of class $y^j \in \mathcal{Y}$ discounted according to the distance $d_j$ from $\textbf{x}$:
\begin{equation}
    \begin{split}
        &m_{j}(y^j) = \alpha e^{-\gamma {d_{j}}^2},\\
        &m_{j}(\mathcal{Y}) \;= 1 - m_{j}(y^j),
    \end{split}
\end{equation}
where $\alpha$ and $\gamma$ are parameters of the method. The parameter $\alpha$ is called the discounting coefficient, and is usually set to $0.8$. The authors suggest to set $\gamma$ as the invert mean distance between all points in the training set.
All $K$ mass functions respectively attributed to the neighbors of $\textbf{x}$ are then combined into $m$ using Dempster's rule of combination~\citep{Dempster1967}.
The epistemic uncertainty is then obtained from the Non-Specificity~\citep{dubois1987} as follows:
\begin{equation}\label{eq:nonspe}
    EU(\textbf{x}) = \sum_{A\subseteq \mathcal{Y}} m(A)\text{log}_2(|A|)\,,
\end{equation}
and the aleatoric uncertainty is obtained from the Discord equation~\citep{klir1990uncertainty}, as follows:
\begin{equation}\label{eq:discord}
    AU(\textbf{x}) = - \sum_{A\subseteq \mathcal{Y}} m(A)\text{log}_2({\mathrm{BetP}(A)})\,,
\end{equation}
where $\mathrm{BetP}$ is the probability measure associated with the following probability mass:
\begin{equation}
    \mathrm{BetP}(\{y\}) = \sum_{A\subseteq \mathcal{Y}, \;\;y\in A}\frac{m(A)}{|A|}\,.
\end{equation}
Aleatoric uncertainty is therefore higher in dense, conflicting regions, while epistemic uncertainty is higher in low-density regions. Note that $\mathrm{BetP}$ actually corresponds to applying a Laplace principle of indifference for each subset having positive mass $m(A)$. For the experiments, we used $K = 7$ neighbors for the Wine dataset and $K = 10$ otherwise.

\section{Application to Wine dataset}\label{app:wine}

The dataset~\citep{Dua2019} consists of wines grown in the same region of Italy but derived from three different cultivars. Each sample is characterized by 13 variables, which we have grouped based on an assigned fictitious cost, defined as follows:
\begin{description}
    \item[Ignition study:] Ash, Alcalinity of ash, Total phenols, Non-flavenoid phenols 
    \item[Visual study:] Hue, Color intensity 
    \item[Chemical study:] Proanthocyanins, Flavanoids, Magnesium, od315 of diluted wines
    \item[Acidity study:] Proline, Malic acid, Alcohol
\end{description}

For the experiments, we randomly split the training and test sets using a 70/30 ratio, run the experiments, and repeat the process 100 times to estimate the mean performance.

Four different models are trained on the dataset according to the train/test split. The first model is trained solely on the variables from the ignition study. The second model is trained on the same variables enriched with those from the visual study, and so on, until the fourth model incorporates all 13 variables.

The results are presented for all the studied models in Figures~\ref{figure:evolution_wine},~\ref{figure:wine_edl},~\ref{figure:wine_rf},~\ref{figure:wine_cent} and~\ref{figure:wine_eknn}.

\begin{figure}
    \centering
    \begin{subfigure}{0.24\linewidth}
    \includegraphics[width=\linewidth]{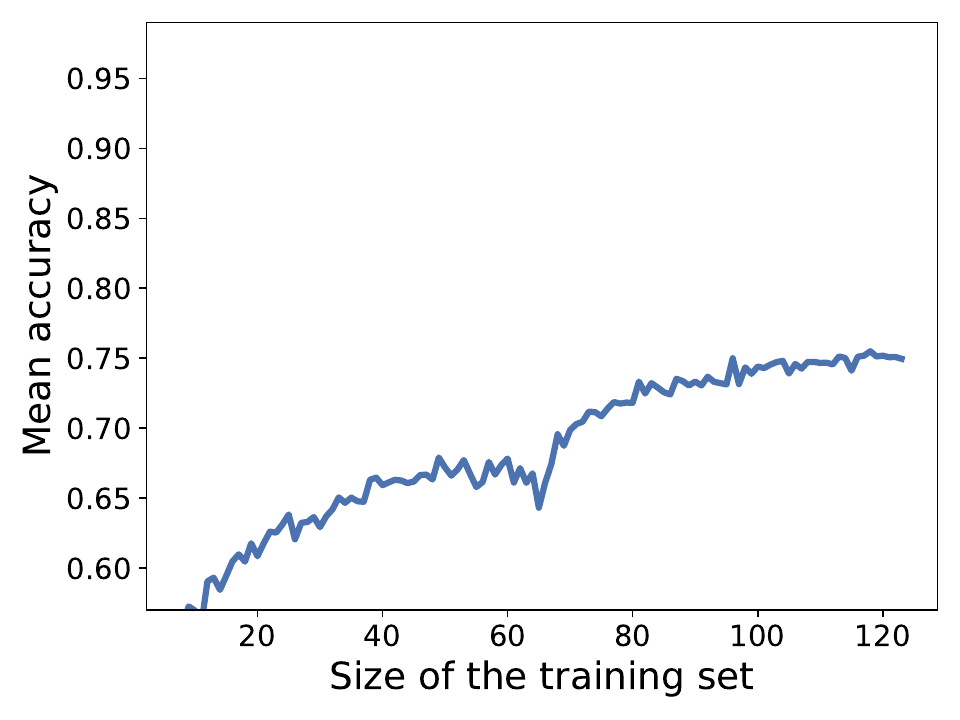}
    \end{subfigure}
    \begin{subfigure}{0.24\linewidth}
    \includegraphics[width=\linewidth]{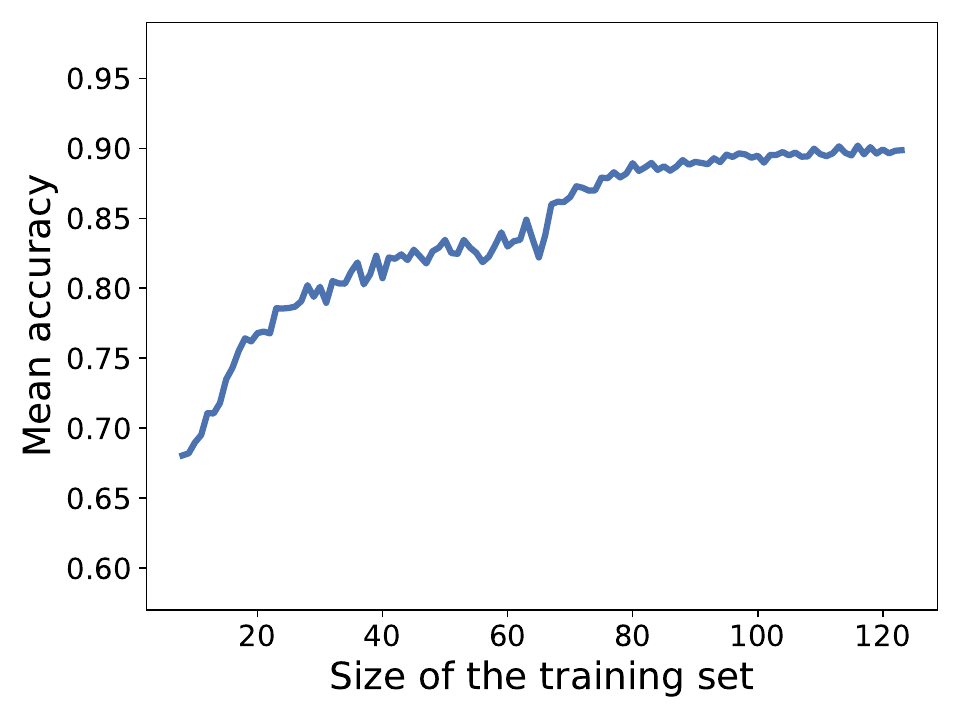}
    \end{subfigure}
    \begin{subfigure}{0.24\linewidth}
    \includegraphics[width=\linewidth]{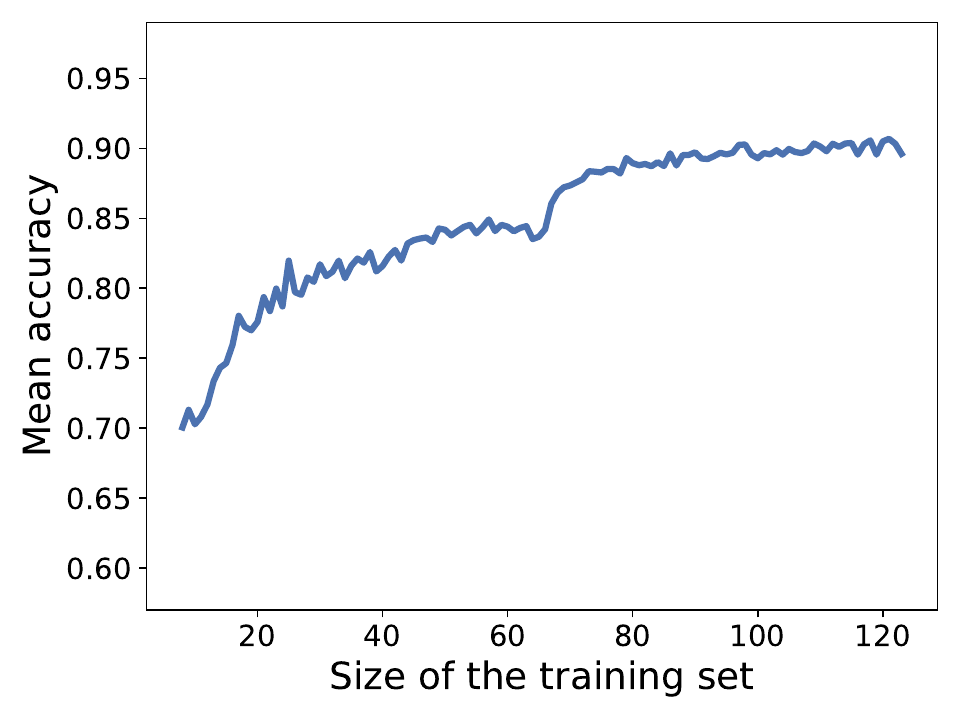}
    \end{subfigure}
    \begin{subfigure}{0.24\linewidth}
    \includegraphics[width=\linewidth]{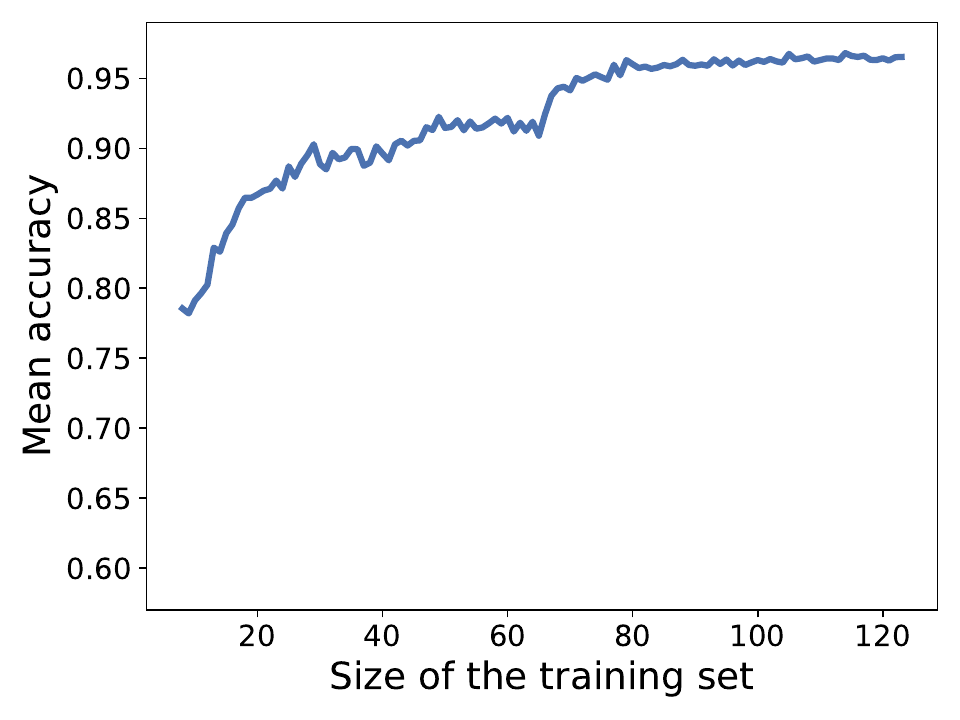}
    \end{subfigure}
    \begin{subfigure}{0.24\linewidth}
    \includegraphics[width=\linewidth]{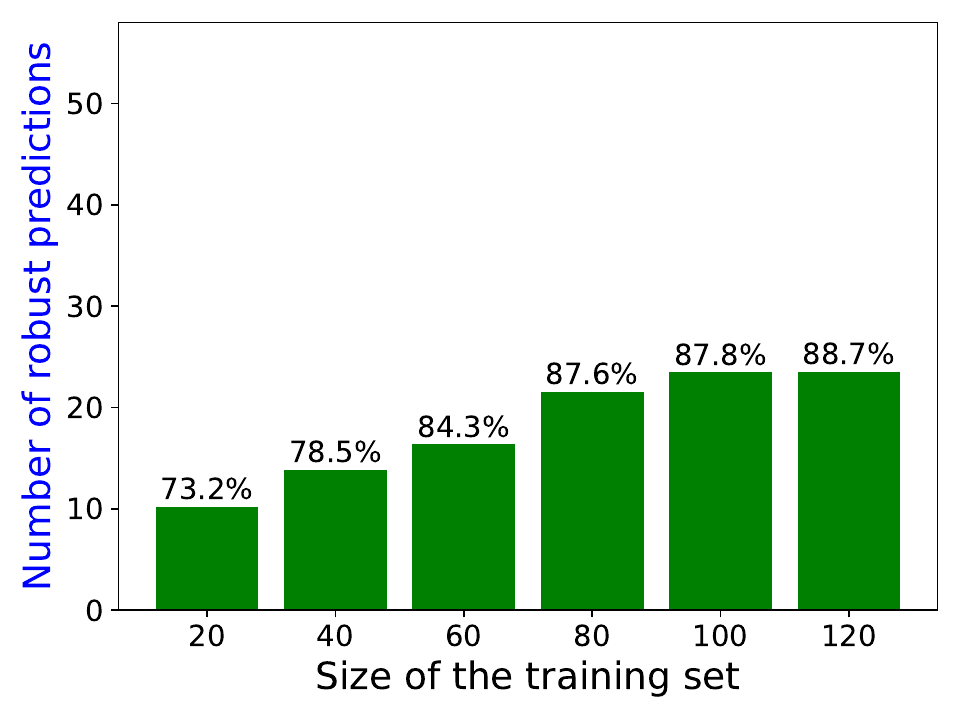}
    \caption{\footnotesize Ignition}
    \end{subfigure}
    \begin{subfigure}{0.24\linewidth}
    \includegraphics[width=\linewidth]{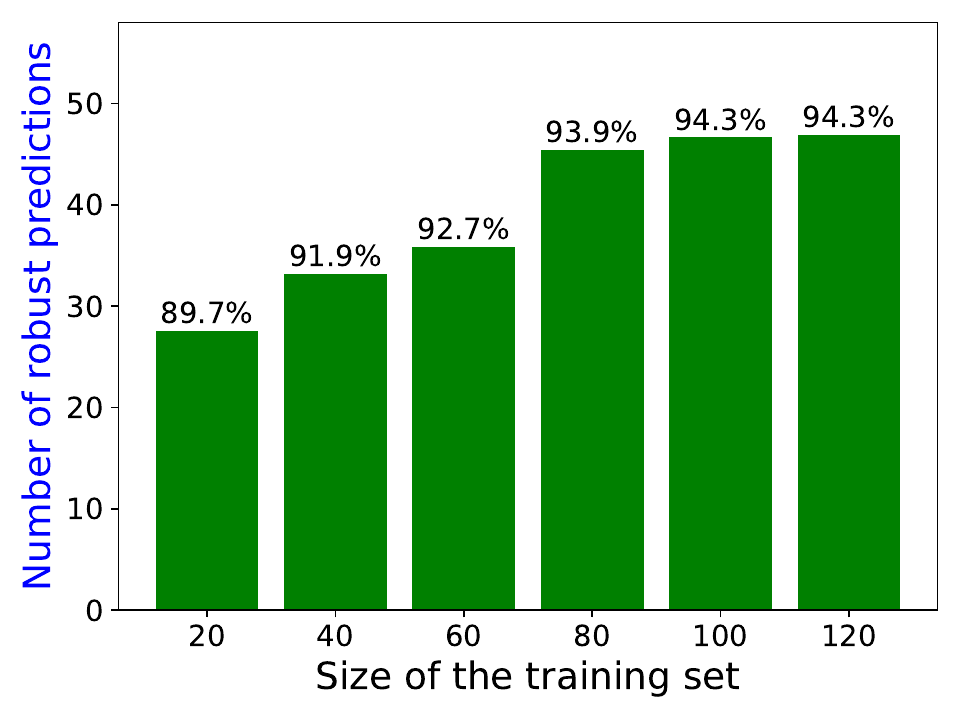}
    \caption{\footnotesize\emph{Former +} Visual}
    \end{subfigure}
    \begin{subfigure}{0.24\linewidth}
    \includegraphics[width=\linewidth]{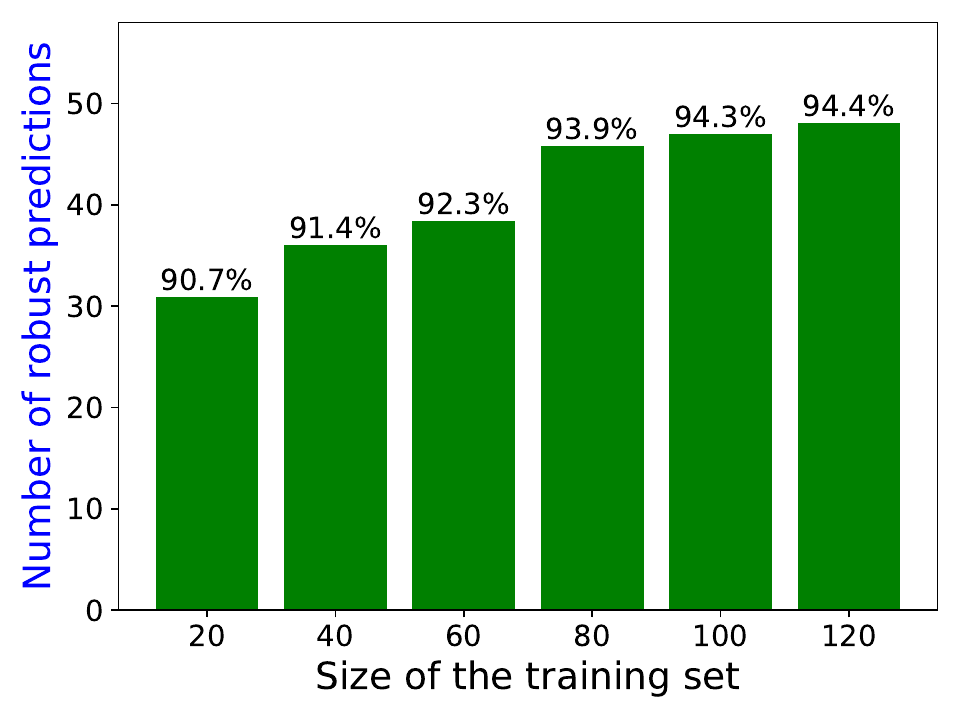}
    \caption{\footnotesize\emph{Former +} Chemical}
    \end{subfigure}
    \begin{subfigure}{0.24\linewidth}
    \includegraphics[width=\linewidth]{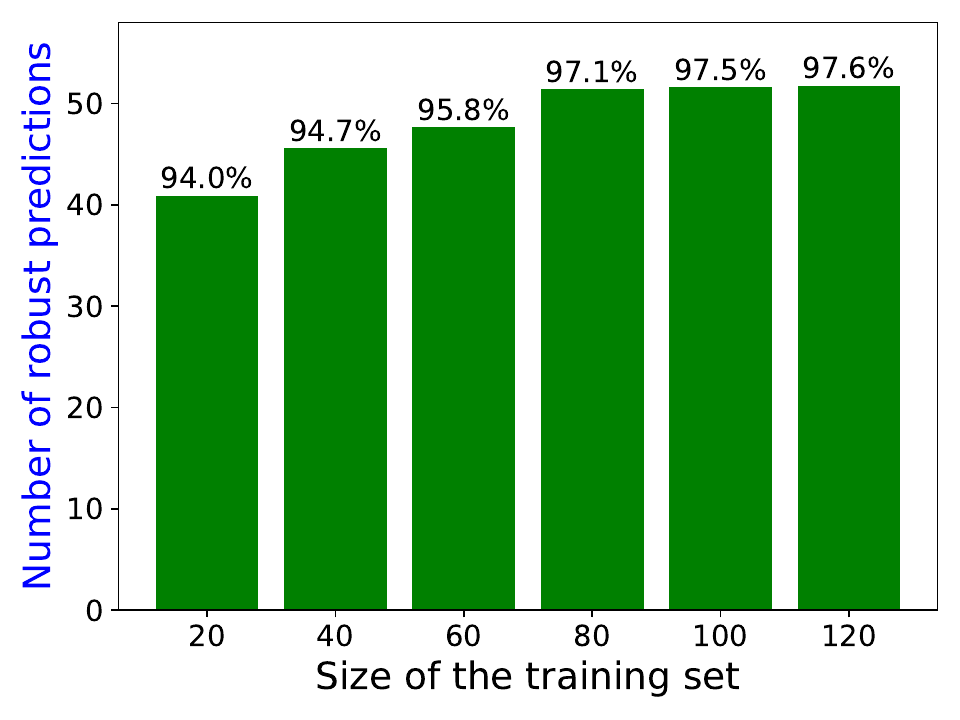}
    \caption{\footnotesize\emph{Former +} Acidity}
    \end{subfigure}
    \caption{Global test performance (top) and number of reliable predictions with test performance for reliable predictions above each bar (bottom) for each model \emph{vs.} size of the training set on Wine dataset with Bayesian Deep Learning.}\label{figure:wine_edl}
\end{figure}

\begin{figure}
    \centering
    \begin{subfigure}{0.24\linewidth}
    \includegraphics[width=\linewidth]{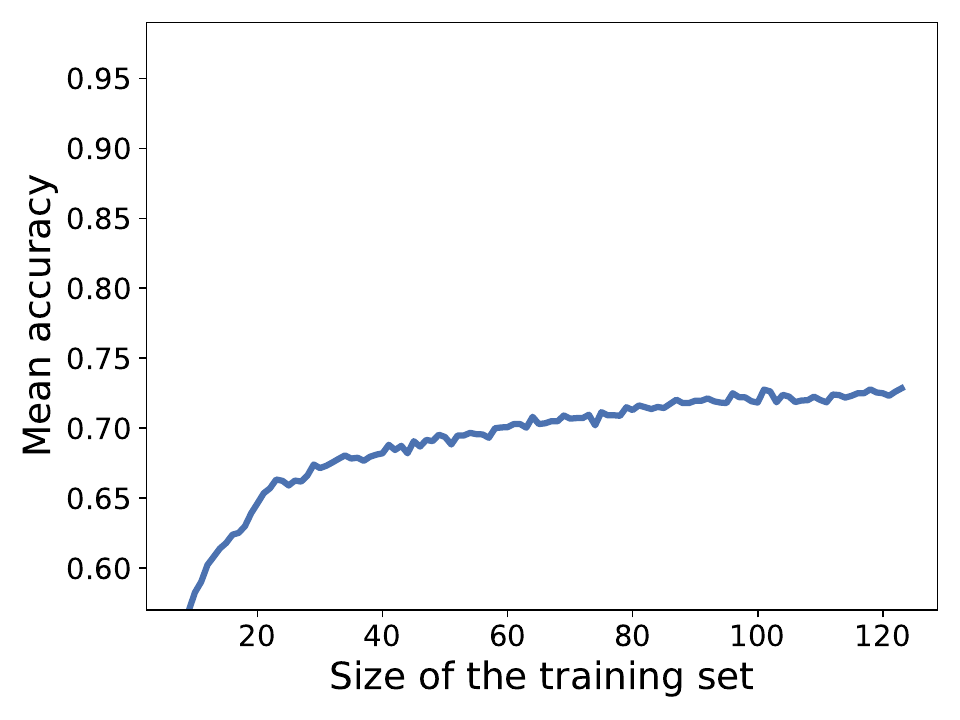}
    \end{subfigure}
    \begin{subfigure}{0.24\linewidth}
    \includegraphics[width=\linewidth]{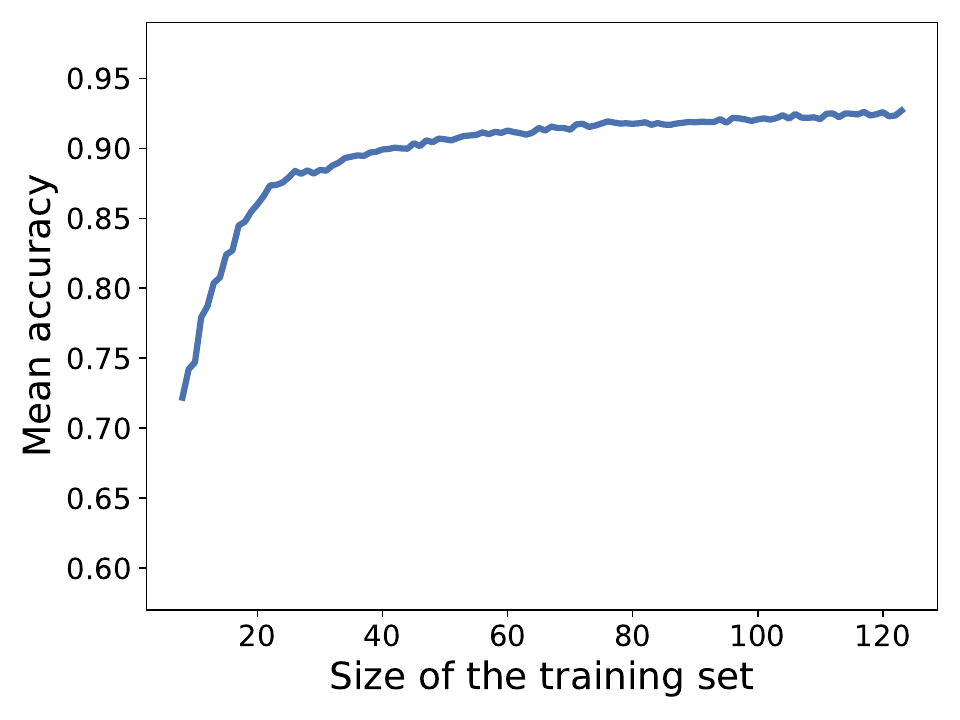}
    \end{subfigure}
    \begin{subfigure}{0.24\linewidth}
    \includegraphics[width=\linewidth]{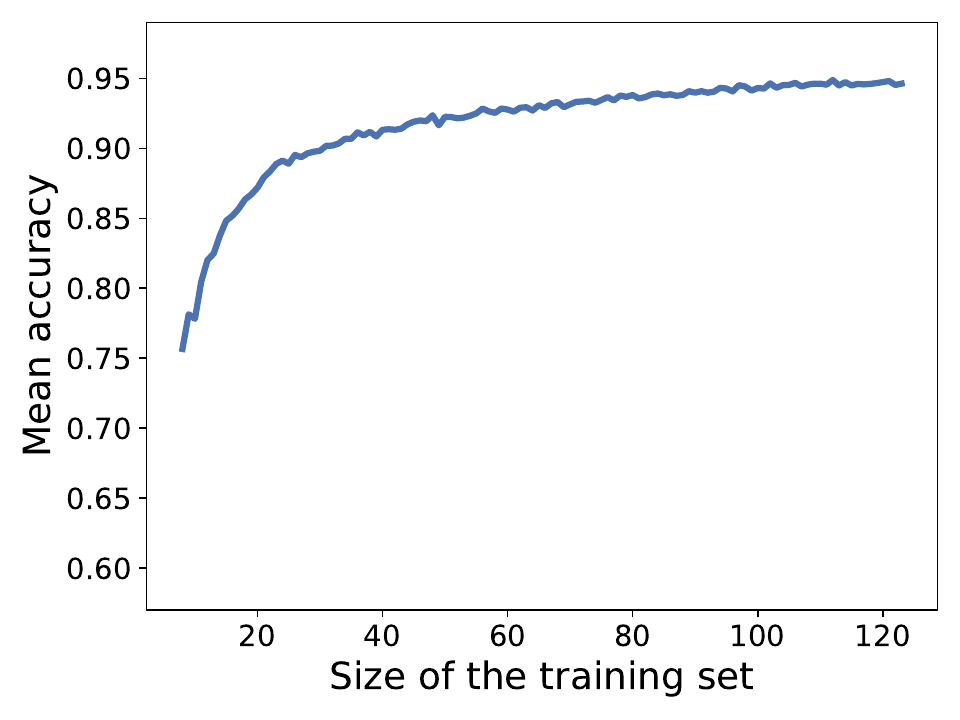}
    \end{subfigure}
    \begin{subfigure}{0.24\linewidth}
    \includegraphics[width=\linewidth]{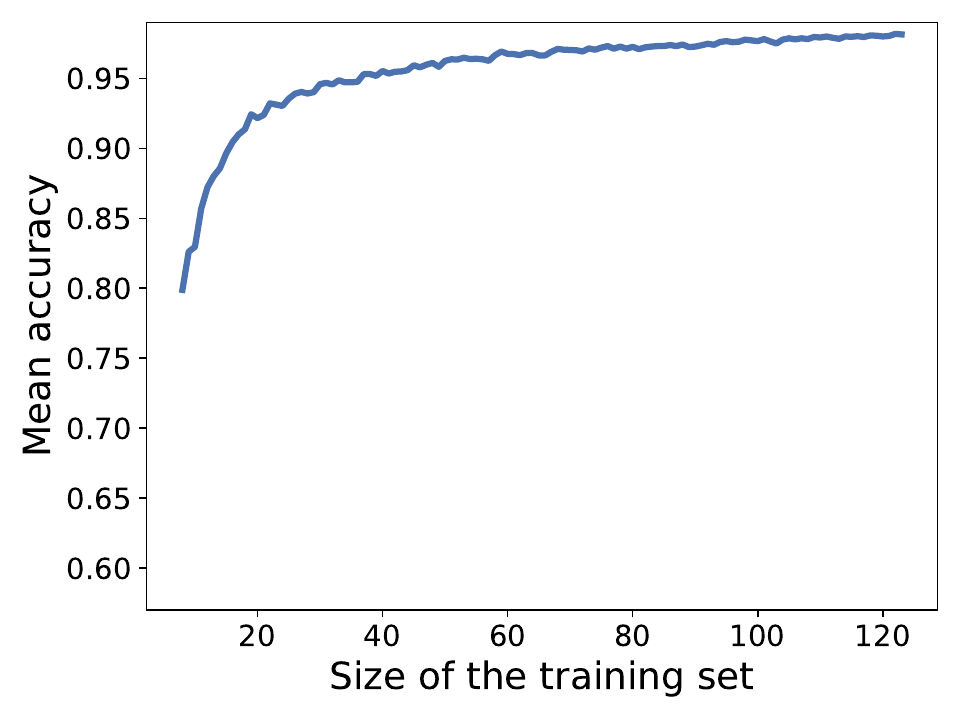}
    \end{subfigure}
    \begin{subfigure}{0.24\linewidth}
    \includegraphics[width=\linewidth]{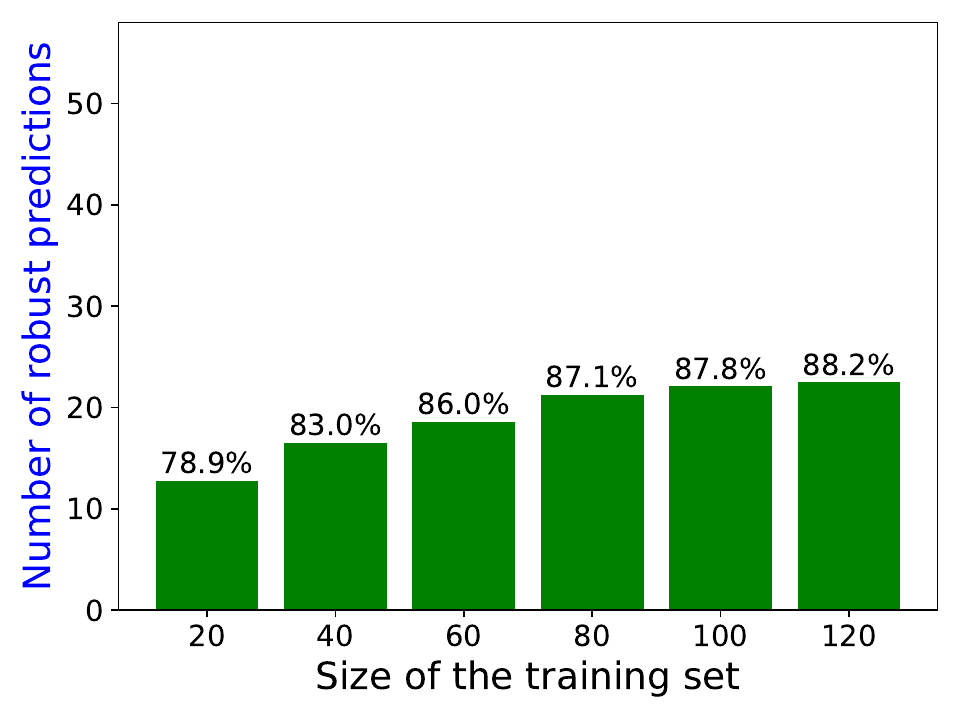}
    \caption{\footnotesize Ignition}
    \end{subfigure}
    \begin{subfigure}{0.24\linewidth}
    \includegraphics[width=\linewidth]{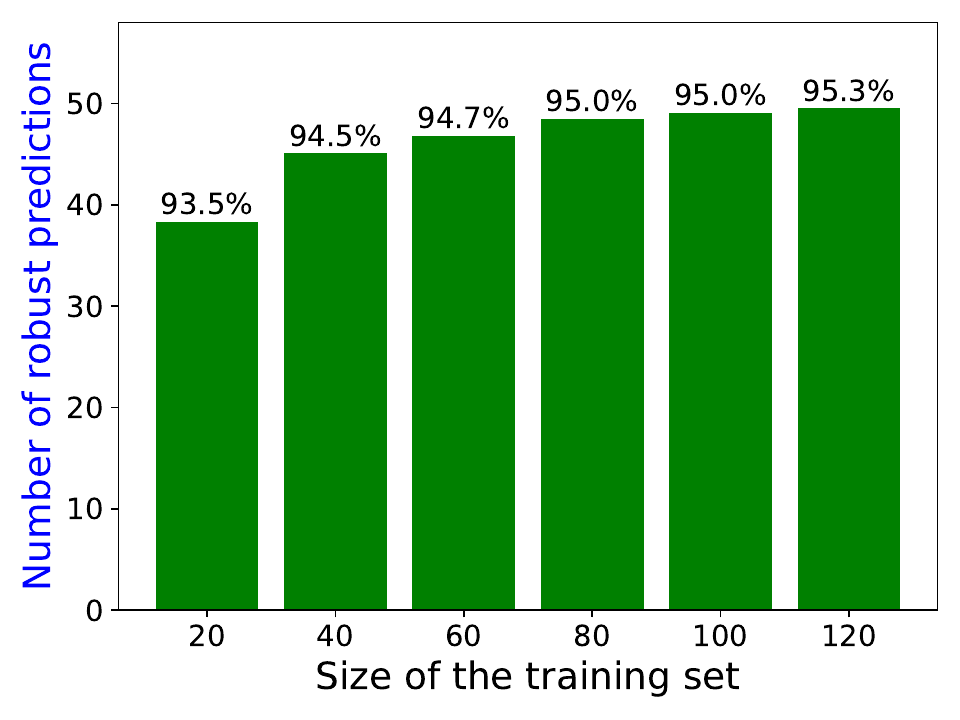}
    \caption{\footnotesize \emph{Former +} Visual}
    \end{subfigure}
    \begin{subfigure}{0.24\linewidth}
    \includegraphics[width=\linewidth]{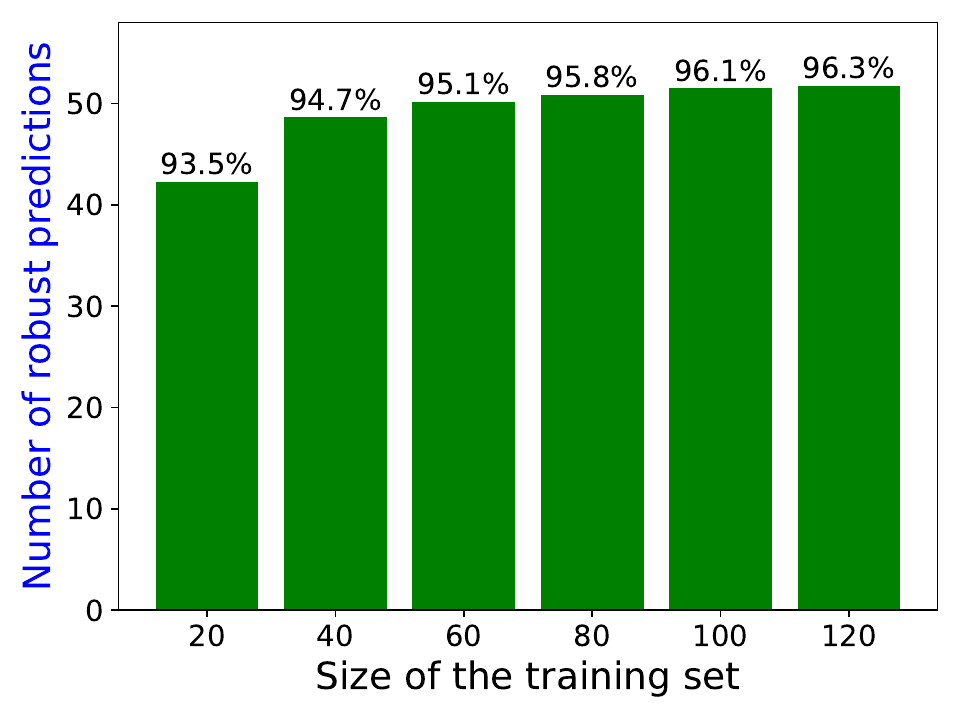}
    \caption{\footnotesize\emph{Former +} Chemical}
    \end{subfigure}
    \begin{subfigure}{0.24\linewidth}
    \includegraphics[width=\linewidth]{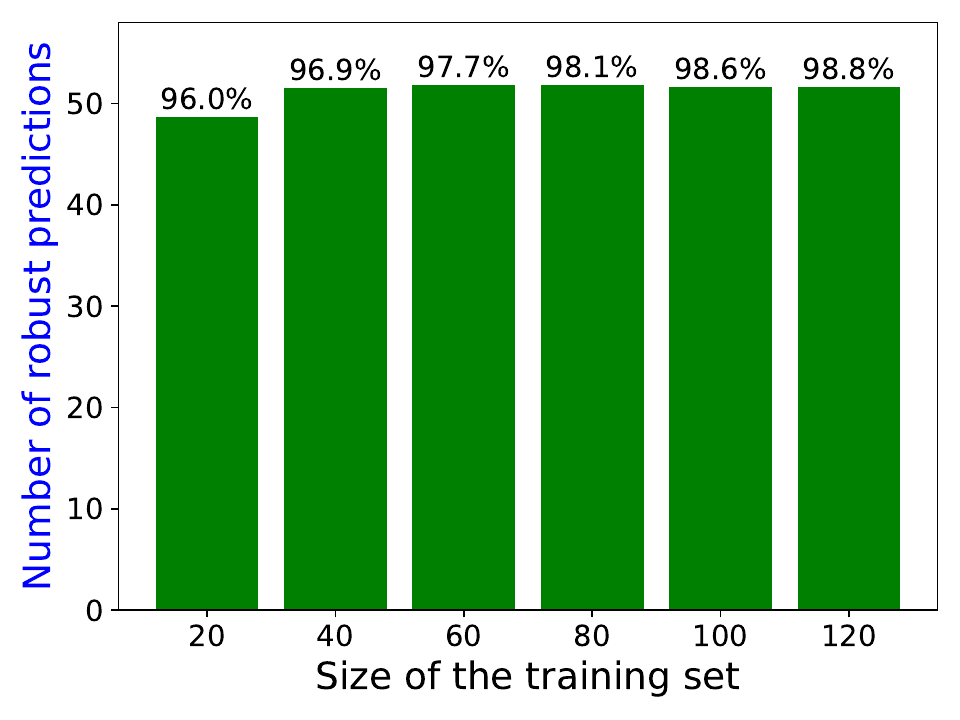}
    \caption{\footnotesize\emph{Former +} Acidity}
    \end{subfigure}
    \caption{Global test performance (top) and number of reliable predictions with test performance for reliable predictions above each bar (bottom) for each model \emph{vs.} size of the training set on Wine dataset with Random Forest.}\label{figure:wine_rf}
\end{figure}

\begin{figure}
    \centering
    \begin{subfigure}{0.24\linewidth}
    \includegraphics[width=\linewidth]{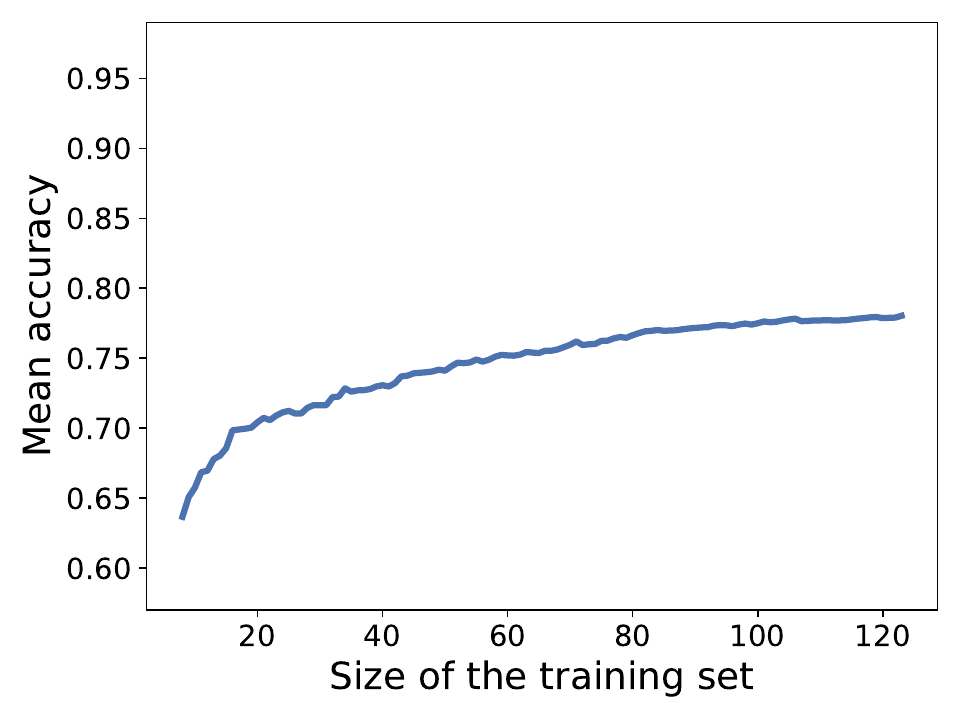}
    \end{subfigure}
    \begin{subfigure}{0.24\linewidth}
    \includegraphics[width=\linewidth]{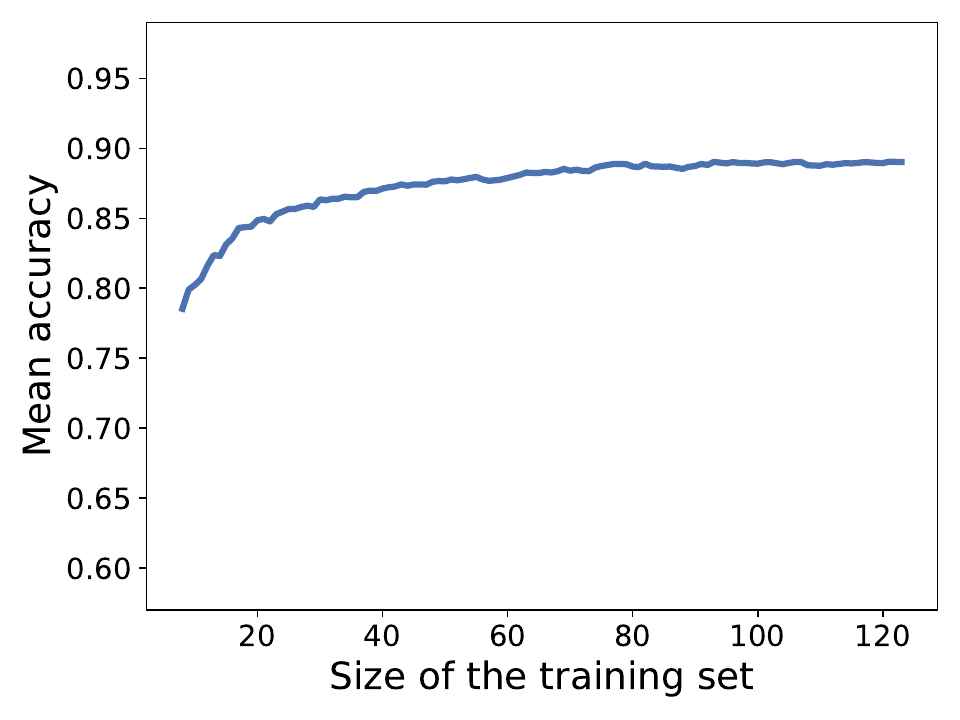}
    \end{subfigure}
    \begin{subfigure}{0.24\linewidth}
    \includegraphics[width=\linewidth]{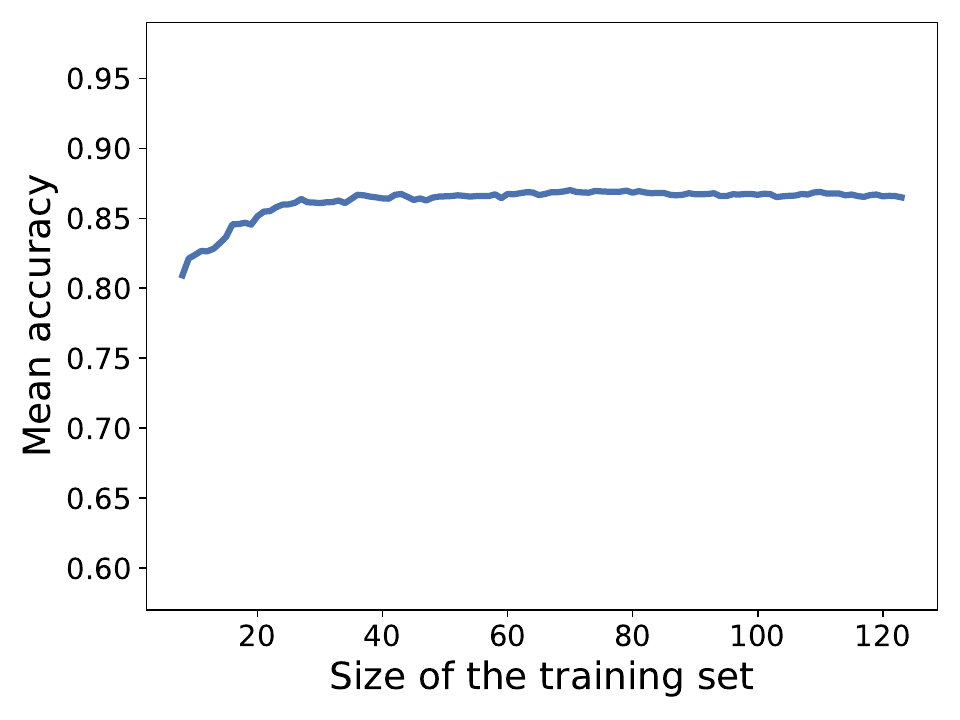}
    \end{subfigure}
    \begin{subfigure}{0.24\linewidth}
    \includegraphics[width=\linewidth]{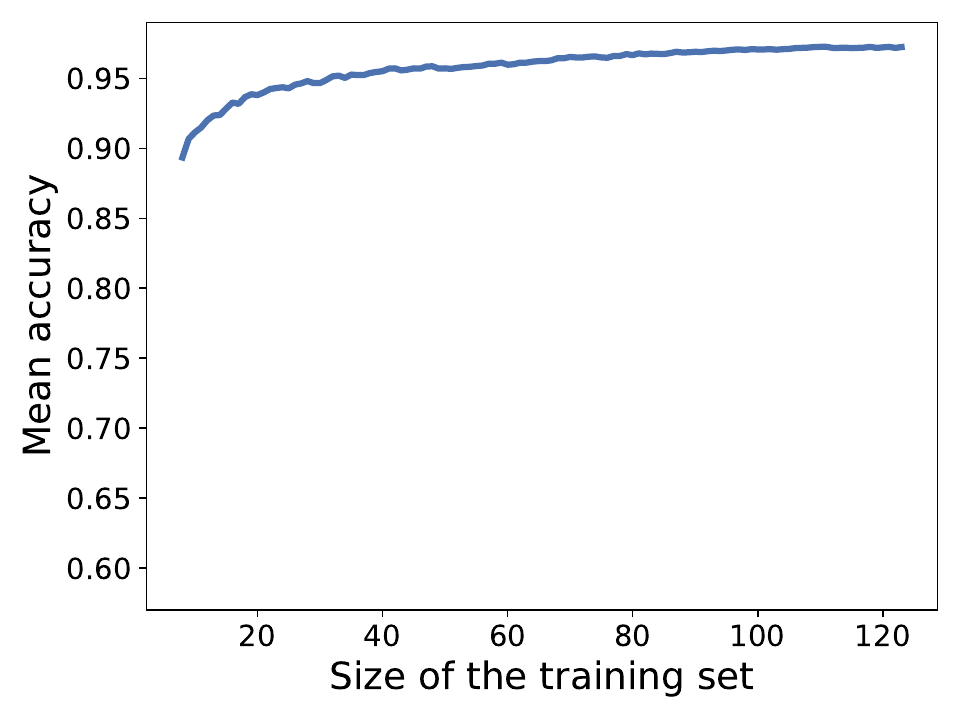}
    \end{subfigure}
    \begin{subfigure}{0.24\linewidth}
    \includegraphics[width=\linewidth]{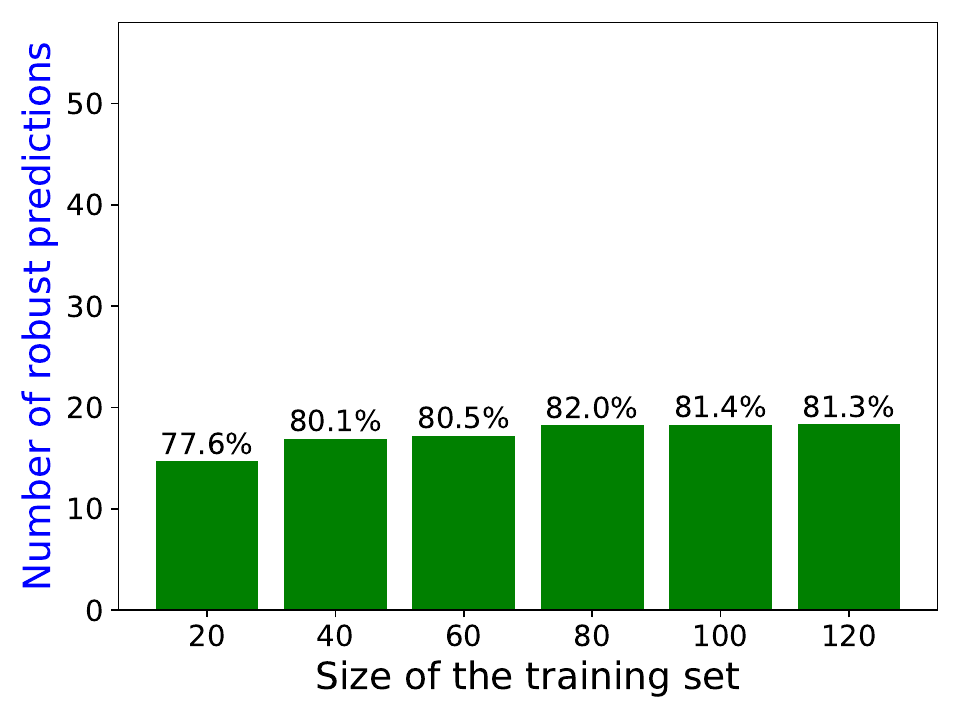}
    \caption{\footnotesize Ignition}
    \end{subfigure}
    \begin{subfigure}{0.24\linewidth}
    \includegraphics[width=\linewidth]{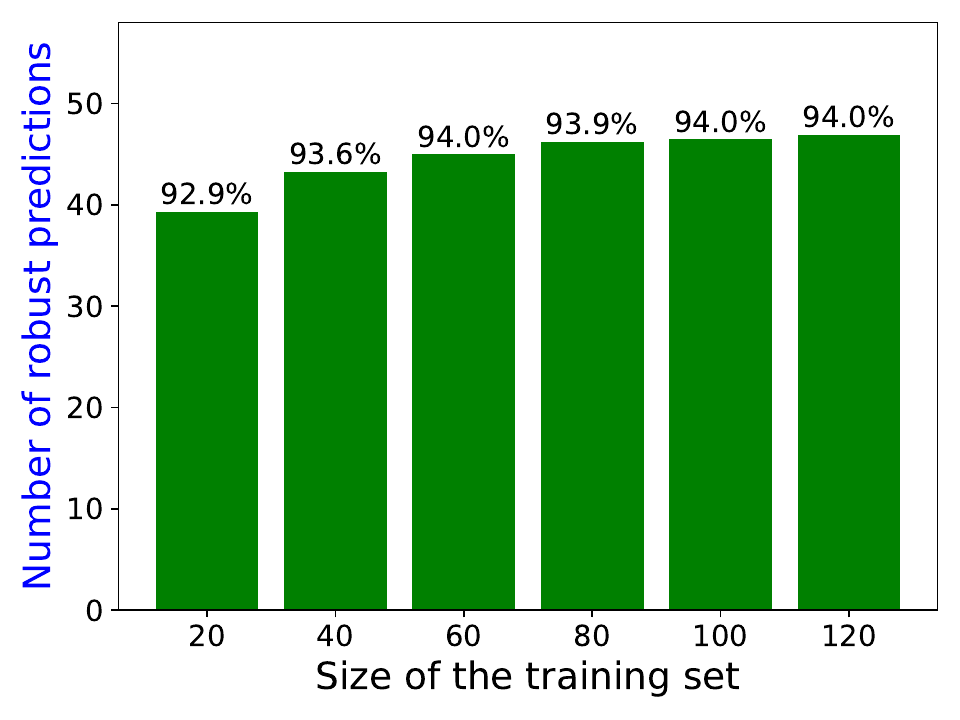}
    \caption{\footnotesize \emph{Former +} Visual}
    \end{subfigure}
    \begin{subfigure}{0.24\linewidth}
    \includegraphics[width=\linewidth]{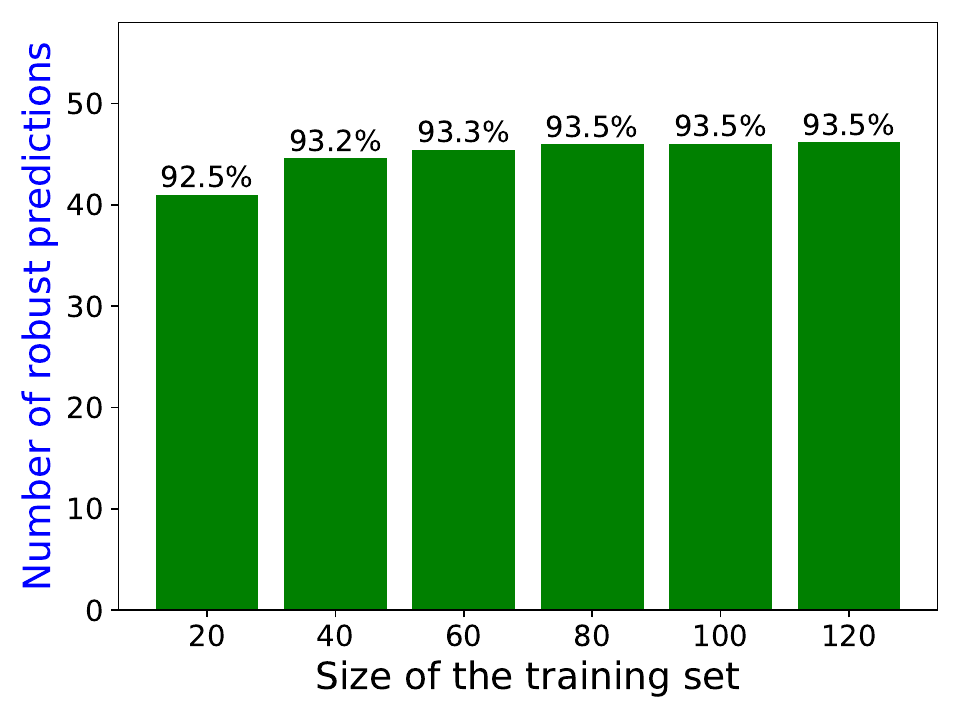}
    \caption{\footnotesize \emph{Former +} Chemical}
    \end{subfigure}
    \begin{subfigure}{0.24\linewidth}
    \includegraphics[width=\linewidth]{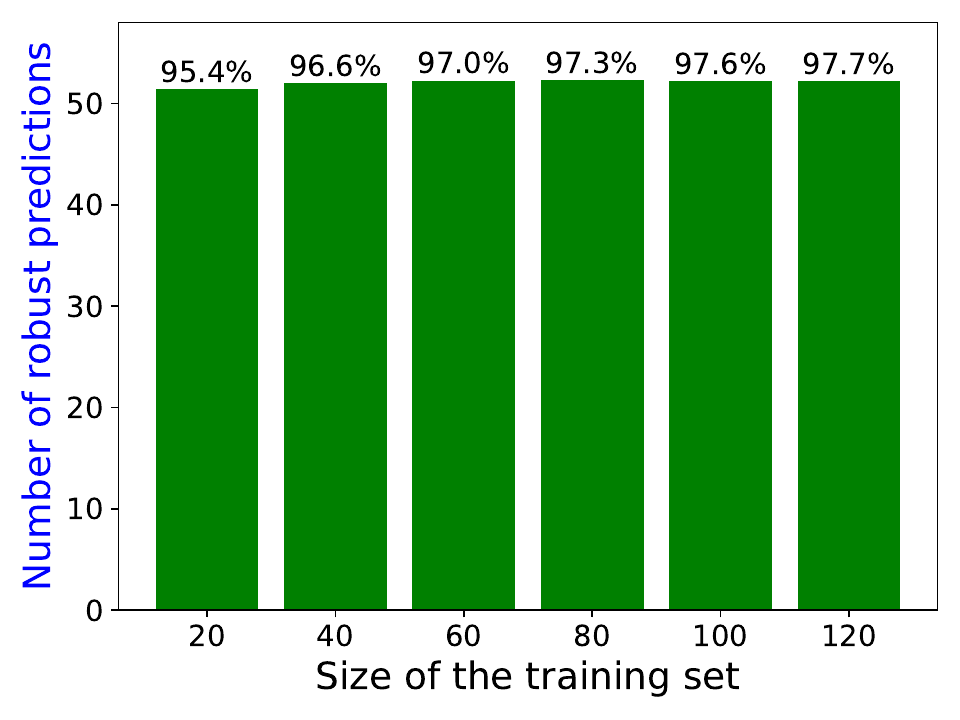}
    \caption{\footnotesize \emph{Former +} Acidity}
    \end{subfigure}
    \caption{Global test performance (top) and number of reliable predictions with test performance for reliable predictions above each bar (bottom) for each model \emph{vs.} size of the training set on Wine dataset with Centroid-based decomposition.}\label{figure:wine_cent}
\end{figure}

\begin{figure}
    \centering
    \begin{subfigure}{0.24\linewidth}
    \includegraphics[width=\linewidth]{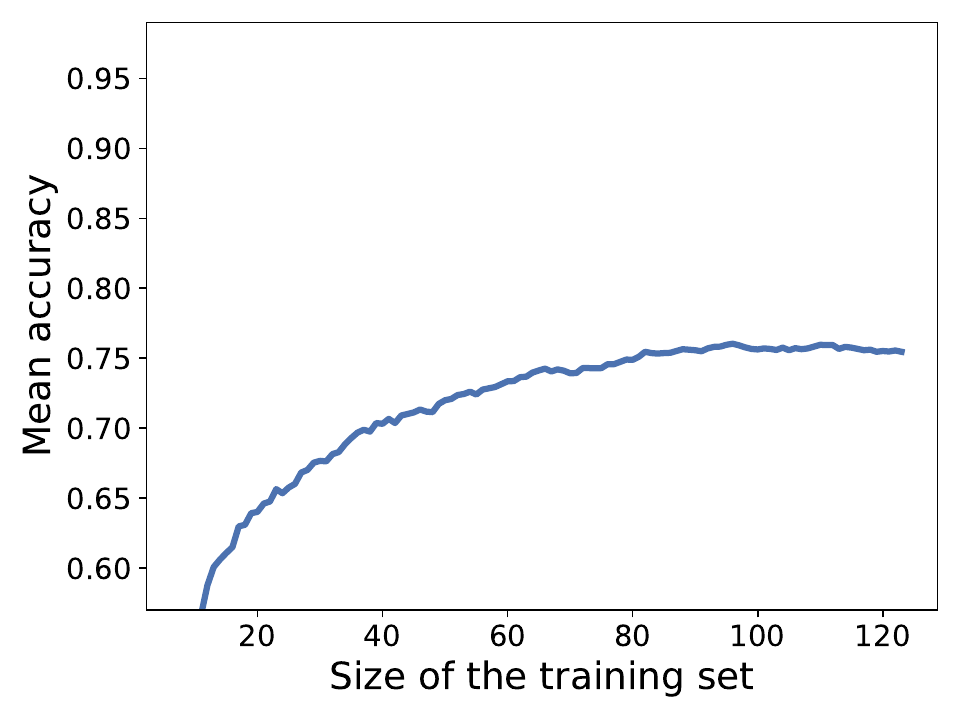}
    \end{subfigure}
    \begin{subfigure}{0.24\linewidth}
    \includegraphics[width=\linewidth]{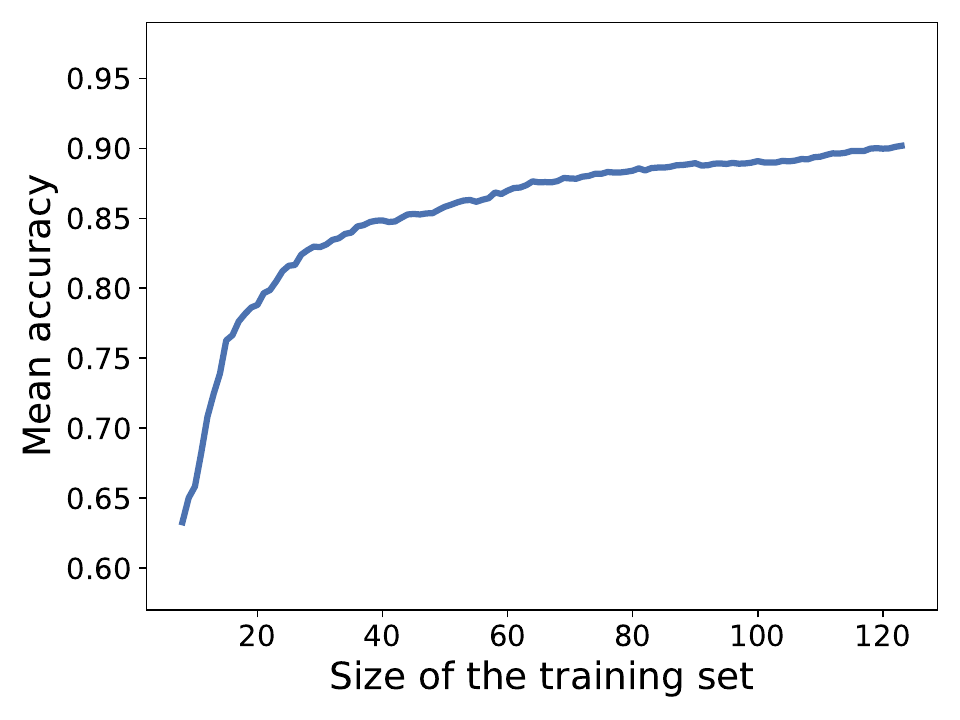}
    \end{subfigure}
    \begin{subfigure}{0.24\linewidth}
    \includegraphics[width=\linewidth]{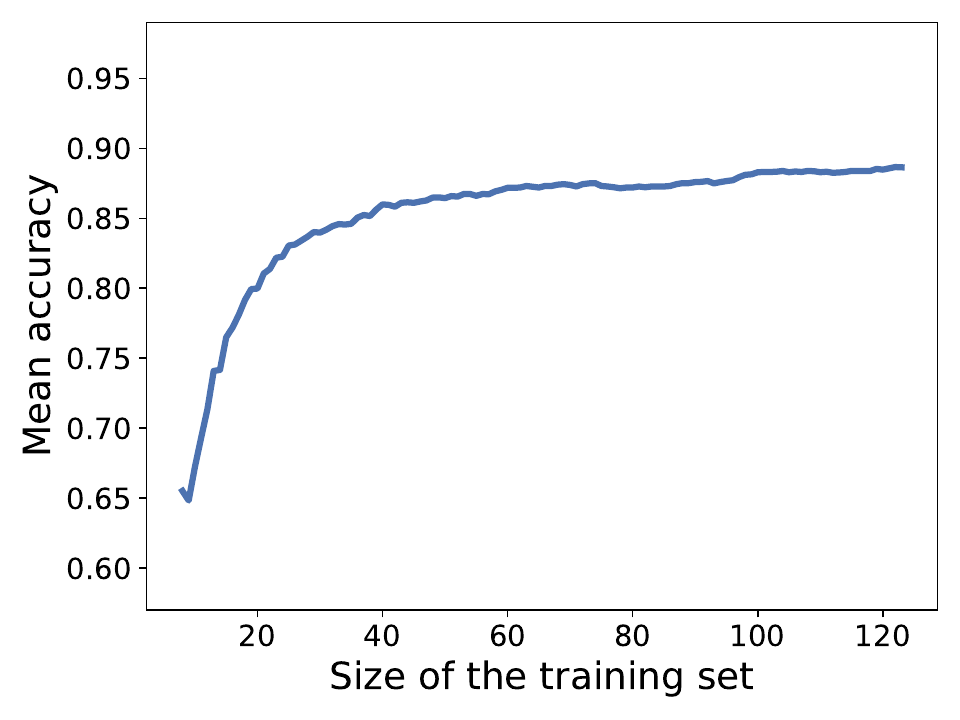}
    \end{subfigure}
    \begin{subfigure}{0.24\linewidth}
    \includegraphics[width=\linewidth]{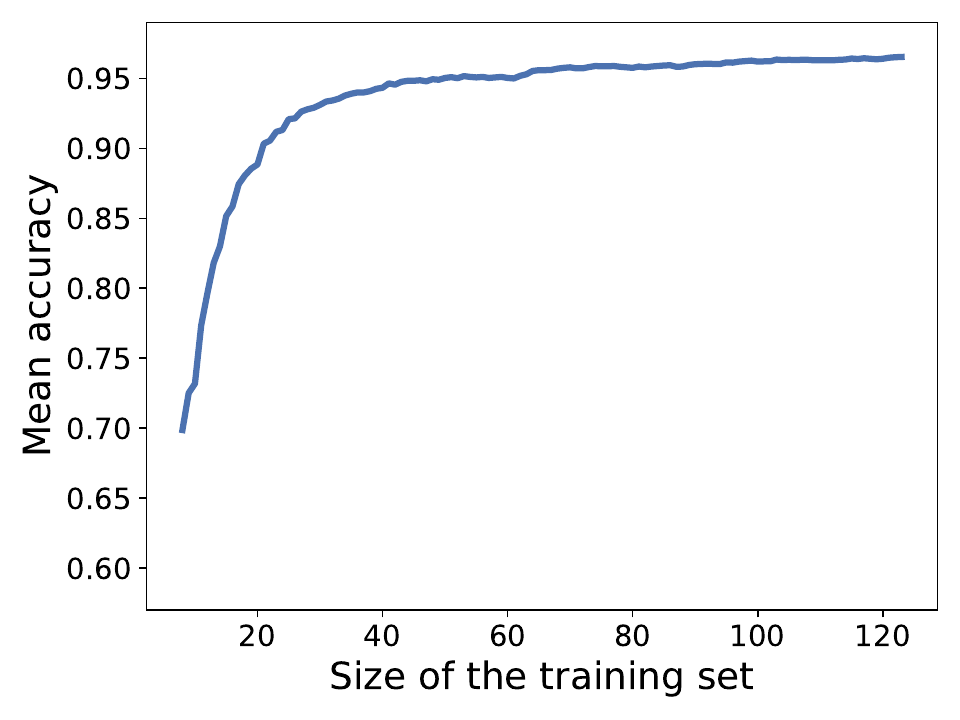}
    \end{subfigure}
    \begin{subfigure}{0.24\linewidth}
    \includegraphics[width=\linewidth]{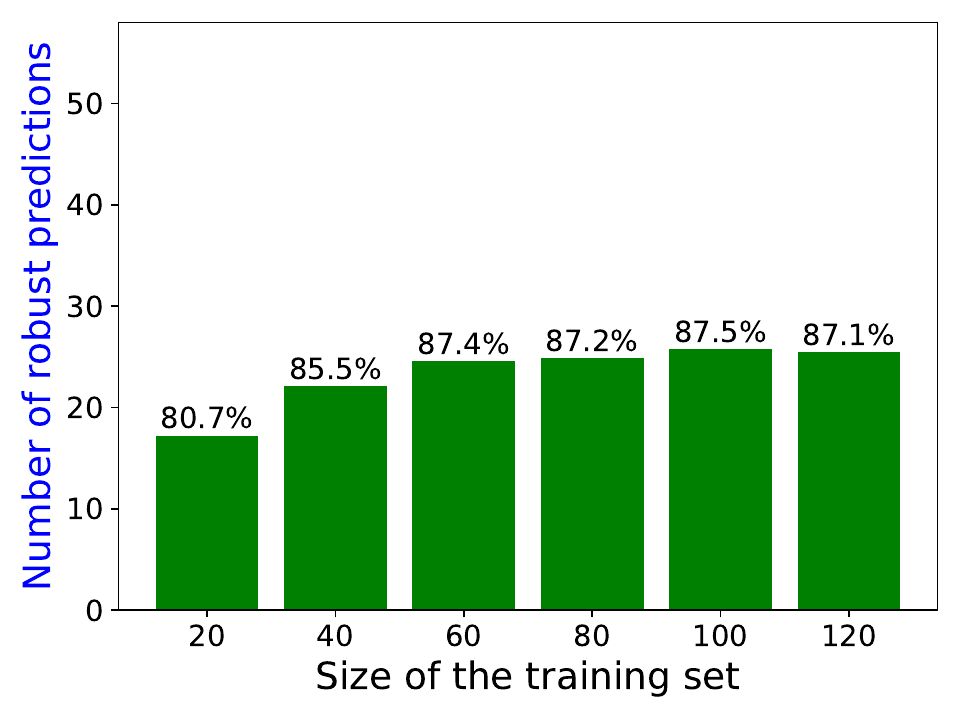}
    \caption{\footnotesize Ignition}
    \end{subfigure}
    \begin{subfigure}{0.24\linewidth}
    \includegraphics[width=\linewidth]{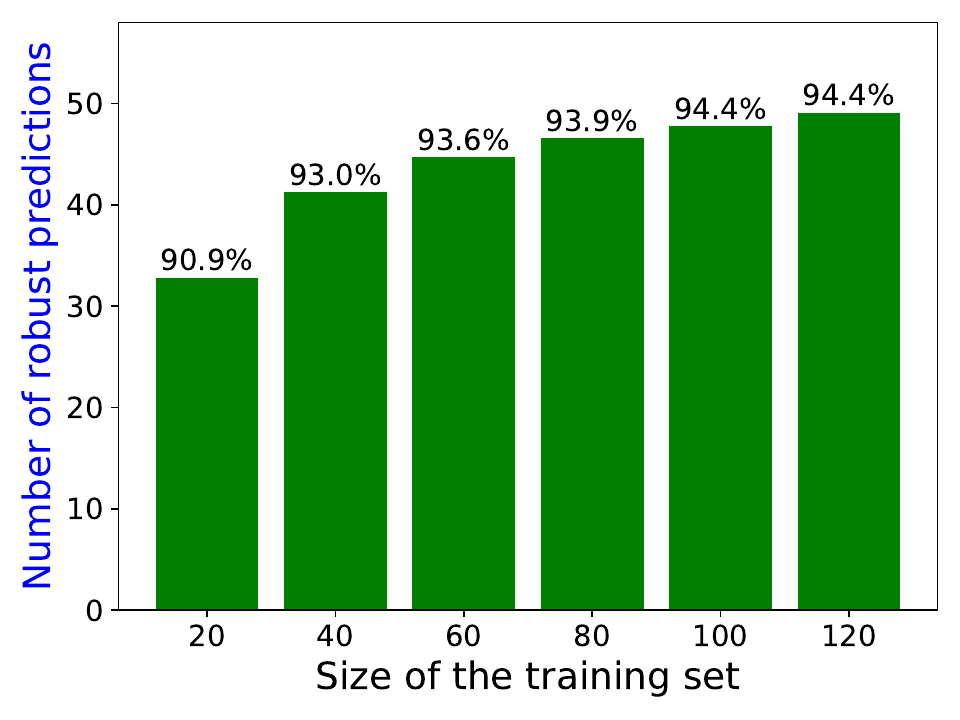}
    \caption{\footnotesize \emph{Former +} Visual}
    \end{subfigure}
    \begin{subfigure}{0.24\linewidth}
    \includegraphics[width=\linewidth]{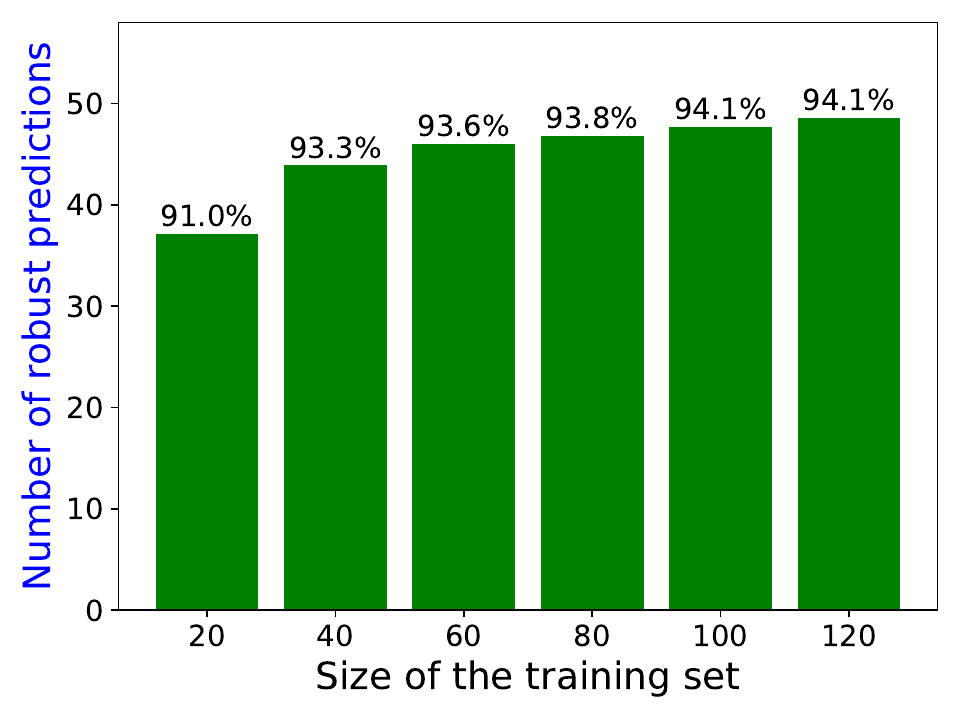}
    \caption{\footnotesize \emph{Former +} Chemical}
    \end{subfigure}
    \begin{subfigure}{0.24\linewidth}
    \includegraphics[width=\linewidth]{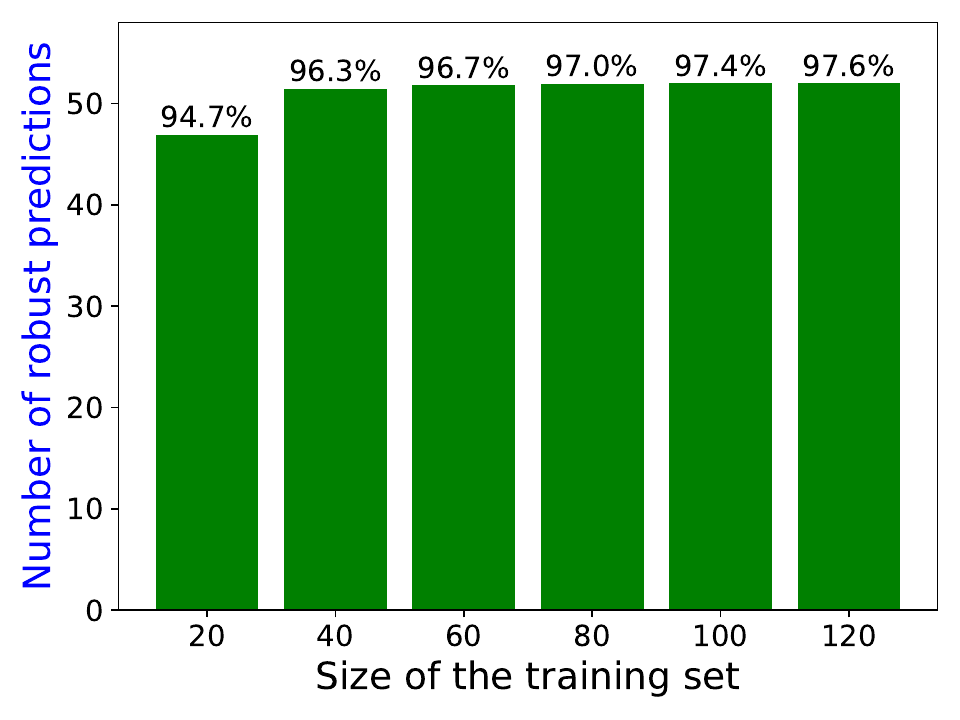}
    \caption{\footnotesize \emph{Former +} Acidity}
    \end{subfigure}
    \caption{Global test performance (top) and number of reliable predictions with test performance for reliable predictions above each bar (bottom) for each model \emph{vs.} size of the training set on Wine dataset with Neighbors-based Forest.}\label{figure:wine_eknn}
\end{figure}

\section{Application to BIOSCAN-5M}\label{app:bioscan}

\begin{figure}
    \centering
    \begin{subfigure}{0.19\linewidth}
    \includegraphics[width=\linewidth]{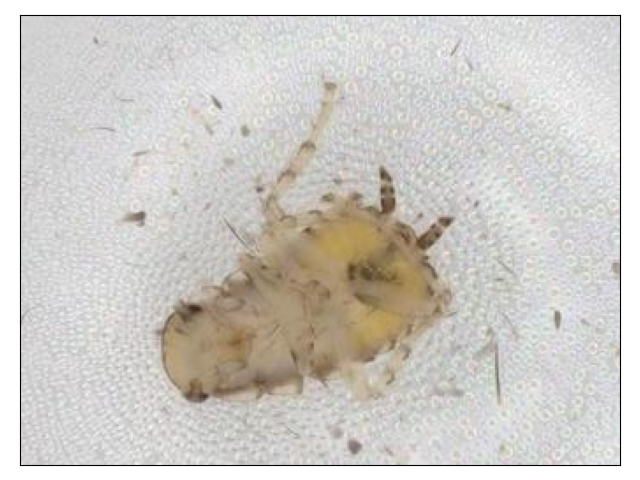}
    \caption{\scriptsize Blattodea}
    \end{subfigure}
    \begin{subfigure}{0.19\linewidth}
    \includegraphics[width=\linewidth]{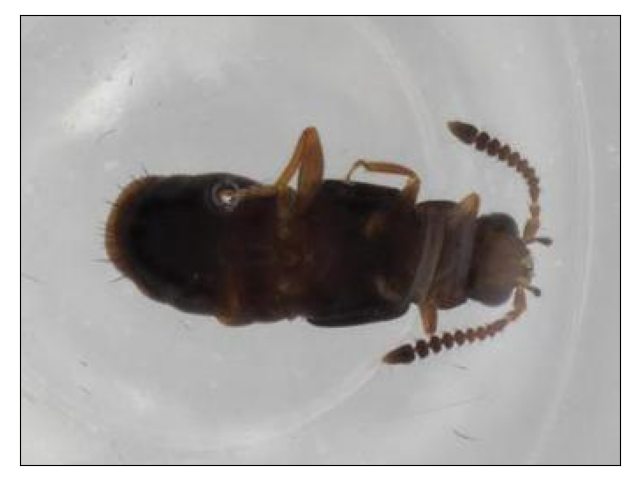}
    \caption{\scriptsize Coleoptera}
    \end{subfigure}
    \begin{subfigure}{0.19\linewidth}
    \includegraphics[width=\linewidth]{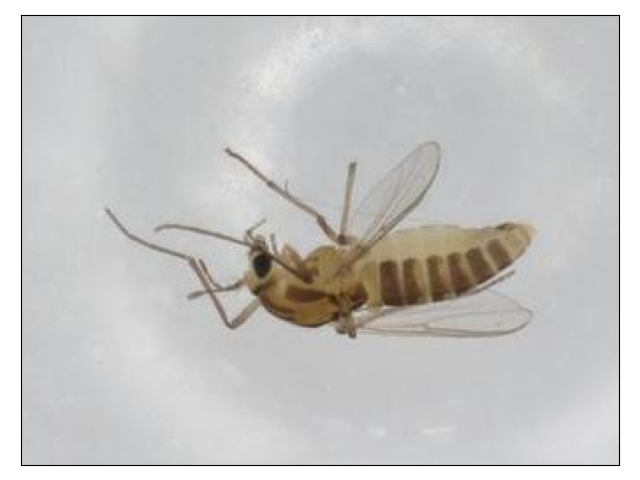}
    \caption{\scriptsize Diptera}
    \end{subfigure}
    \begin{subfigure}{0.19\linewidth}
    \includegraphics[width=\linewidth]{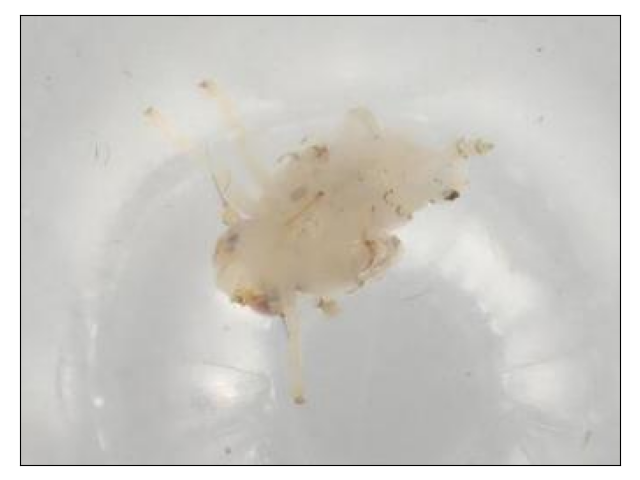}
    \caption{\scriptsize Hemiptera}
    \end{subfigure}
    \begin{subfigure}{0.19\linewidth}
    \includegraphics[width=\linewidth]{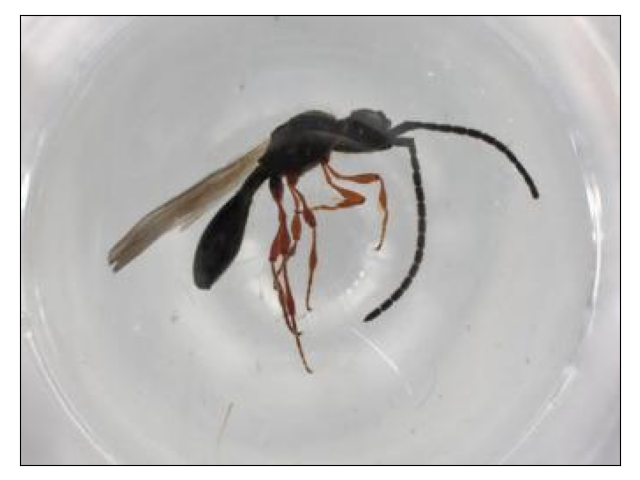}
    \caption{\scriptsize Hymenoptera}
    \end{subfigure}
    \begin{subfigure}{0.19\linewidth}
    \includegraphics[width=\linewidth]{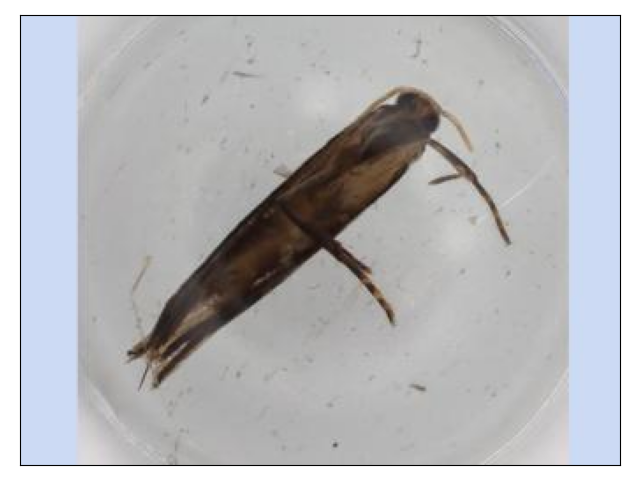}
    \caption{\scriptsize Lepidoptera}
    \end{subfigure}
    \begin{subfigure}{0.19\linewidth}
    \includegraphics[width=\linewidth]{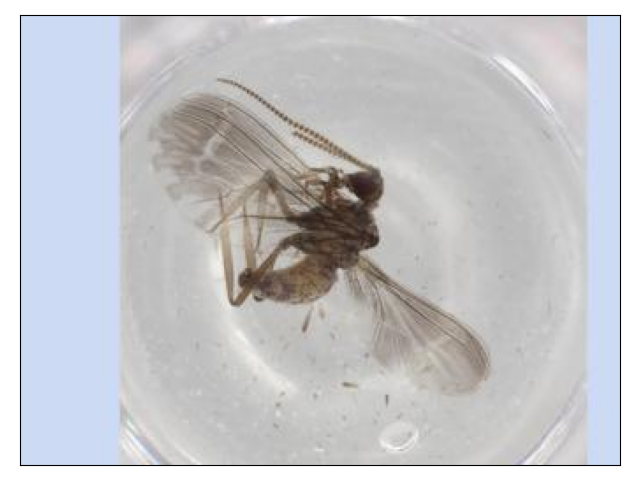}
    \caption{\scriptsize Neuroptera}
    \end{subfigure}
    \begin{subfigure}{0.19\linewidth}
    \includegraphics[width=\linewidth]{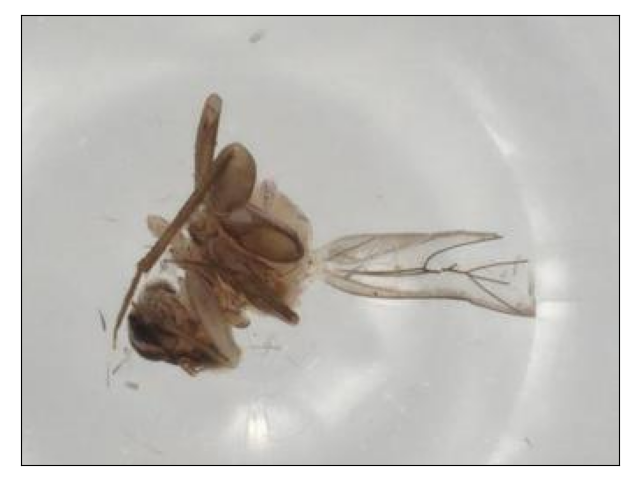}
    \caption{\scriptsize Psocodea}
    \end{subfigure}
    \begin{subfigure}{0.19\linewidth}
    \includegraphics[width=\linewidth]{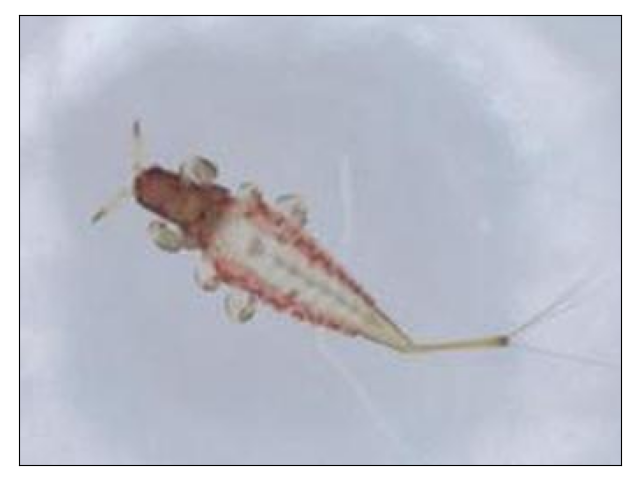}
    \caption{\scriptsize Thysanoptera}
    \end{subfigure}
    \begin{subfigure}{0.19\linewidth}
    \includegraphics[width=\linewidth]{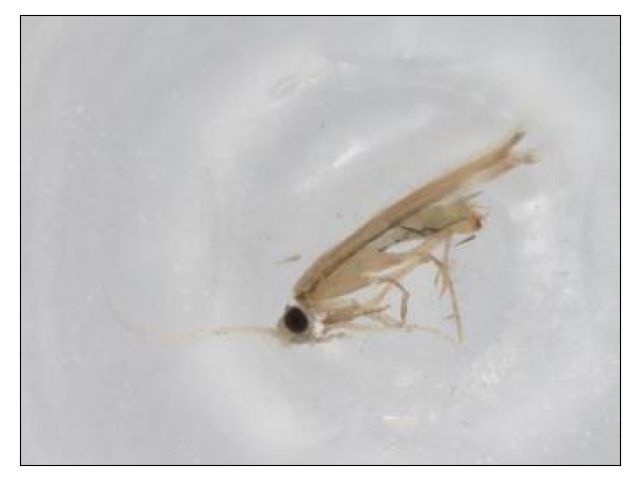}
    \caption{\scriptsize Trichoptera}
    \end{subfigure}
    \caption{BIOSCAN-5M: 10 oders for the class \emph{Insecta}.}\label{figure:bioscan}
\end{figure}

The BIOSCAN-5M~\citep{gharaee2024bioscan5m} dataset is a multi modal dataset for insect biodiversity. Pictures of insects are collected along with geographical information and DNA sequences are collected. For our experiment we used a subset of BIOSCAN-5M of $50k$ training instances and $5k$ test instances uniformly distributed over 10 classes (biological orders of the class \emph{Insecta}), they are presented in Figure~\ref{figure:bioscan}.

Three different models are trained on the dataset according to the train/test split. The first model is trained solely on the images, the second model is trained on both the images and the geographical information, and the third model is trained solely on the DNA sequences.
For the images, we used a ResNet18 model. The images were resized to a $64 \times 64 \times 3$ tensor, and normalization was applied based on ImageNet statistics. We employed a cross-entropy loss with stochastic gradient descent (SGD), using a learning rate of 0.1\footnote{The learning rate for each parameter group was also set using a cosine annealing schedule.}, momentum of 0.9, and a weight decay of $5 \times 10^{-4}$. Each batch had a size of $100$, and the models were trained for 10 epochs.

For the second model, after training, a reduced feature space is extracted from the ultimate layer of the ResNet18. This feature space is then enriched with the geographical information: country and state are encoded using one-hot encoding, while latitude and longitude are included as numerical values. Both sets of features are standardized. 

For the third model, only DNA sequences are considered. We used $k$-mer frequency feature extraction with $k=3$ to encode the DNA sequences.

The centroid method differs slightly from the other, as centroids must be updated during every batch in a deep learning scenario. Consequently, the implementation is somewhat different. However, to ensure a fair comparison, a similar ResNet18~\citep{he2016} architecture is trained using common parameters and the same number of epochs. The authors of the method~\citep{van-amersfoort20a} provide a PyTorch implementation of their approach, which we used for training the model.
The images were resized to a $64 \times 64 \times 3$ tensor, and normalization was applied using ImageNet statistics. A cross-entropy loss function was used alongside stochastic gradient descent (SGD) with a learning rate of 0.05, momentum of 0.9, and a weight decay of $5 \times 10^{-4}$. Each batch consisted of 100 samples, and the models were trained for 10 epochs. This method require the last layer of the ResNet to be modified (from 10 outputs to 512). In each batch, the centroids are updated according to the proposed method described by the authors.

The results are presented for all the studied models in Figures~\ref{figure:evolution}~\ref{figure:bios-bdl},~\ref{figure:bios-rf},~\ref{figure:bios-cent} and~\ref{figure:bios-eknn}.

\begin{figure}
    \centering
    \begin{subfigure}{0.32\linewidth}
    \includegraphics[width=\linewidth]{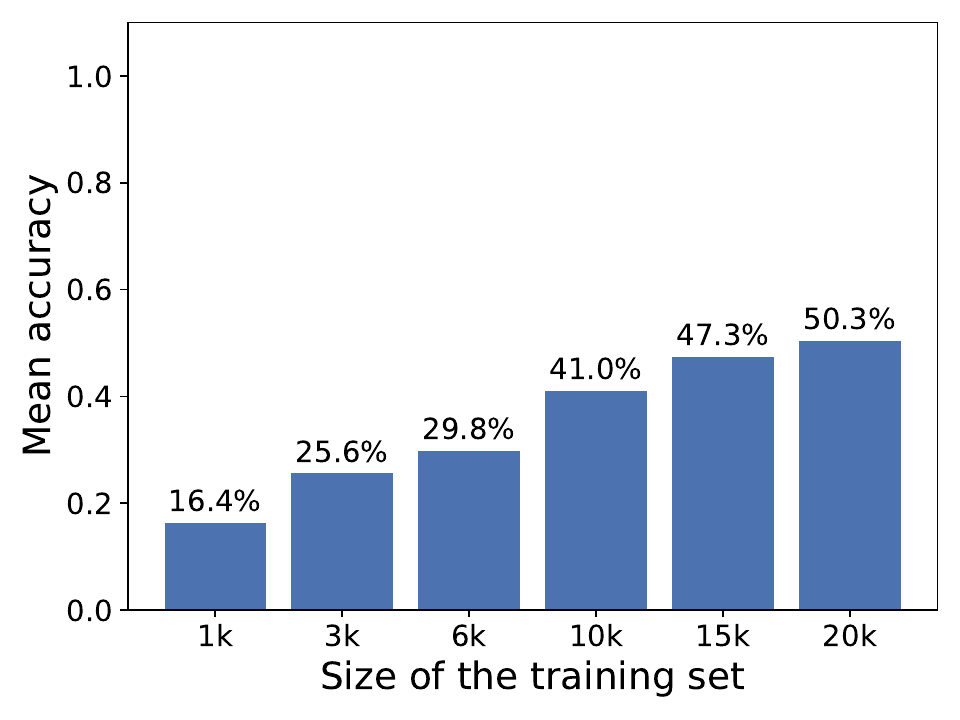}
    \end{subfigure}
    \begin{subfigure}{0.32\linewidth}
    \includegraphics[width=\linewidth]{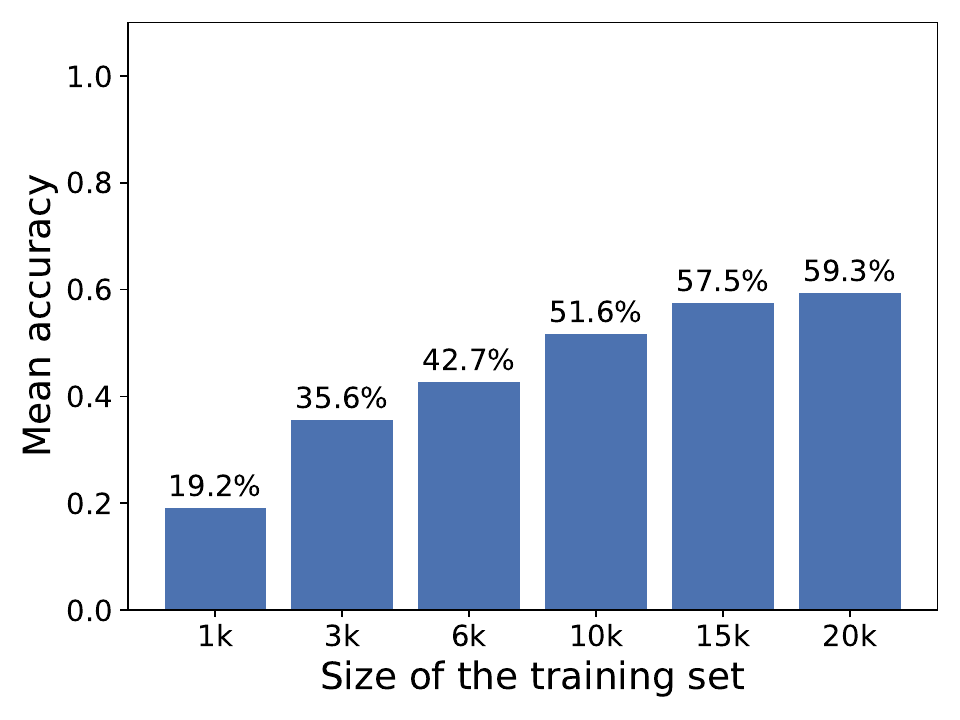}
    \end{subfigure}
    \begin{subfigure}{0.32\linewidth}
    \includegraphics[width=\linewidth]{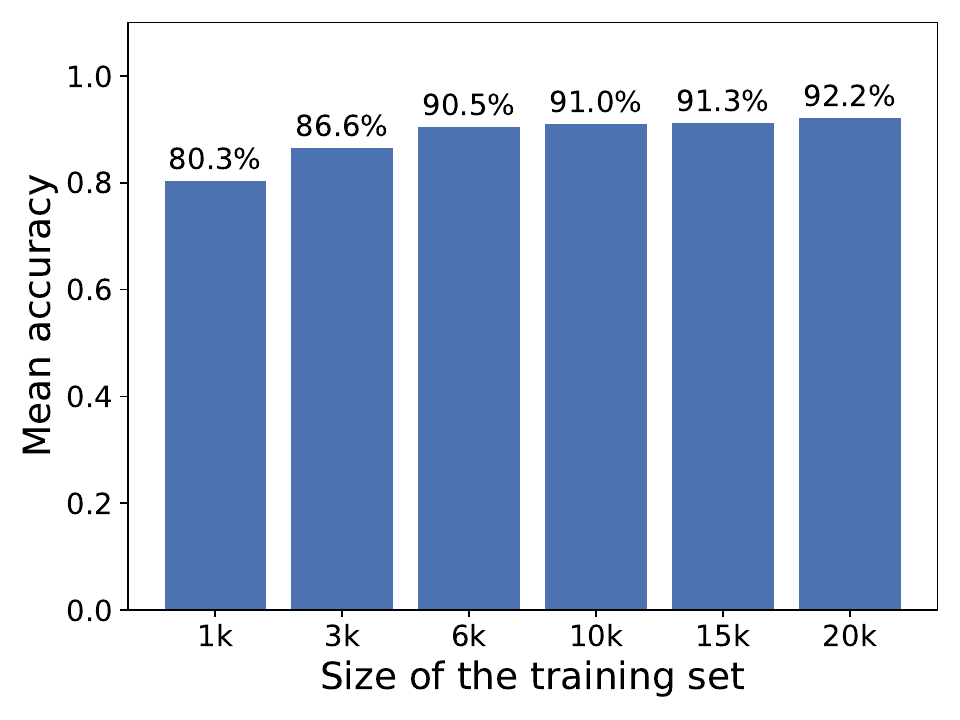}
    \end{subfigure}
    \begin{subfigure}{0.32\linewidth}
    \includegraphics[width=\linewidth]{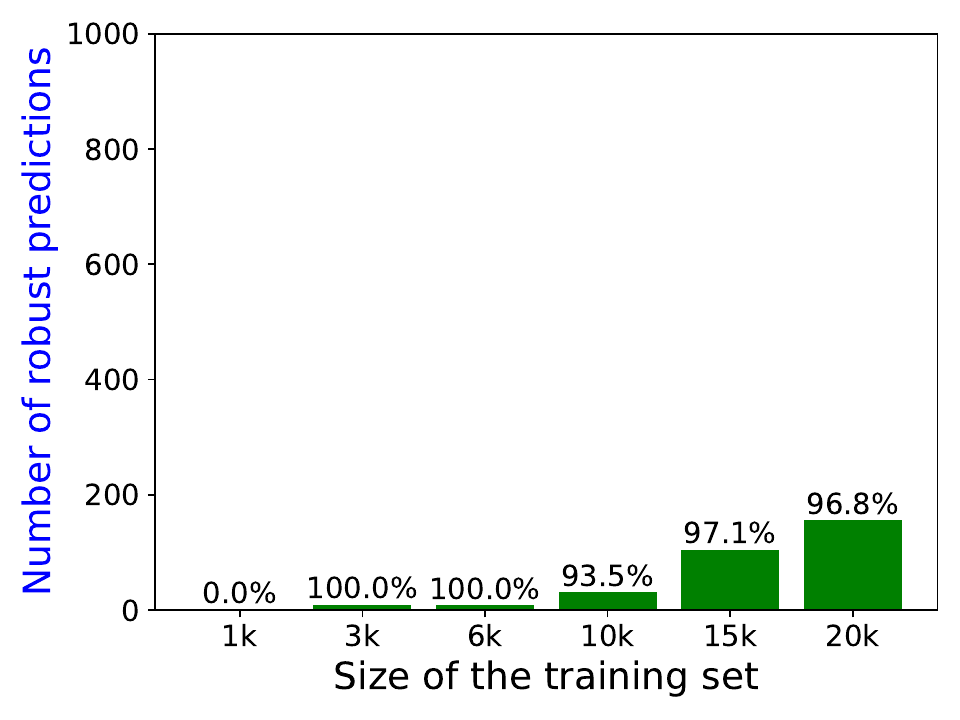}
    \caption{Image}
    \end{subfigure}
    \begin{subfigure}{0.32\linewidth}
    \includegraphics[width=\linewidth]{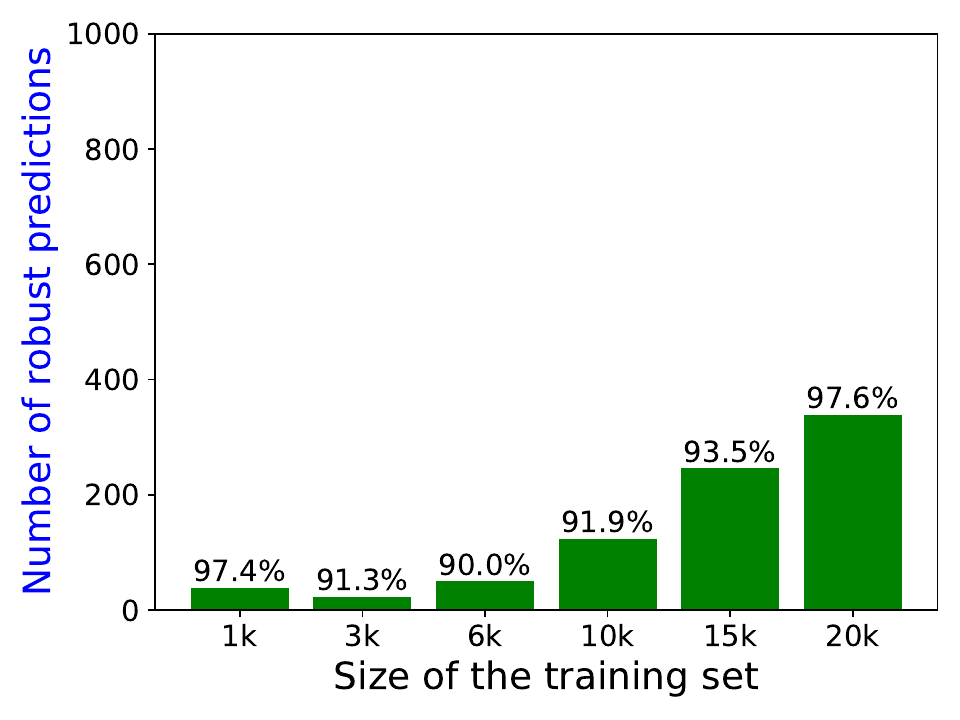}
    \caption{Image + Geographical}
    \end{subfigure}
    \begin{subfigure}{0.32\linewidth}
    \includegraphics[width=\linewidth]{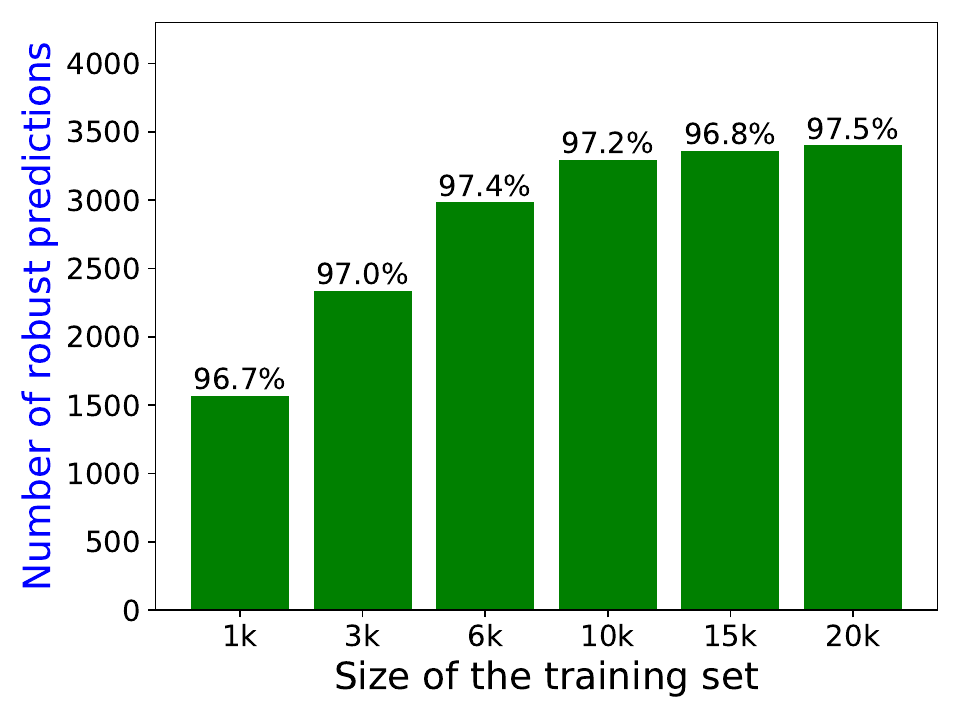}
    \caption{DNA Sequence}
    \end{subfigure}
    \caption{Global test performance (top) and number of reliable predictions with test performance for reliable predictions above each bar (bottom) for each model \emph{vs.} size of the training set on BIOSCAN-5M dataset with Bayesian Deep Learning}\label{figure:bios-bdl}
\end{figure}

\begin{figure}
    \centering
    \begin{subfigure}{0.32\linewidth}
    \includegraphics[width=\linewidth]{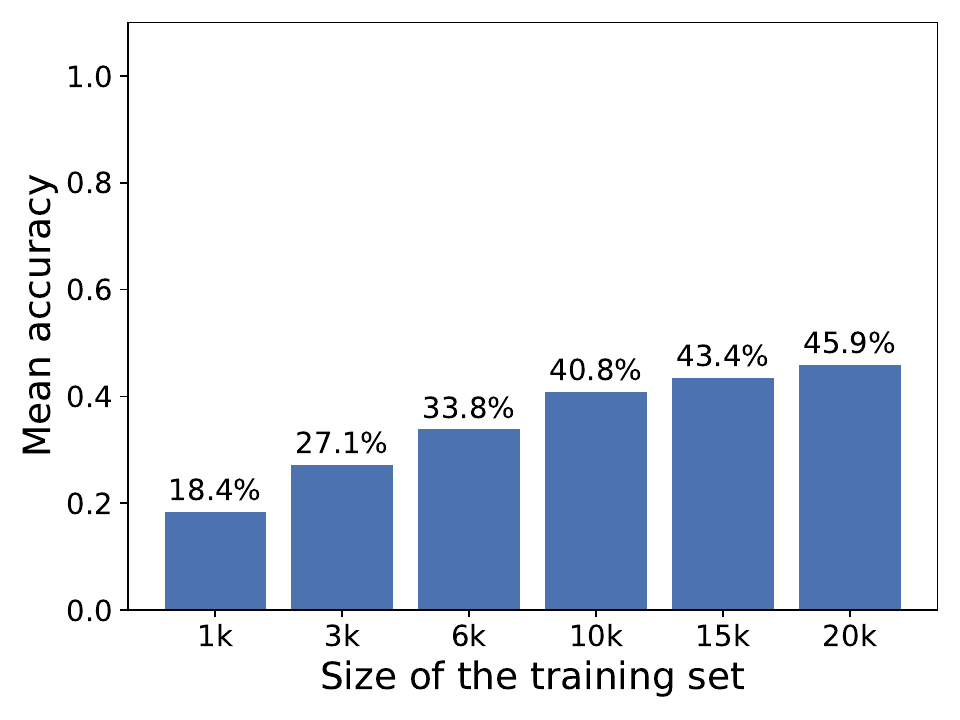}
    \end{subfigure}
    \begin{subfigure}{0.32\linewidth}
    \includegraphics[width=\linewidth]{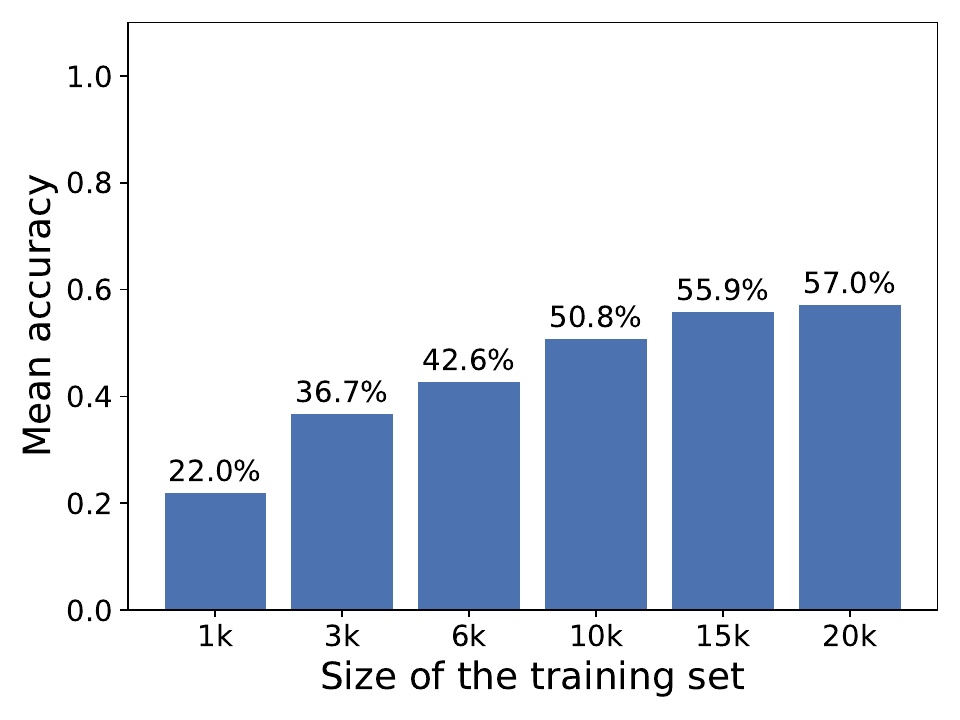}
    \end{subfigure}
    \begin{subfigure}{0.32\linewidth}
    \includegraphics[width=\linewidth]{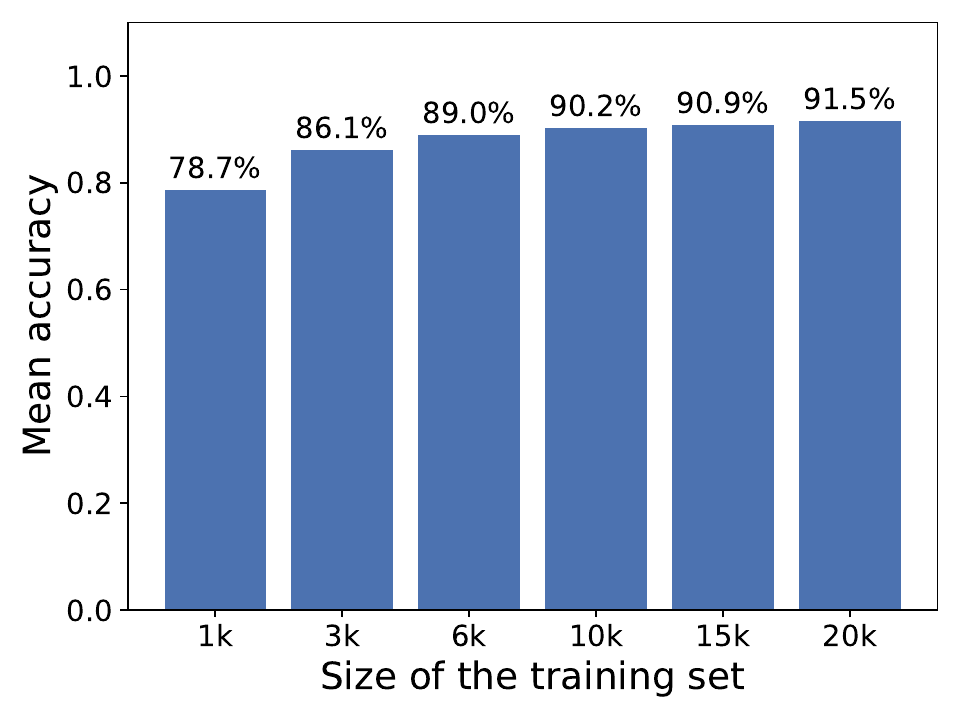}
    \end{subfigure}
    \begin{subfigure}{0.32\linewidth}
    \includegraphics[width=\linewidth]{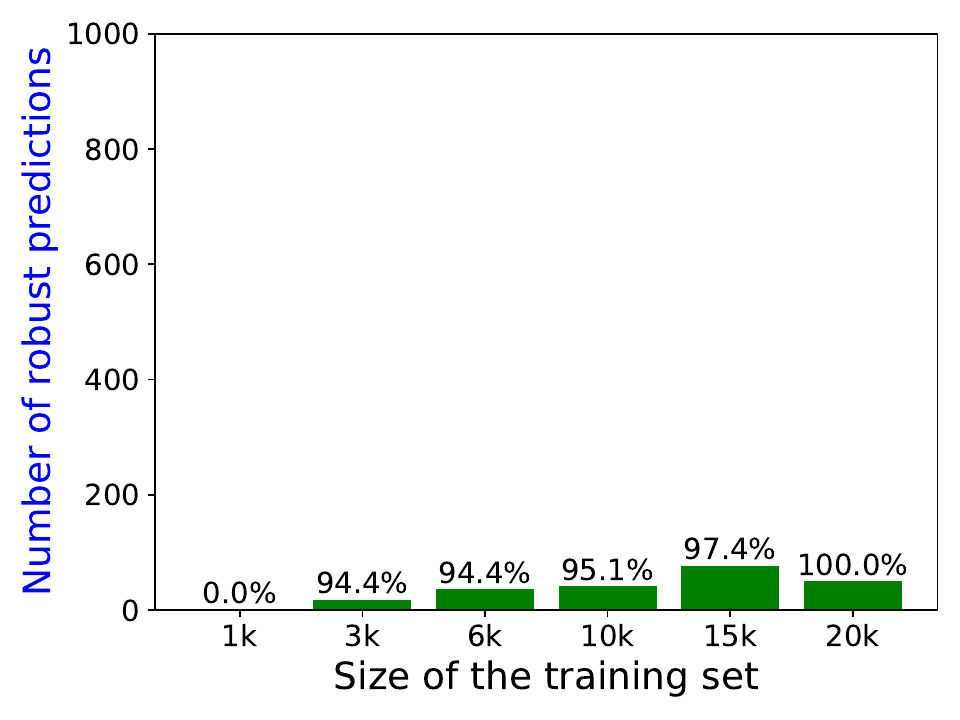}
    \caption{Image}
    \end{subfigure}
    \begin{subfigure}{0.32\linewidth}
    \includegraphics[width=\linewidth]{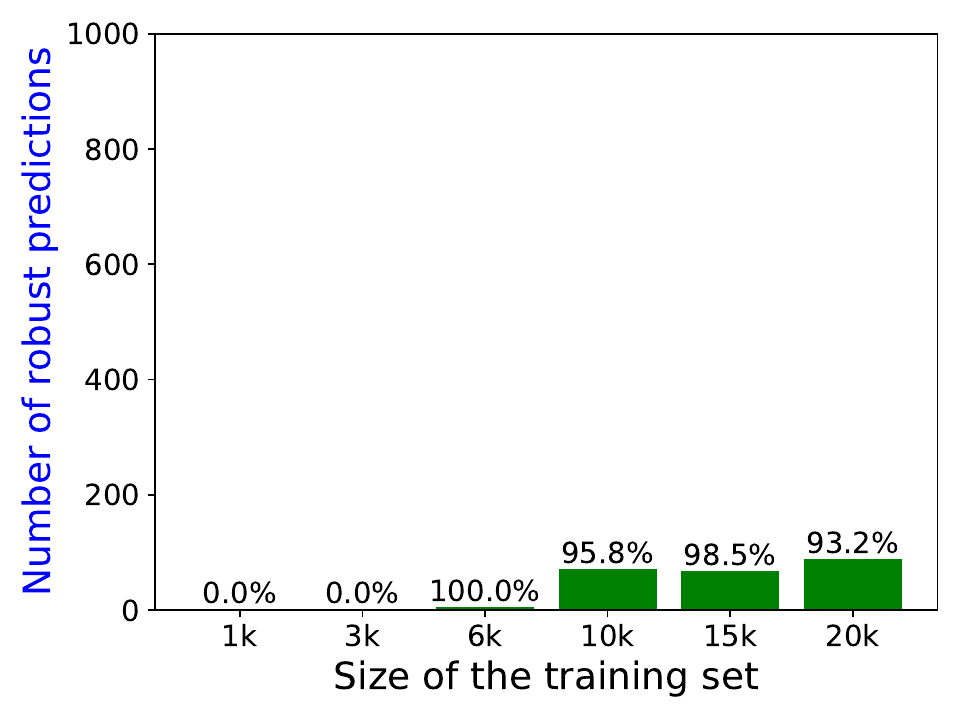}
    \caption{Image + Geographical}
    \end{subfigure}
    \begin{subfigure}{0.32\linewidth}
    \includegraphics[width=\linewidth]{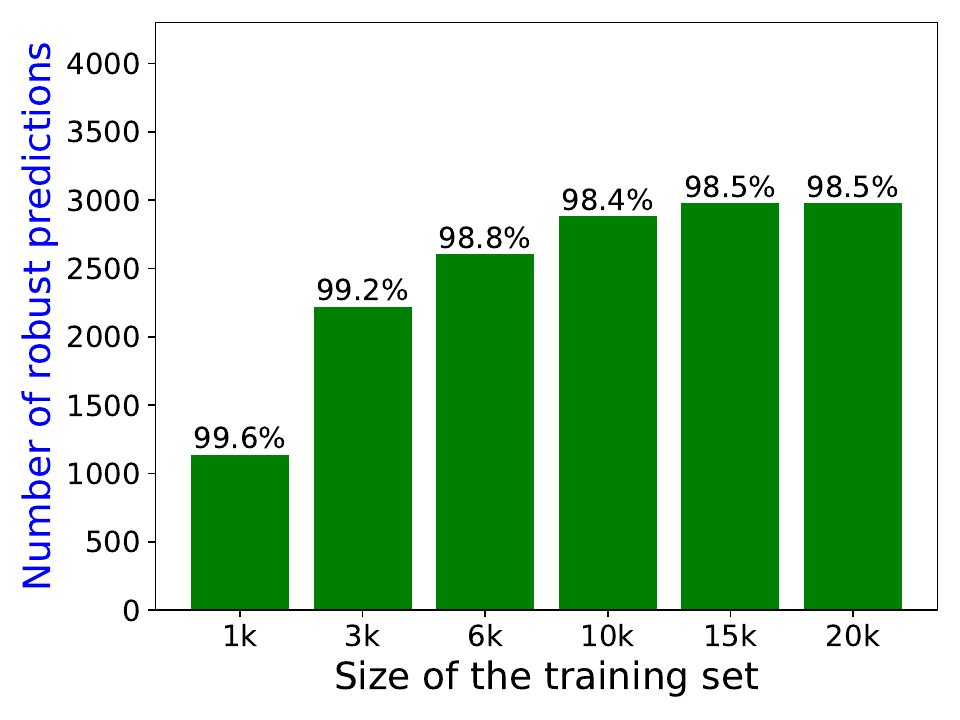}
    \caption{DNA Sequence}
    \end{subfigure}
    \caption{Global test performance (top) and number of reliable predictions with test performance for reliable predictions above each bar (bottom) for each model \emph{vs.} size of the training set on BIOSCAN-5M dataset with Random Forest}\label{figure:bios-rf}
\end{figure}

\begin{figure}
    \centering
    \begin{subfigure}{0.32\linewidth}
    \includegraphics[width=\linewidth]{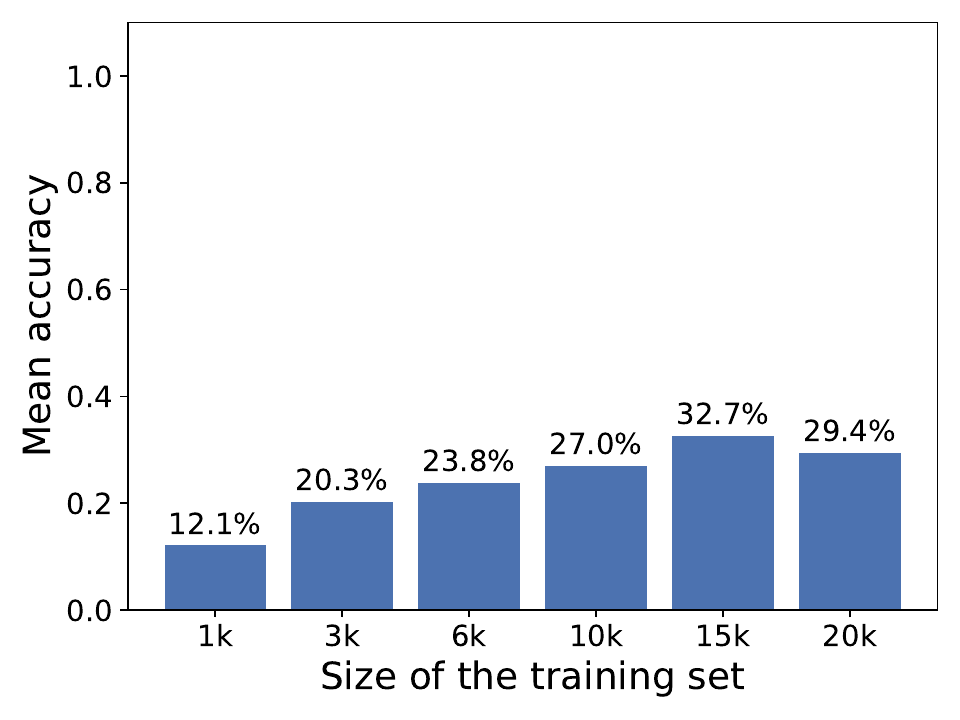}
    \end{subfigure}
    \begin{subfigure}{0.32\linewidth}
    \includegraphics[width=\linewidth]{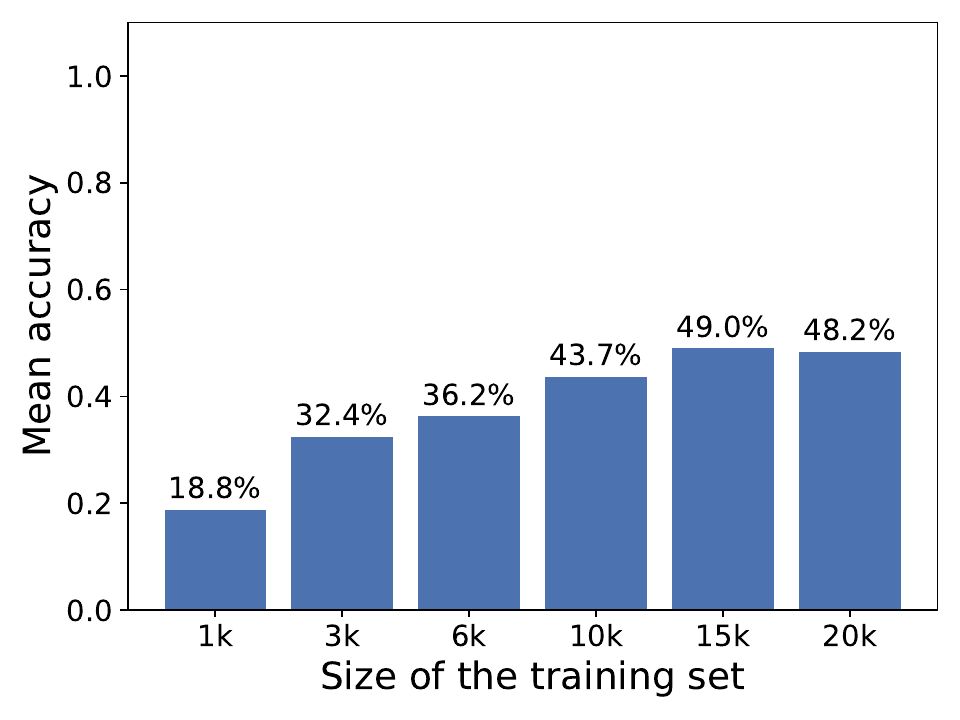}
    \end{subfigure}
    \begin{subfigure}{0.32\linewidth}
    \includegraphics[width=\linewidth]{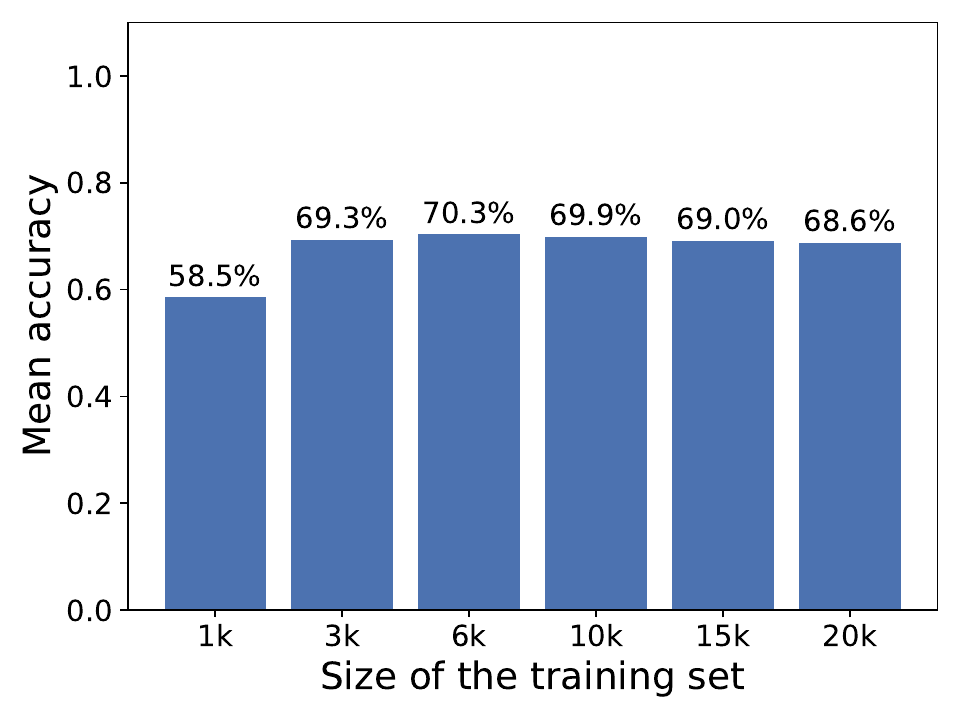}
    \end{subfigure}
    \begin{subfigure}{0.32\linewidth}
    \includegraphics[width=\linewidth]{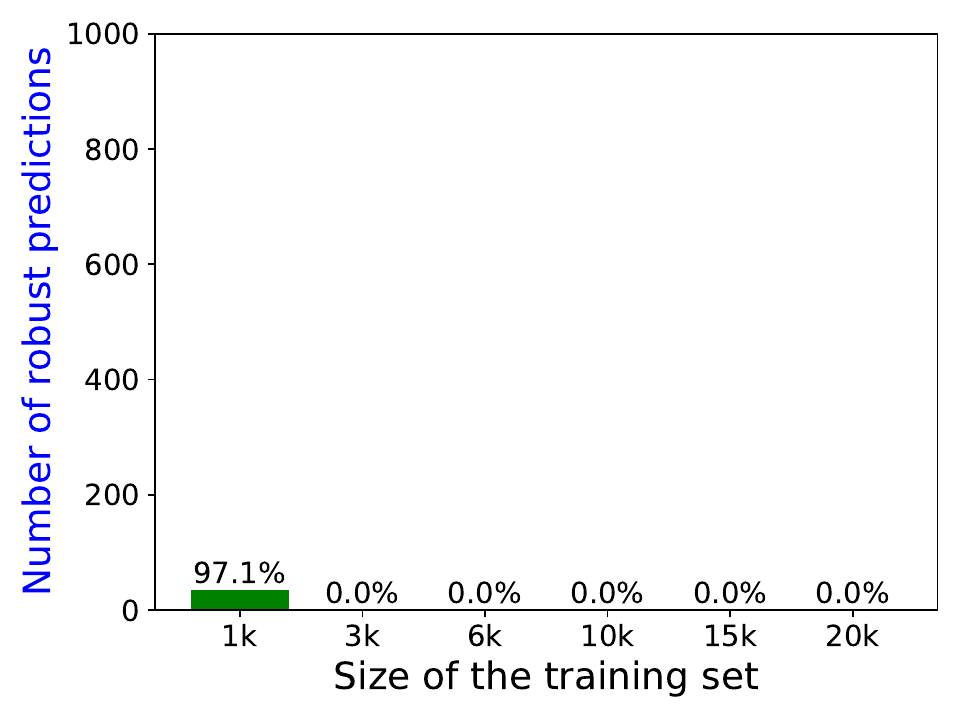}
    \caption{Image}
    \end{subfigure}
    \begin{subfigure}{0.32\linewidth}
    \includegraphics[width=\linewidth]{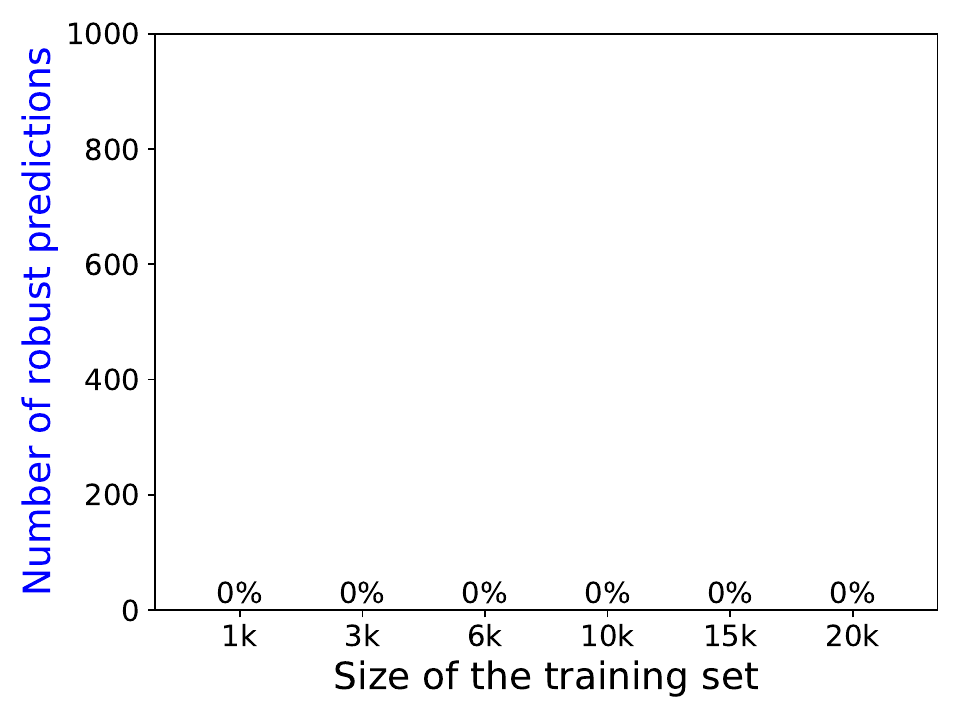}
    \caption{Image + Geographical}
    \end{subfigure}
    \begin{subfigure}{0.32\linewidth}
    \includegraphics[width=\linewidth]{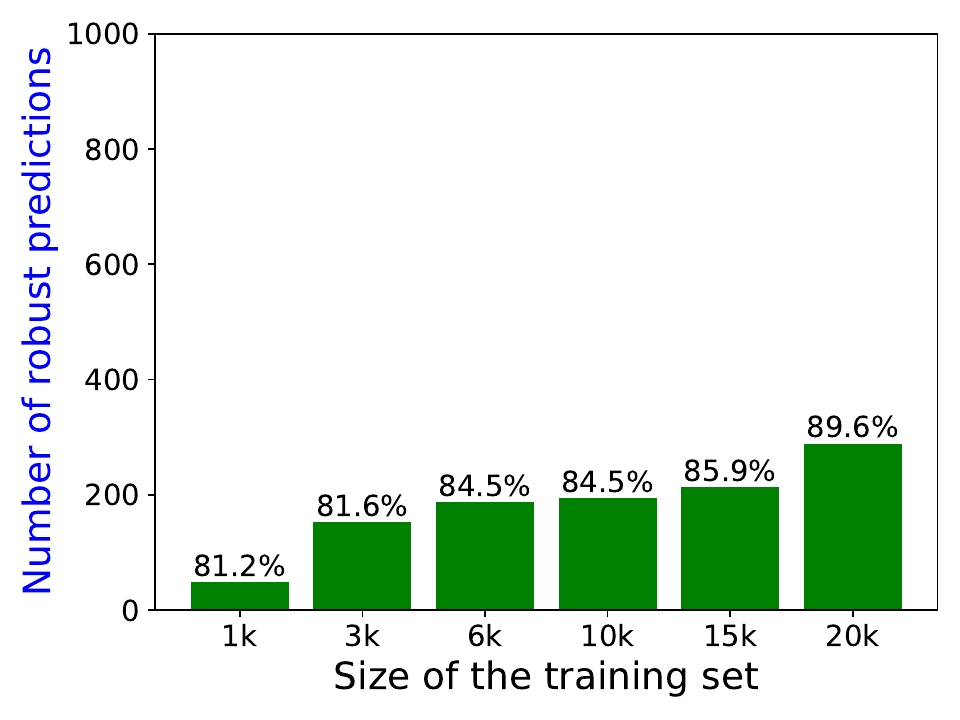}
    \caption{DNA Sequence}
    \end{subfigure}
    \caption{Global test performance (top) and number of reliable predictions with test performance for reliable predictions above each bar (bottom) for each model \emph{vs.} size of the training set on BIOSCAN-5M dataset with Centroid-based decomposition.}\label{figure:bios-cent}
\end{figure}

\begin{figure}
    \centering
    \begin{subfigure}{0.32\linewidth}
    \includegraphics[width=\linewidth]{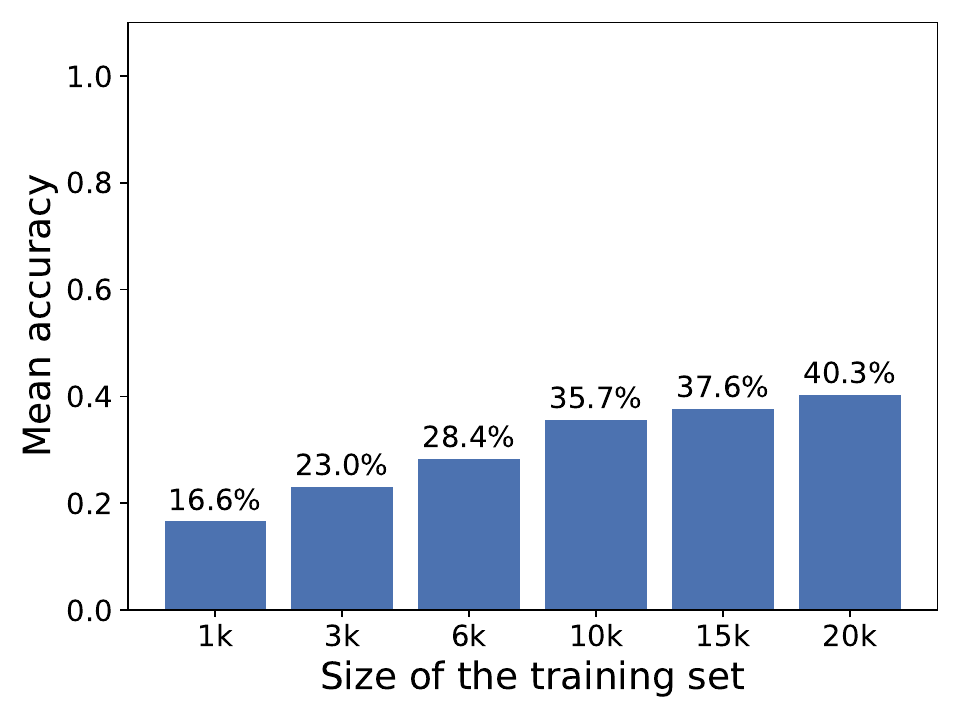}
    \end{subfigure}
    \begin{subfigure}{0.32\linewidth}
    \includegraphics[width=\linewidth]{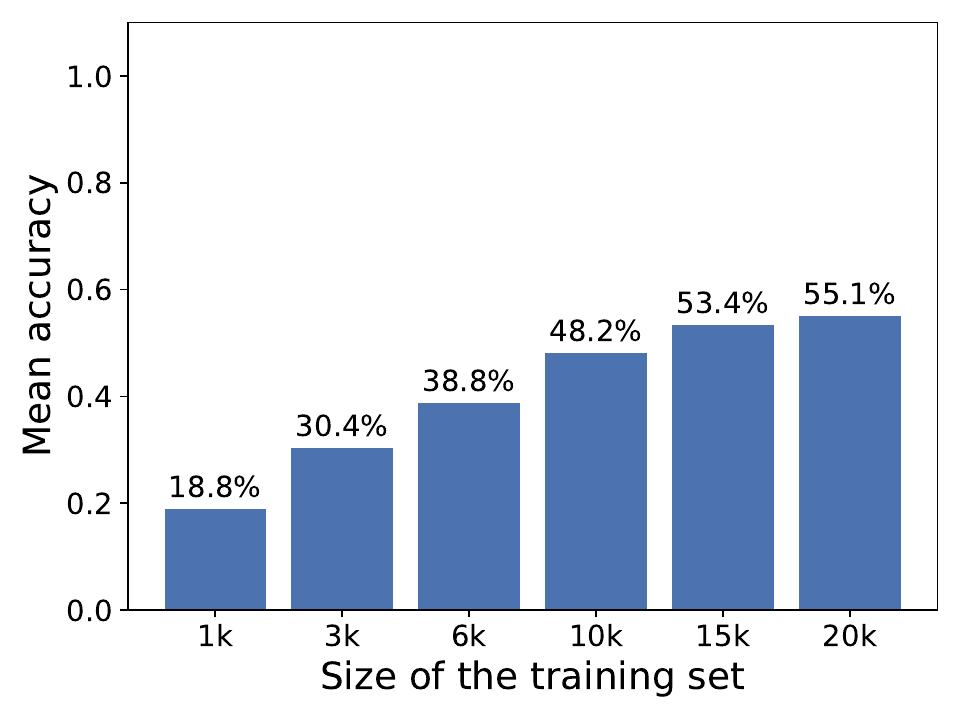}
    \end{subfigure}
    \begin{subfigure}{0.32\linewidth}
    \includegraphics[width=\linewidth]{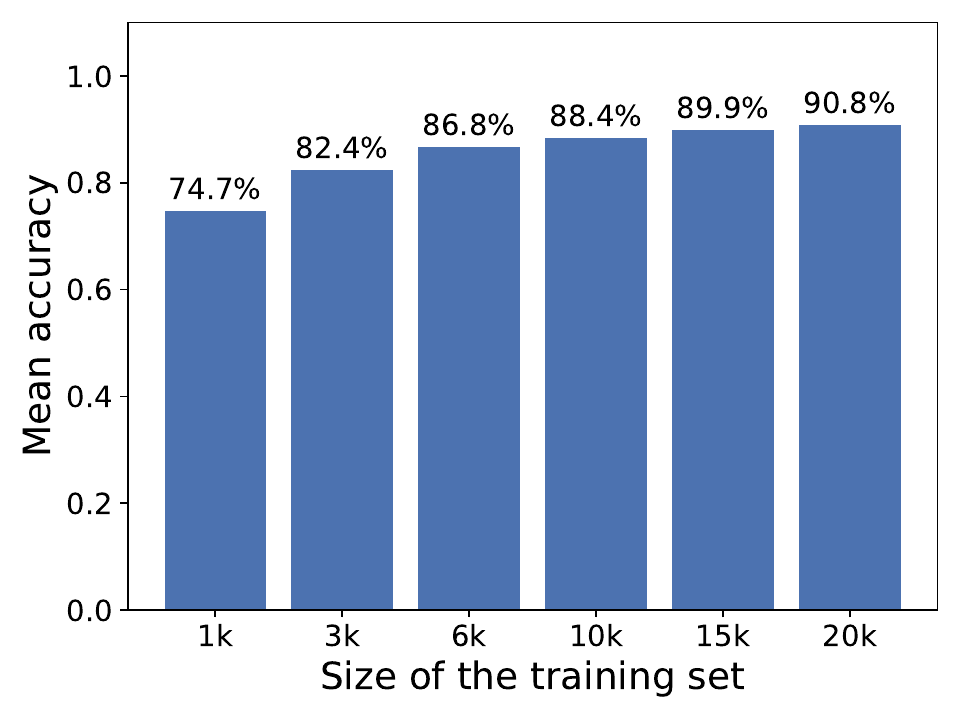}
    \end{subfigure}
    \begin{subfigure}{0.32\linewidth}
    \includegraphics[width=\linewidth]{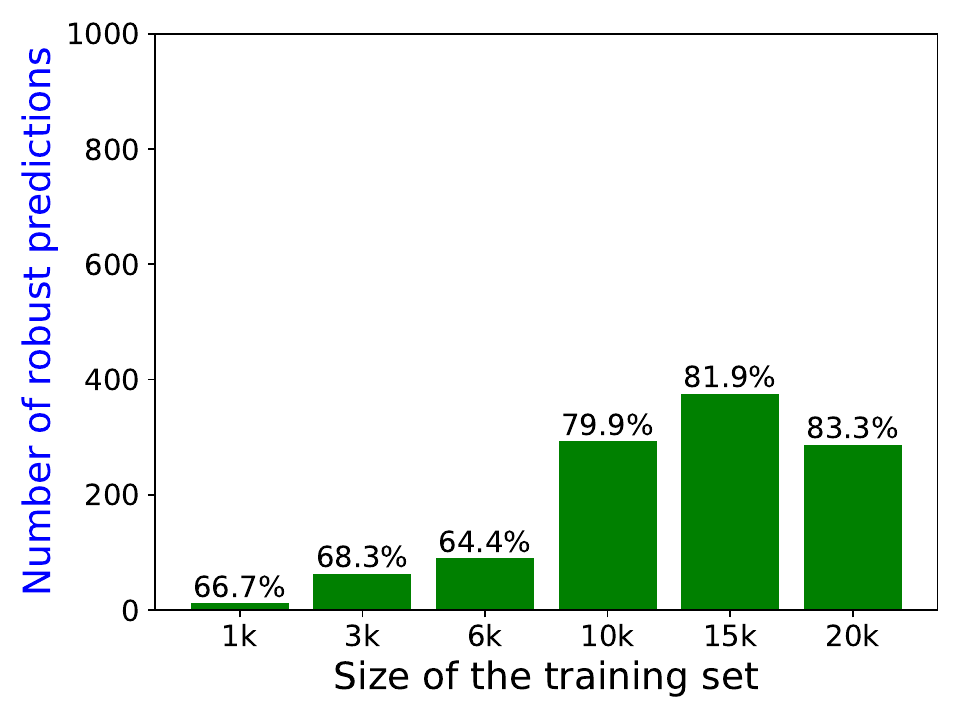}
    \caption{Image}
    \end{subfigure}
    \begin{subfigure}{0.32\linewidth}
    \includegraphics[width=\linewidth]{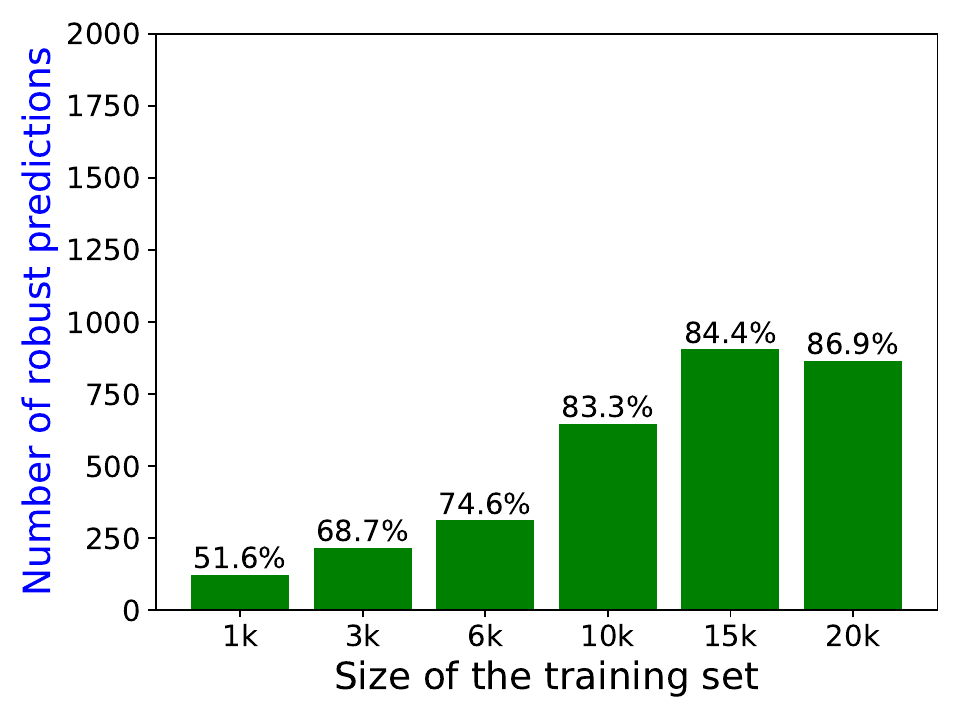}
    \caption{Image + Geographical}
    \end{subfigure}
    \begin{subfigure}{0.32\linewidth}
    \includegraphics[width=\linewidth]{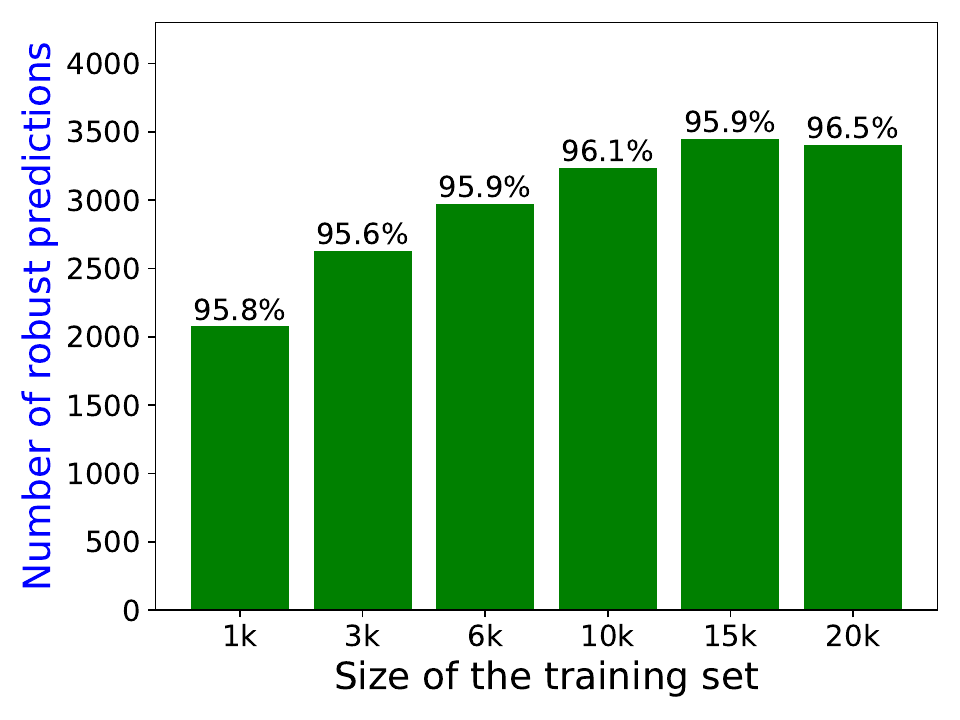}
    \caption{DNA Sequence}
    \end{subfigure}
    \caption{Global test performance (top) and number of reliable predictions with test performance for reliable predictions above each bar (bottom) for each model \emph{vs.} size of the training set on BIOSCAN-5M dataset with Neighbors-based decomposition.}\label{figure:bios-eknn}
\end{figure}

\section{Application to MIMIC-IV}\label{app:mimic}

MIMIC-IV\citep{Johnson2023} is a freely accessible electronic health record dataset. Information available includes patient measurements, orders, diagnoses, procedures, treatments, and de-identified free-text clinical notes. We extracted three different modalities to detect the eight diseases presented in Table~\ref{tab:mimic}. The Online Medical Record (OMR) contains information such as gender, age, height, and weight, which do not require expensive procedures. Classical laboratory tests (from blood samples) are also conducted, along with a third modality comprising results from microbiological cultures.
For our experiment we used a subset of MIMIC-IV of $100k$ training instances and $4k$ test instances distributed with high imbalance over 8 diseases, they are presented in Table~\ref{tab:mimic}. For each modality, we considered only the patient's first test during the hospital stay. A patient may also be diagnosed with multiple diseases, in such cases, we only consider the most severe one, following a subjective order.
The dataset also contains several missing values, so we applied a simple mean imputation method.

Three different models are trained on the dataset according to the train/test split. The first model is trained solely on the patient's Online Medical Record (OMR), which includes gender, age, weight, height, BMI, and blood pressure. The second model is trained on both the OMR and laboratory test results (blood samples), comprising 40 variables. The third model is trained on the previously mentioned variables, enriched with microbiological culture data for three different bacteria, resulting in 50 additional features.

The results are presented for all the studied models in Figures~\ref{fig:mimic-de}~\ref{fig:mimic-edl},~\ref{fig:mimic-rf},~\ref{fig:mimic-cent} and~\ref{fig:mimic-eknn}.

\begin{figure}
    \centering
    \begin{subfigure}{0.29\linewidth}
    \includegraphics[width=\linewidth]{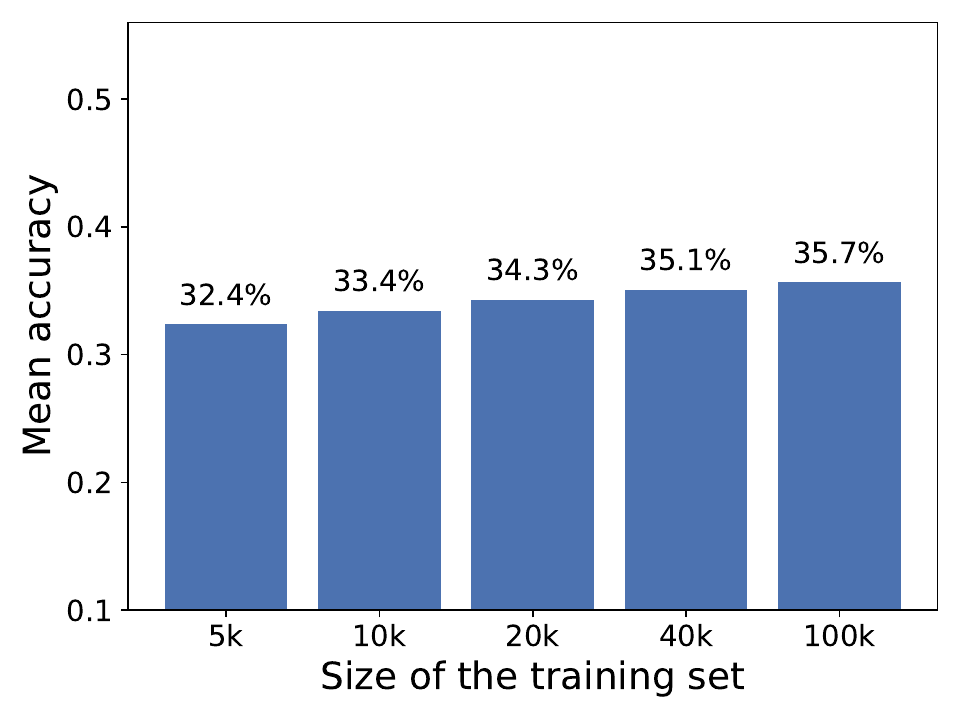}
    \end{subfigure}
    \begin{subfigure}{0.29\linewidth}
    \includegraphics[width=\linewidth]{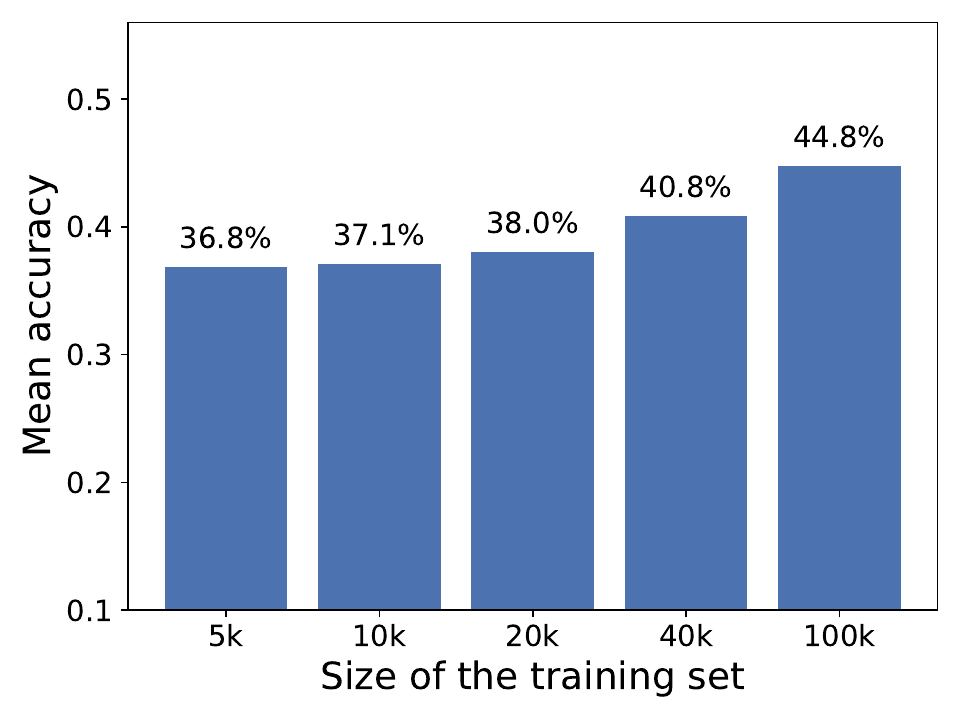}
    \end{subfigure}
    \begin{subfigure}{0.29\linewidth}
    \includegraphics[width=\linewidth]{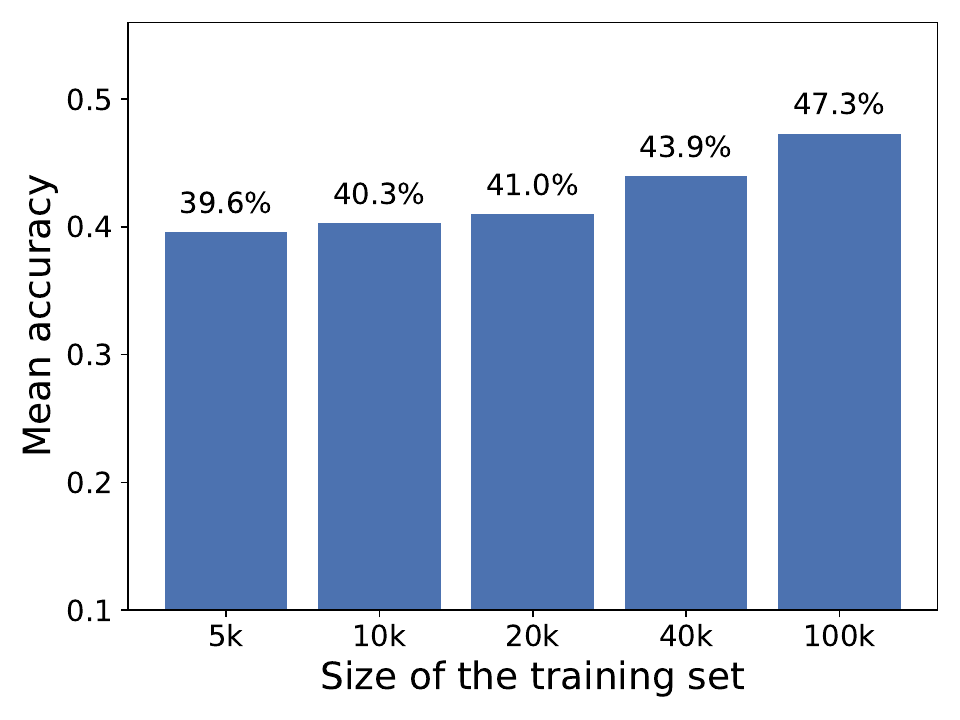}
    \end{subfigure}
    \begin{subfigure}{0.29\linewidth}
    \includegraphics[width=\linewidth]{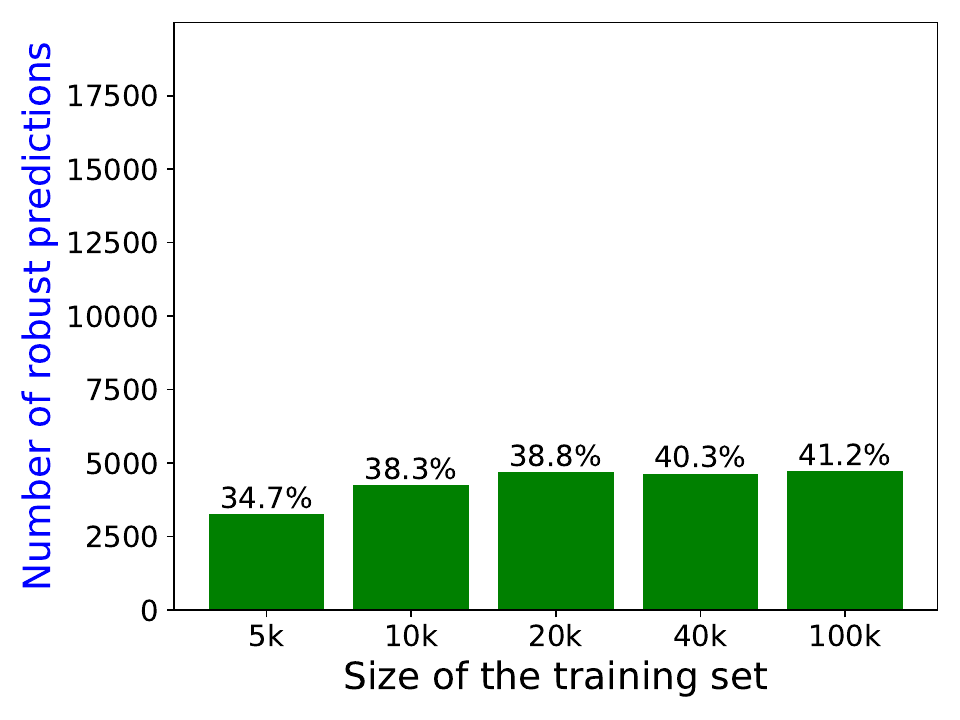}
    \caption{Online Medical Record}
    \end{subfigure}
    \begin{subfigure}{0.29\linewidth}
    \includegraphics[width=\linewidth]{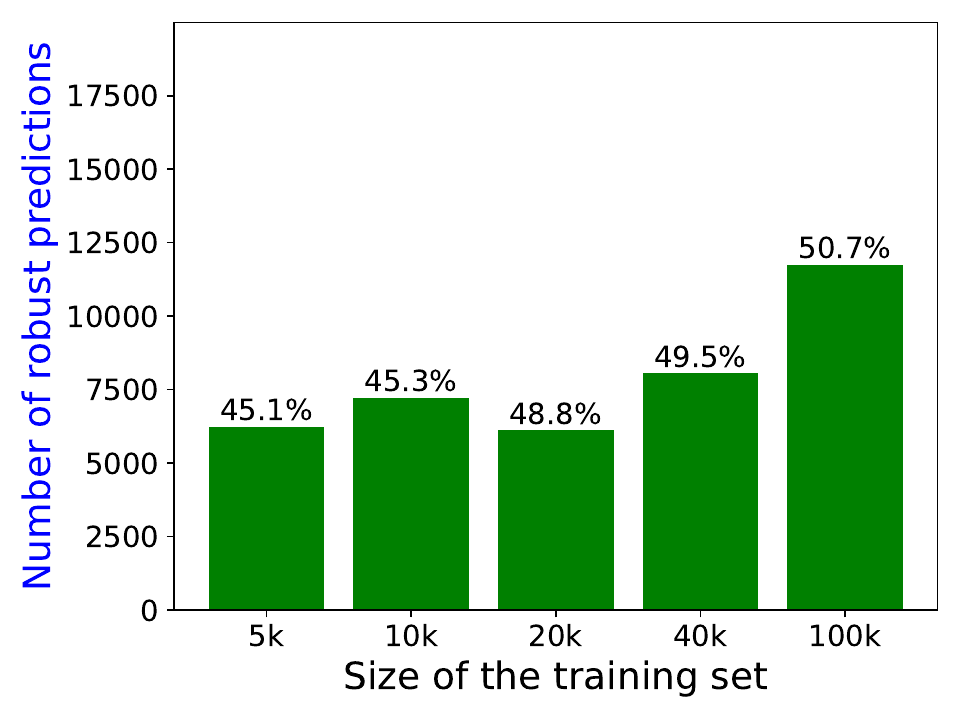}
    \caption{OMR + Lab. tests}
    \end{subfigure}
    \begin{subfigure}{0.29\linewidth}
    \includegraphics[width=\linewidth]{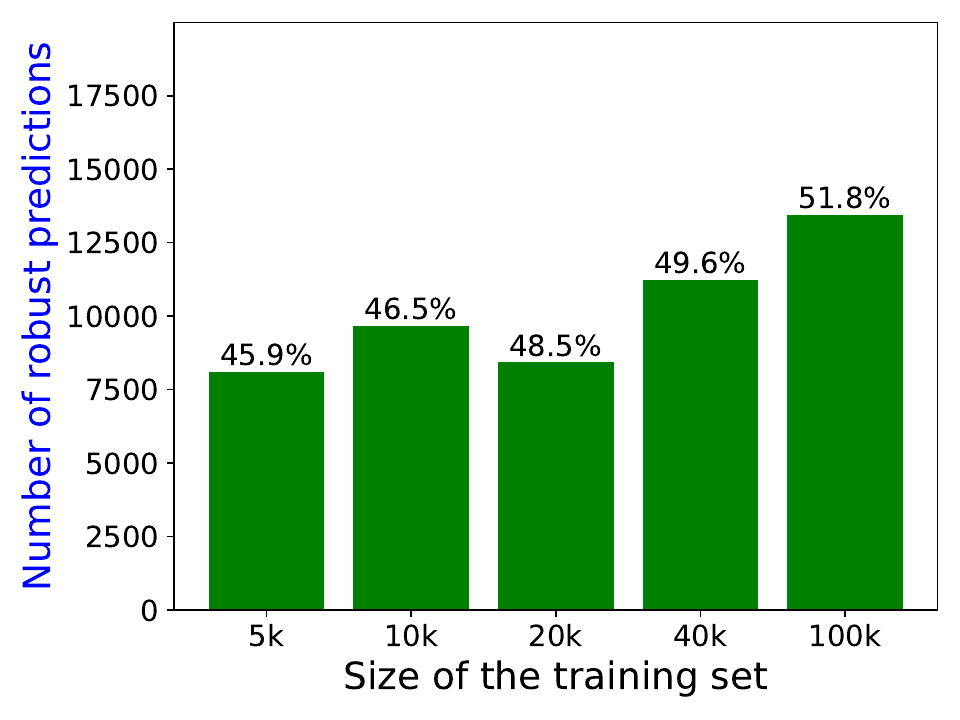}
    \caption{All + Microb. culture}
    \end{subfigure}
    \caption{Global test performance (top) and reliable predictions with performance on reliable predictions above each bar (bottom) for each model \emph{vs.} size of the training set on MIMIC-IV dataset with Deep Ensemble.}\label{fig:mimic-de}
\end{figure}

\begin{figure}
    \centering
    \begin{subfigure}{0.32\linewidth}
    \includegraphics[width=\linewidth]{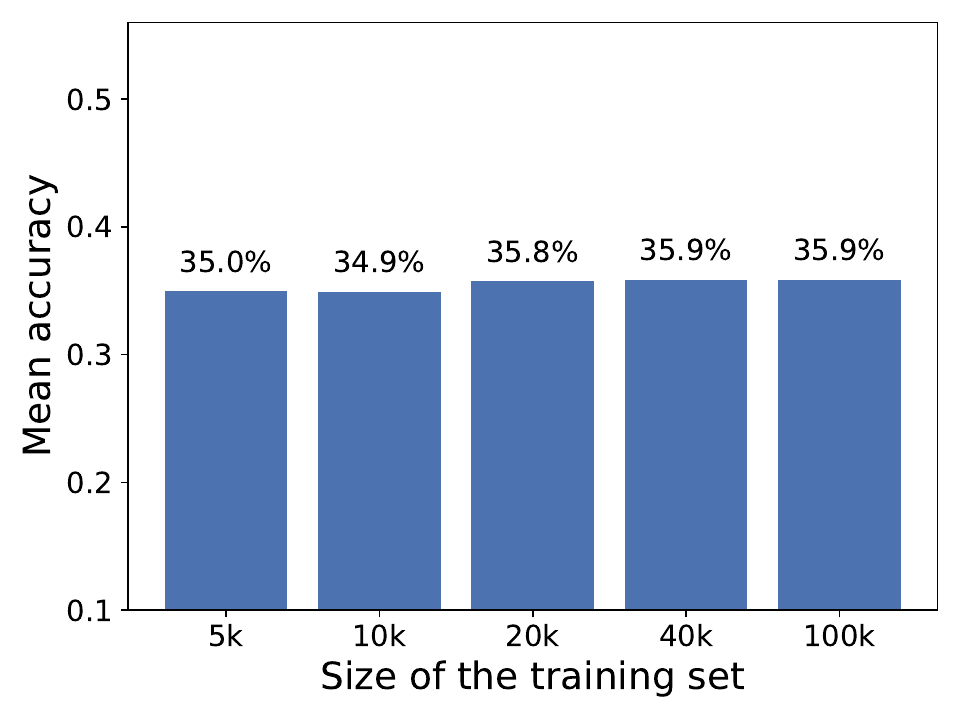}
    \end{subfigure}
    \begin{subfigure}{0.32\linewidth}
    \includegraphics[width=\linewidth]{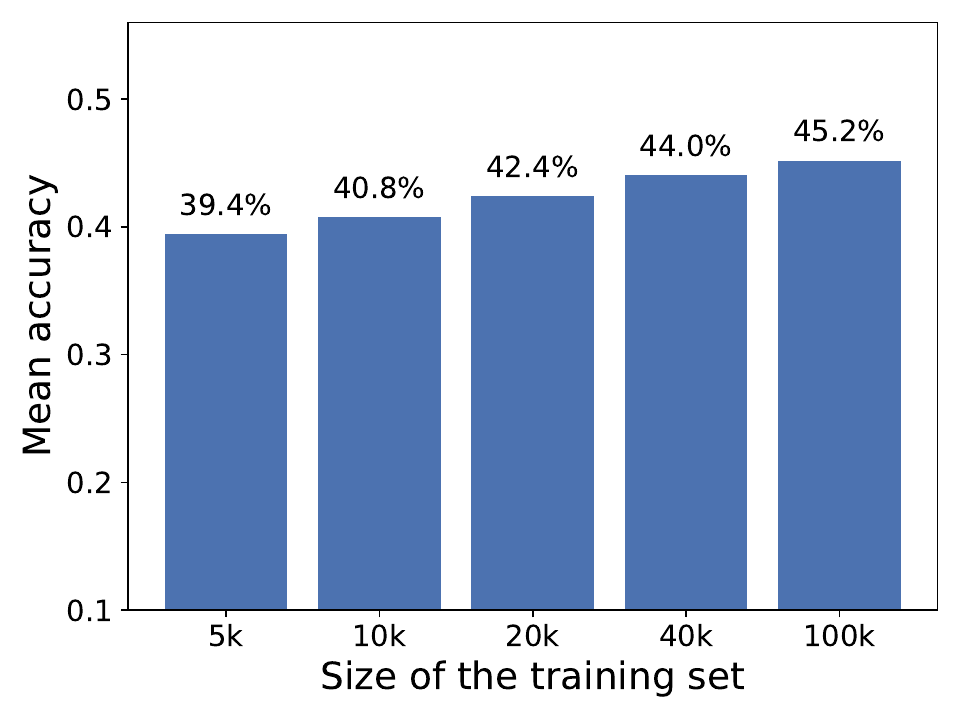}
    \end{subfigure}
    \begin{subfigure}{0.32\linewidth}
    \includegraphics[width=\linewidth]{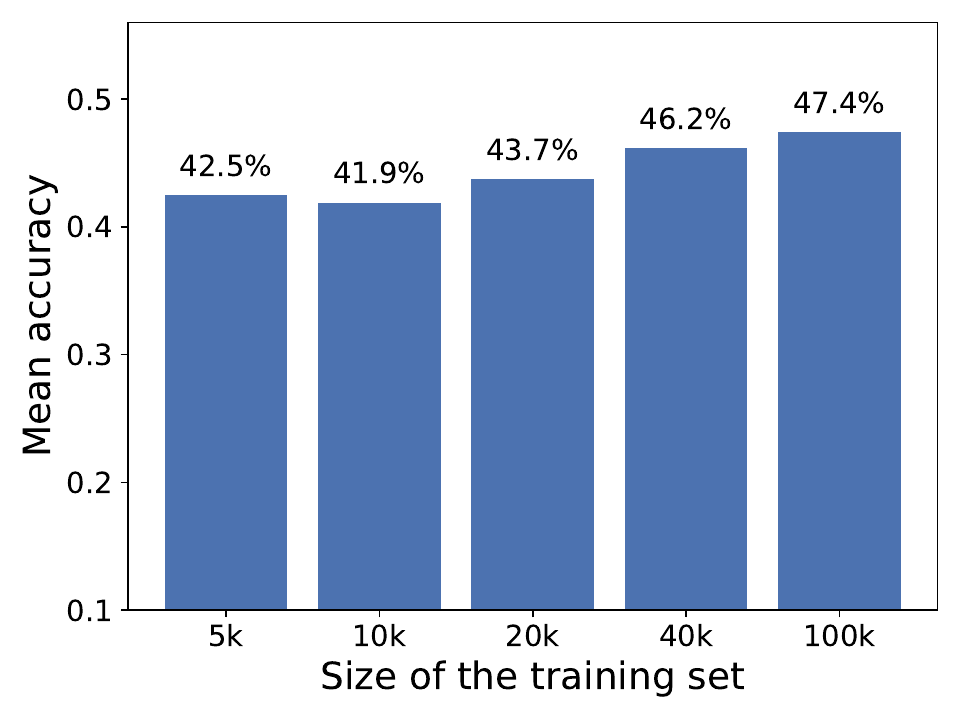}
    \end{subfigure}
    \begin{subfigure}{0.32\linewidth}
    \includegraphics[width=\linewidth]{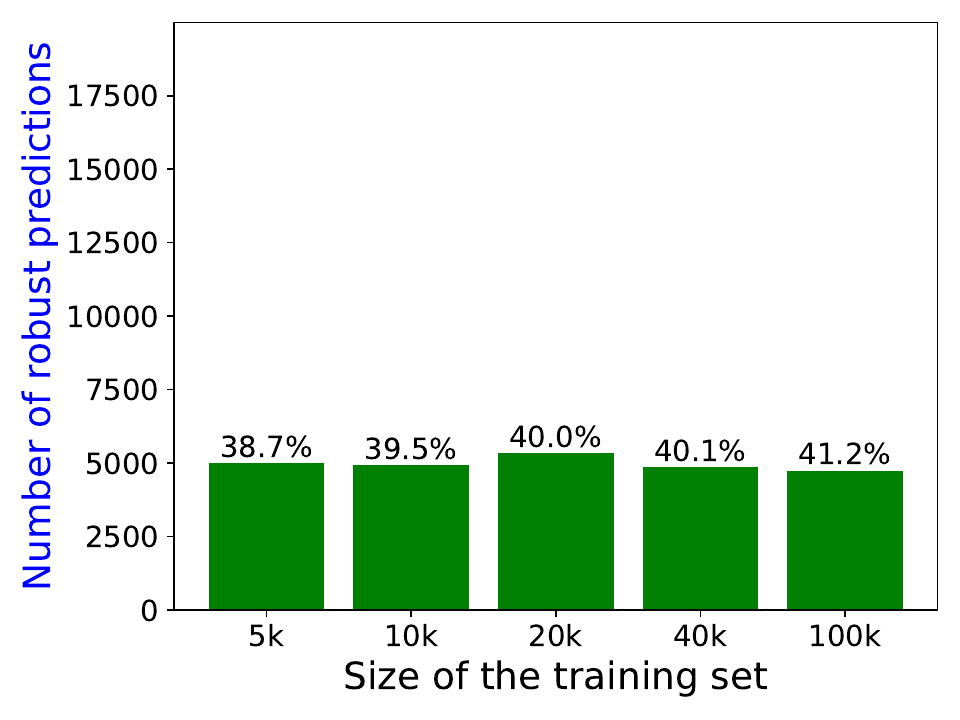}
    \caption{Online Medical Record}
    \end{subfigure}
    \begin{subfigure}{0.32\linewidth}
    \includegraphics[width=\linewidth]{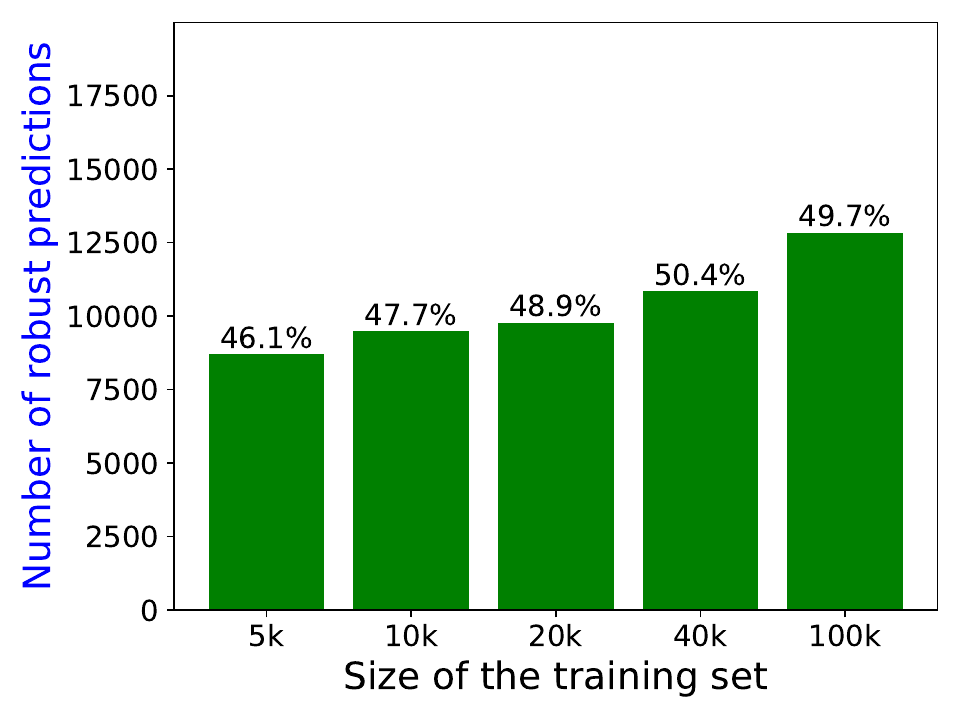}
    \caption{OMR + Lab. tests}
    \end{subfigure}
    \begin{subfigure}{0.32\linewidth}
    \includegraphics[width=\linewidth]{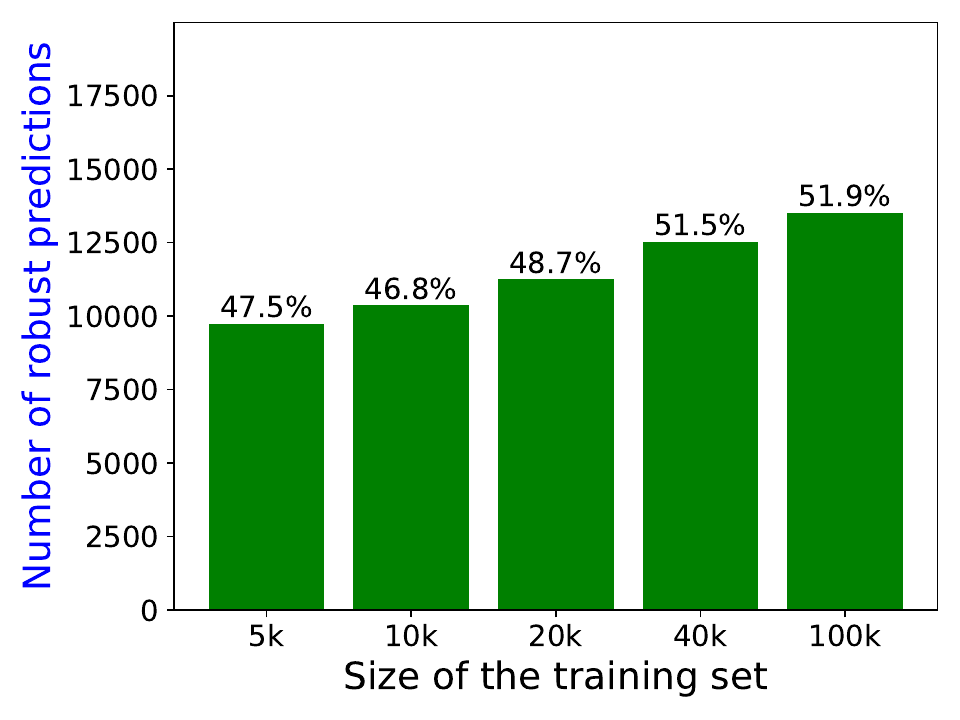}
    \caption{All + Microb. culture}
    \end{subfigure}
    \caption{Global test performance (top) and reliable predictions with performance on reliable predictions above each bar (bottom) for each model \emph{vs.} size of the training set on MIMIC-IV dataset with Bayesian Deep Learning.}\label{fig:mimic-edl}
\end{figure}

\begin{figure}
    \centering
    \begin{subfigure}{0.32\linewidth}
    \includegraphics[width=\linewidth]{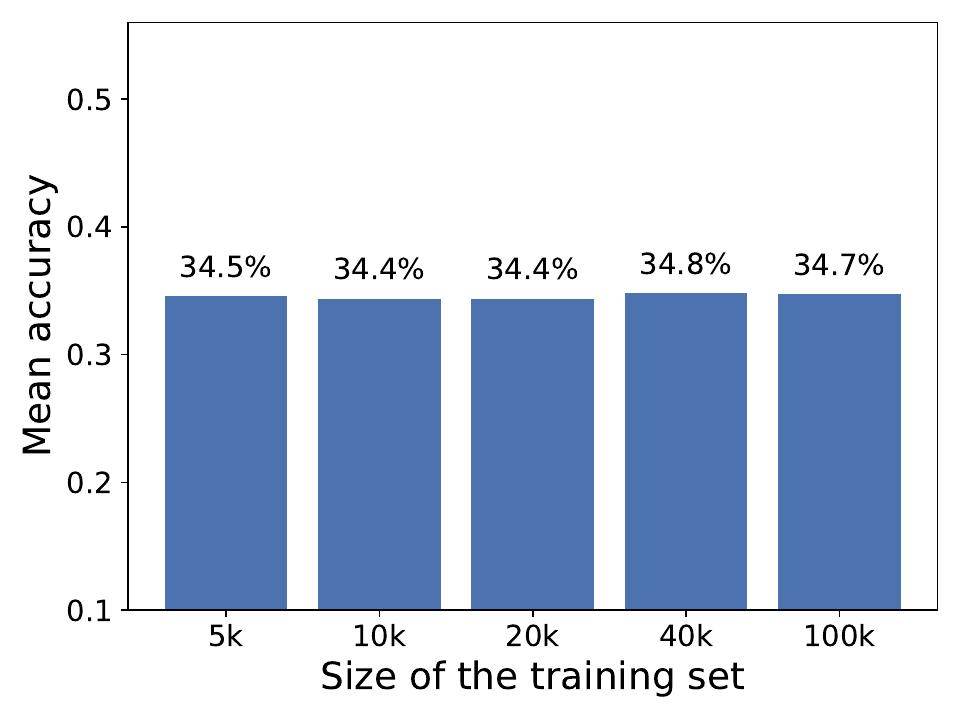}
    \end{subfigure}
    \begin{subfigure}{0.32\linewidth}
    \includegraphics[width=\linewidth]{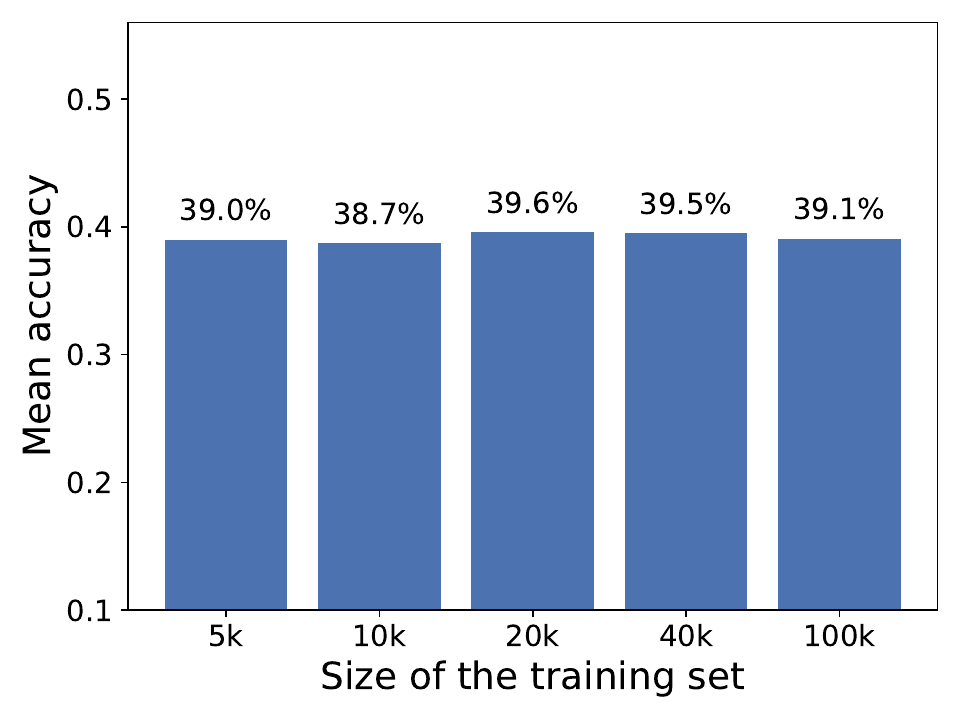}
    \end{subfigure}
    \begin{subfigure}{0.32\linewidth}
    \includegraphics[width=\linewidth]{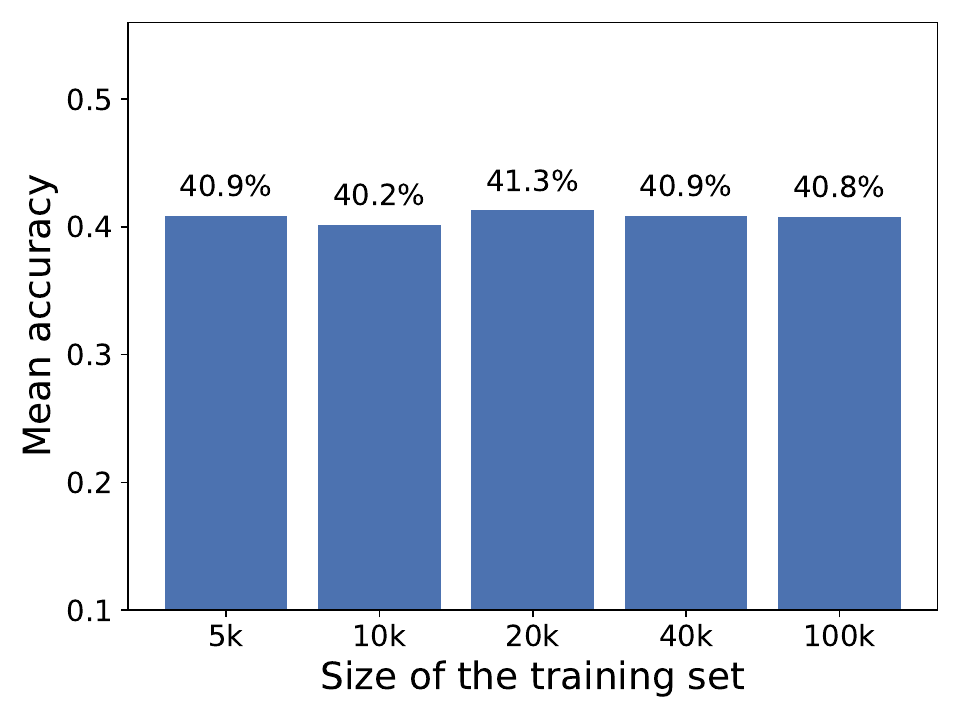}
    \end{subfigure}
    \begin{subfigure}{0.32\linewidth}
    \includegraphics[width=\linewidth]{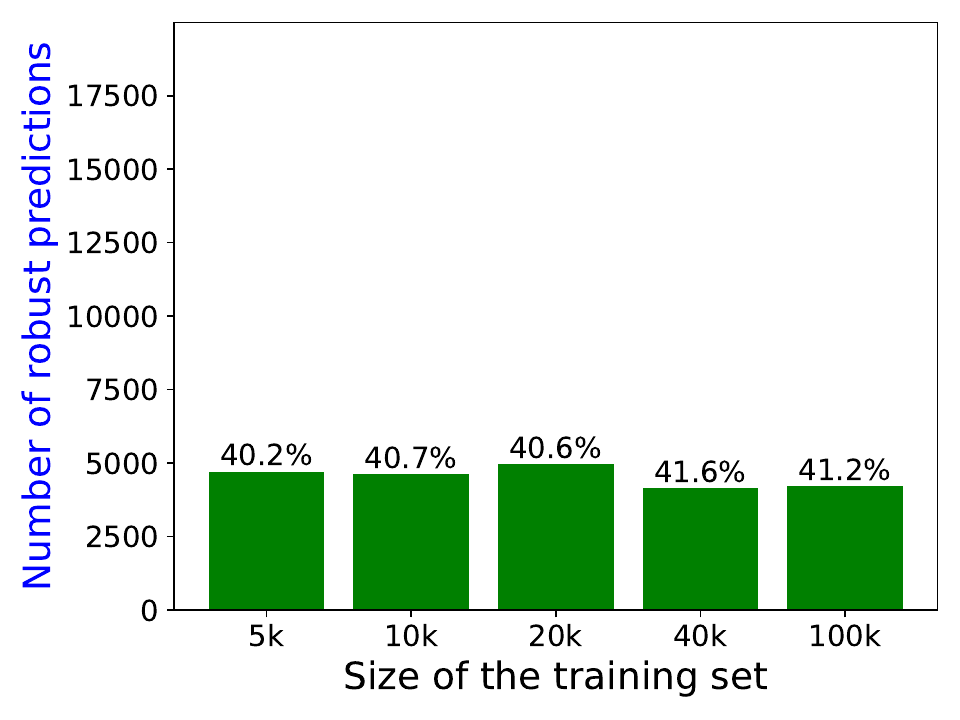}
    \caption{Online Medical Record}
    \end{subfigure}
    \begin{subfigure}{0.32\linewidth}
    \includegraphics[width=\linewidth]{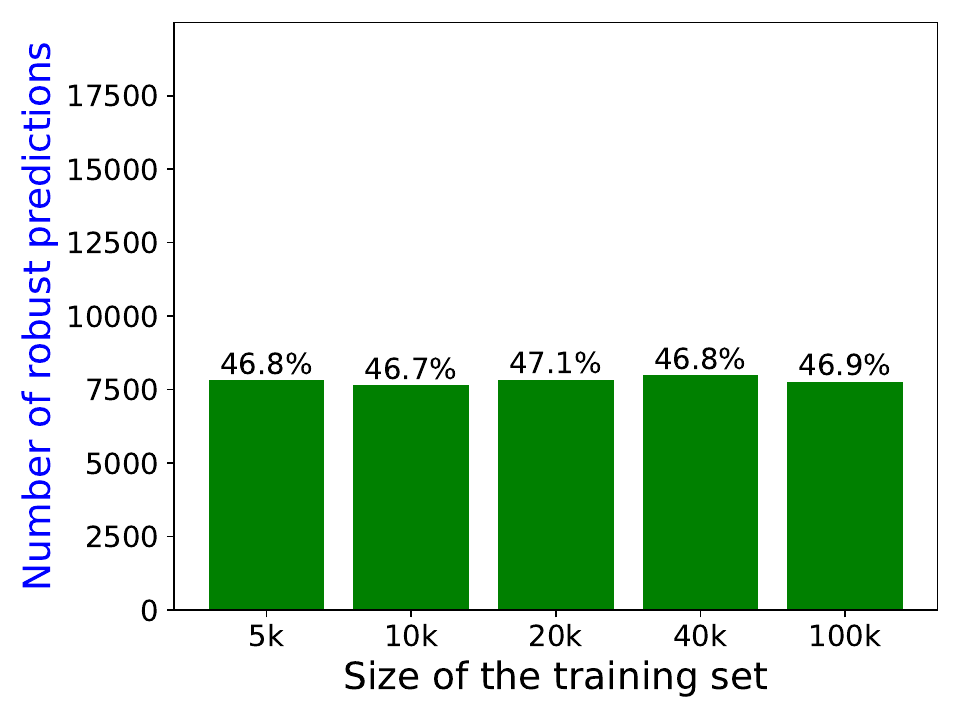}
    \caption{OMR + Lab. tests}
    \end{subfigure}
    \begin{subfigure}{0.32\linewidth}
    \includegraphics[width=\linewidth]{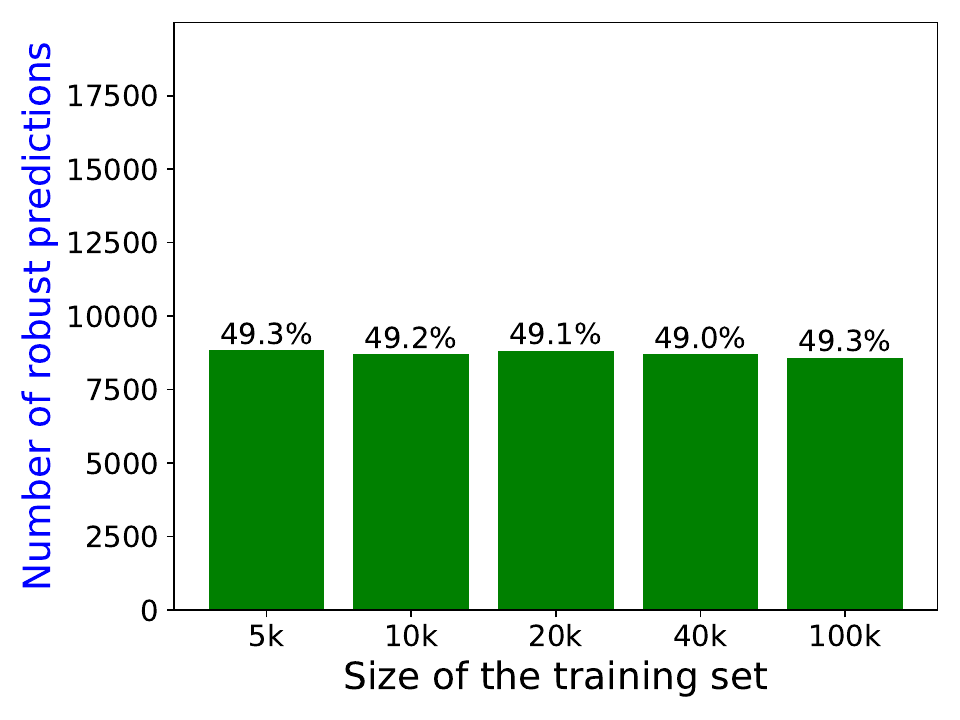}
    \caption{All + Microb. culture}
    \end{subfigure}
    \caption{Global test performance (top) and reliable predictions with performance on reliable predictions above each bar (bottom) for each model \emph{vs.} size of the training set on MIMIC-IV dataset with Random Forest.}\label{fig:mimic-rf}
\end{figure}

\begin{figure}
    \centering
    \begin{subfigure}{0.32\linewidth}
    \includegraphics[width=\linewidth]{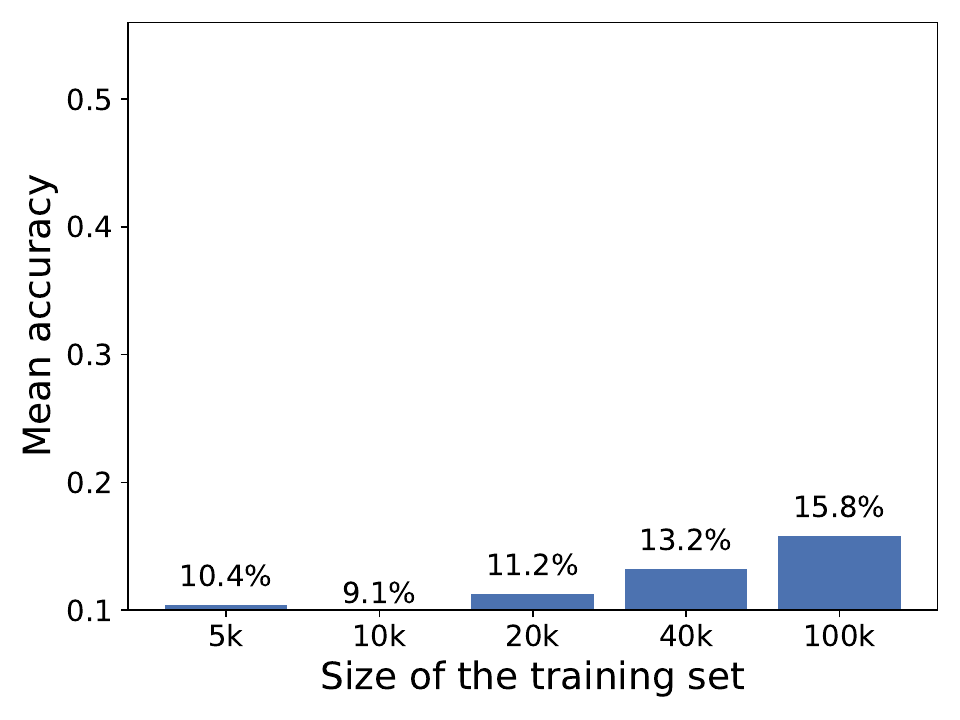}
    \end{subfigure}
    \begin{subfigure}{0.32\linewidth}
    \includegraphics[width=\linewidth]{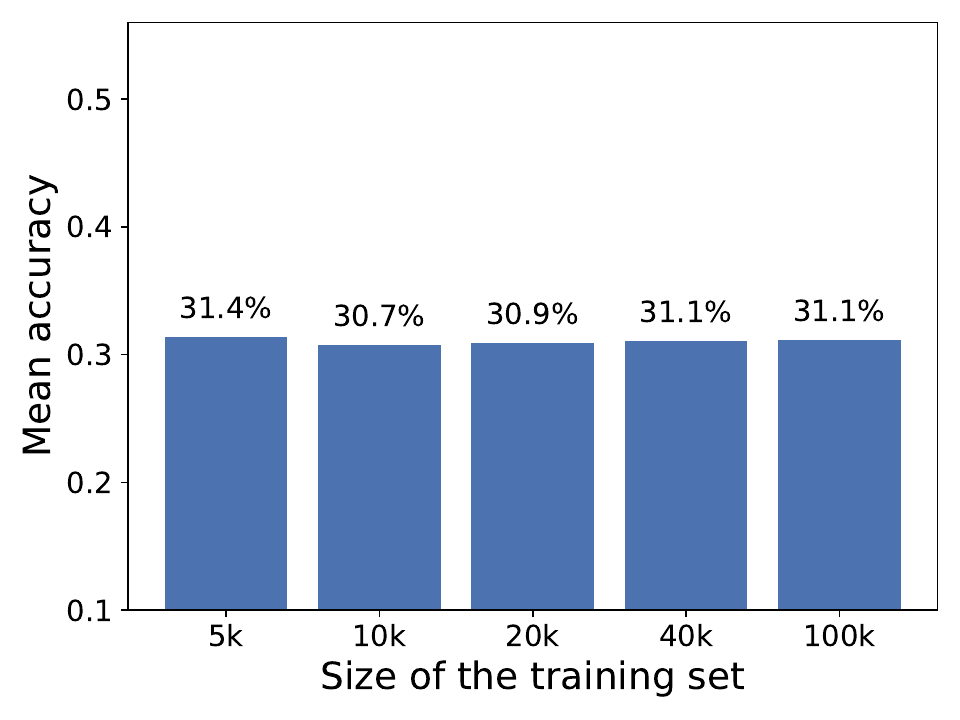}
    \end{subfigure}
    \begin{subfigure}{0.32\linewidth}
    \includegraphics[width=\linewidth]{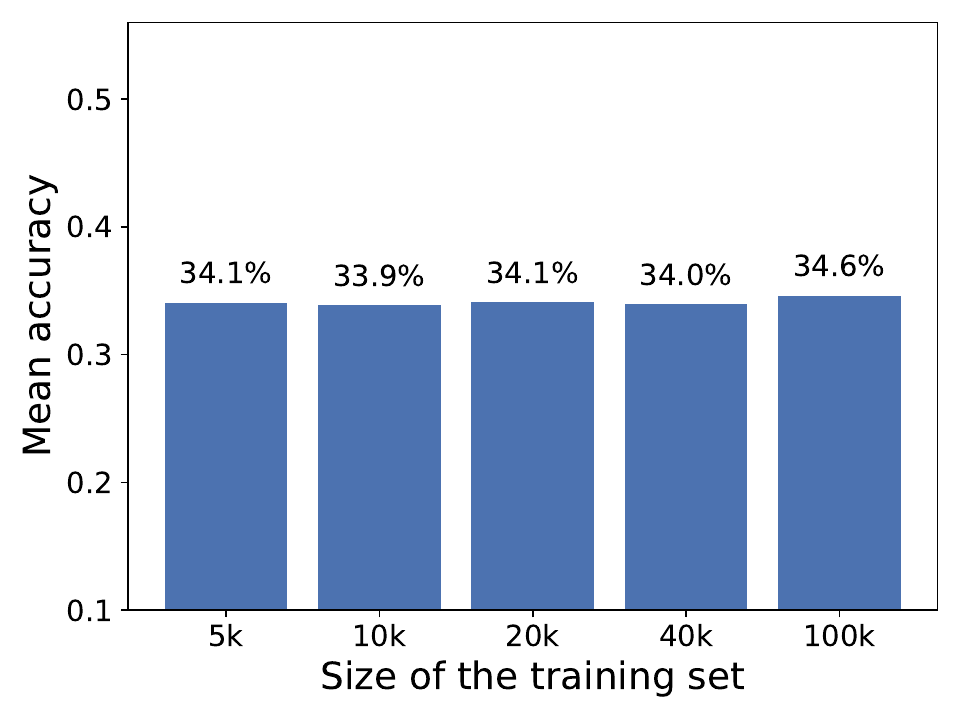}
    \end{subfigure}
    \begin{subfigure}{0.32\linewidth}
    \includegraphics[width=\linewidth]{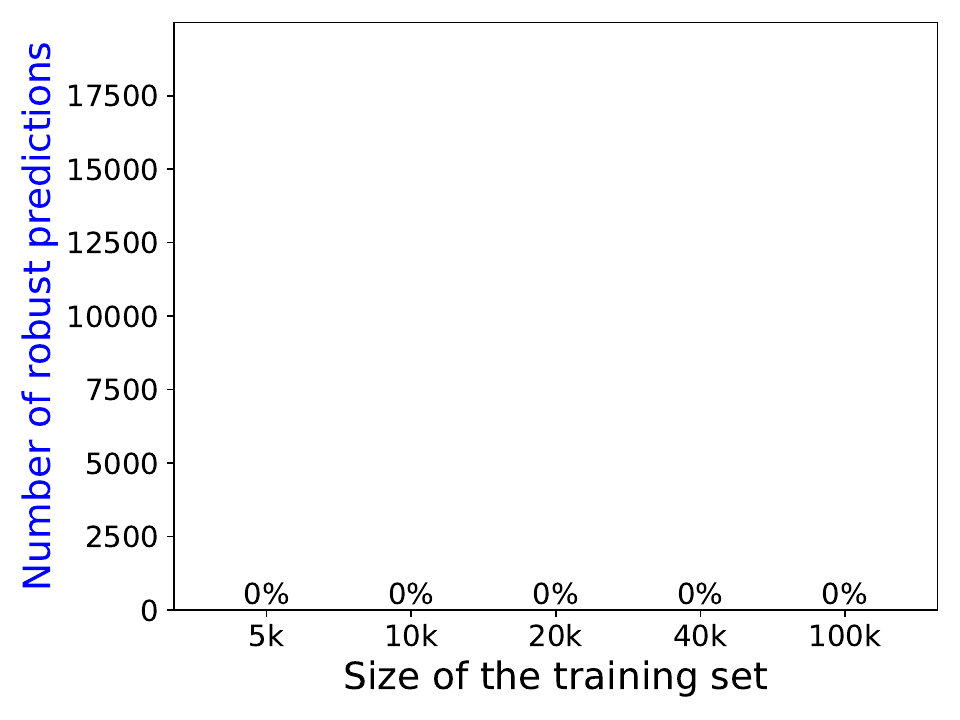}
    \caption{Online Medical Record}
    \end{subfigure}
    \begin{subfigure}{0.32\linewidth}
    \includegraphics[width=\linewidth]{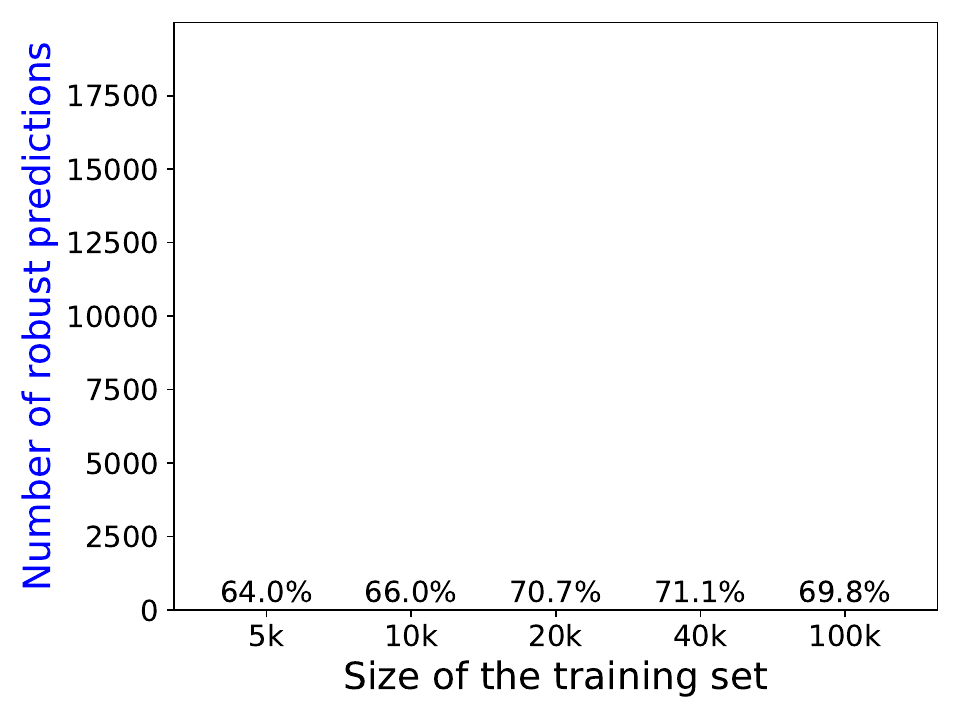}
    \caption{OMR + Lab. tests}
    \end{subfigure}
    \begin{subfigure}{0.32\linewidth}
    \includegraphics[width=\linewidth]{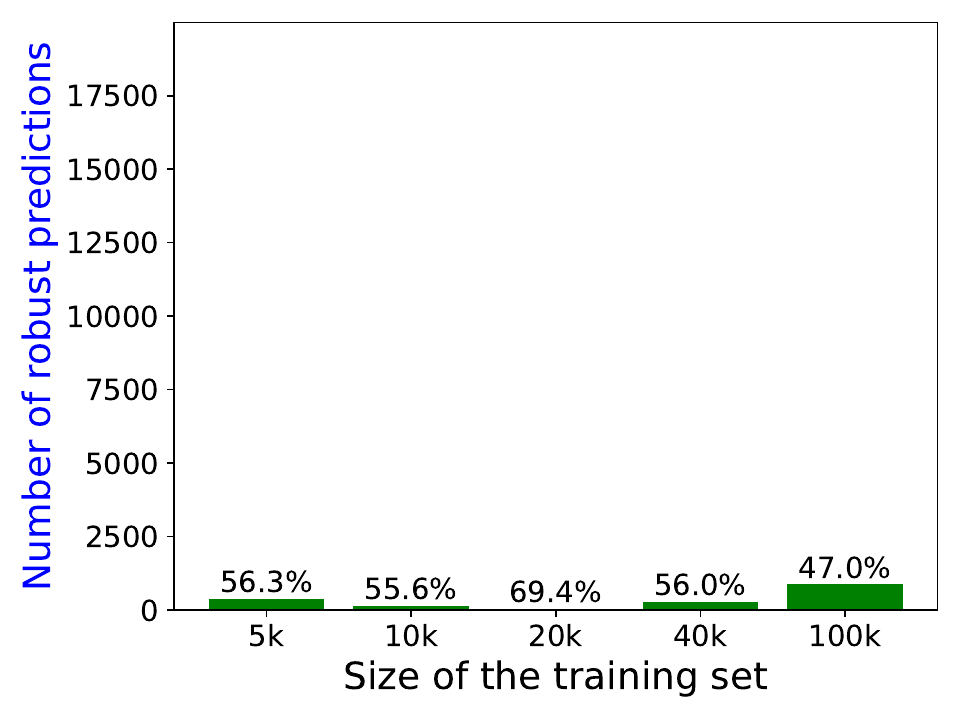}
    \caption{All + Microb. culture}
    \end{subfigure}
    \caption{Global test performance (top) and reliable predictions with performance on reliable predictions above each bar (bottom) for each model \emph{vs.} size of the training set on MIMIC-IV dataset with Centroid-based decomposition.}\label{fig:mimic-cent}
\end{figure}

\begin{figure}
    \centering
    \begin{subfigure}{0.32\linewidth}
    \includegraphics[width=\linewidth]{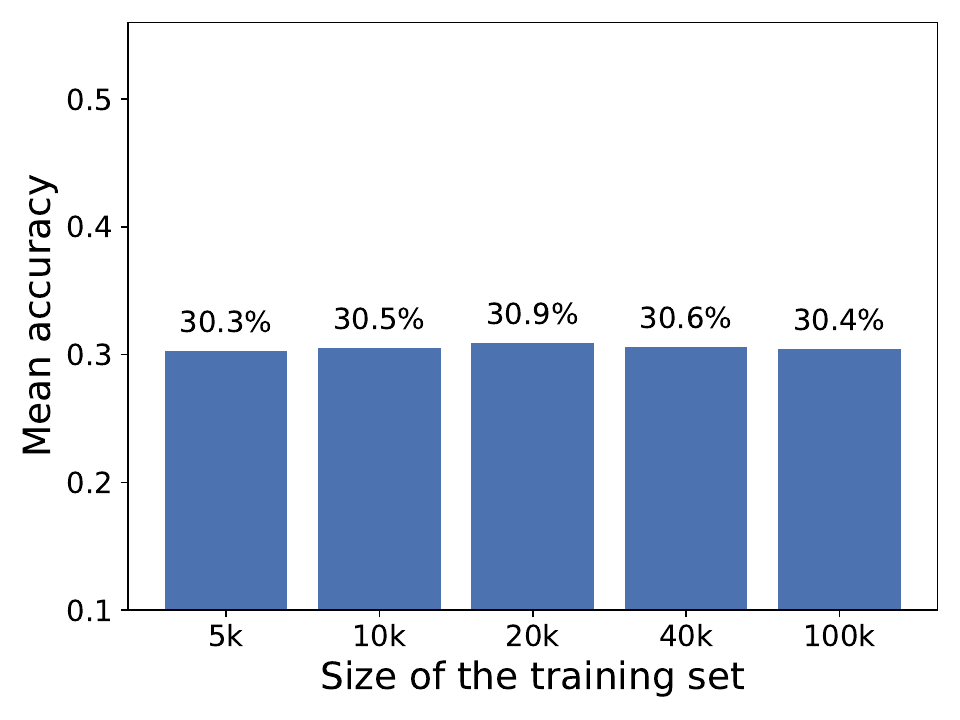}
    \end{subfigure}
    \begin{subfigure}{0.32\linewidth}
    \includegraphics[width=\linewidth]{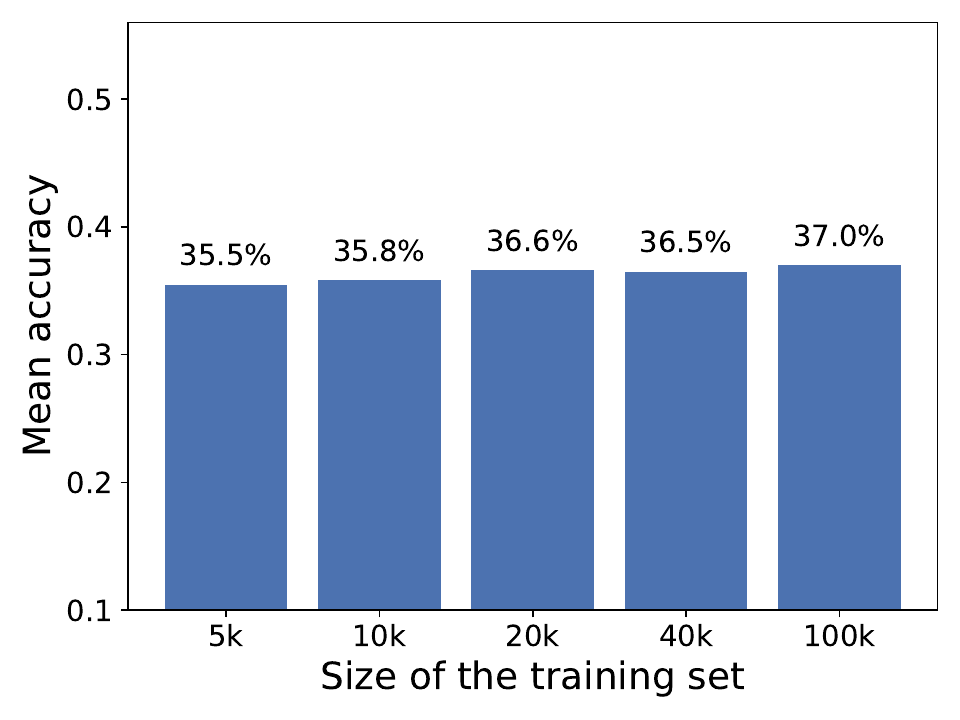}
    \end{subfigure}
    \begin{subfigure}{0.32\linewidth}
    \includegraphics[width=\linewidth]{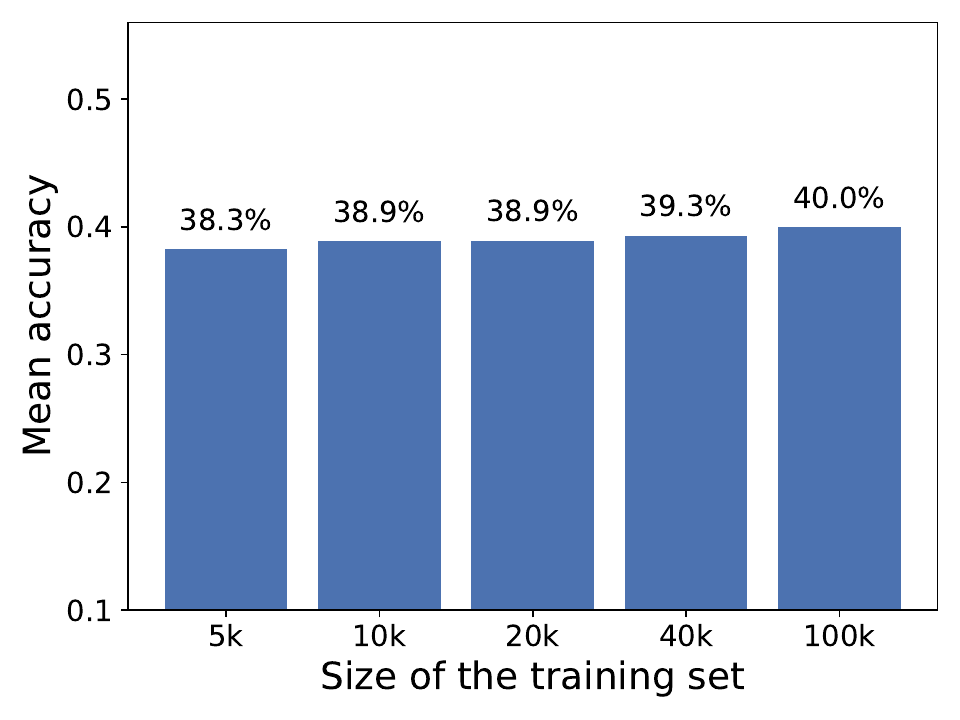}
    \end{subfigure}
    \begin{subfigure}{0.32\linewidth}
    \includegraphics[width=\linewidth]{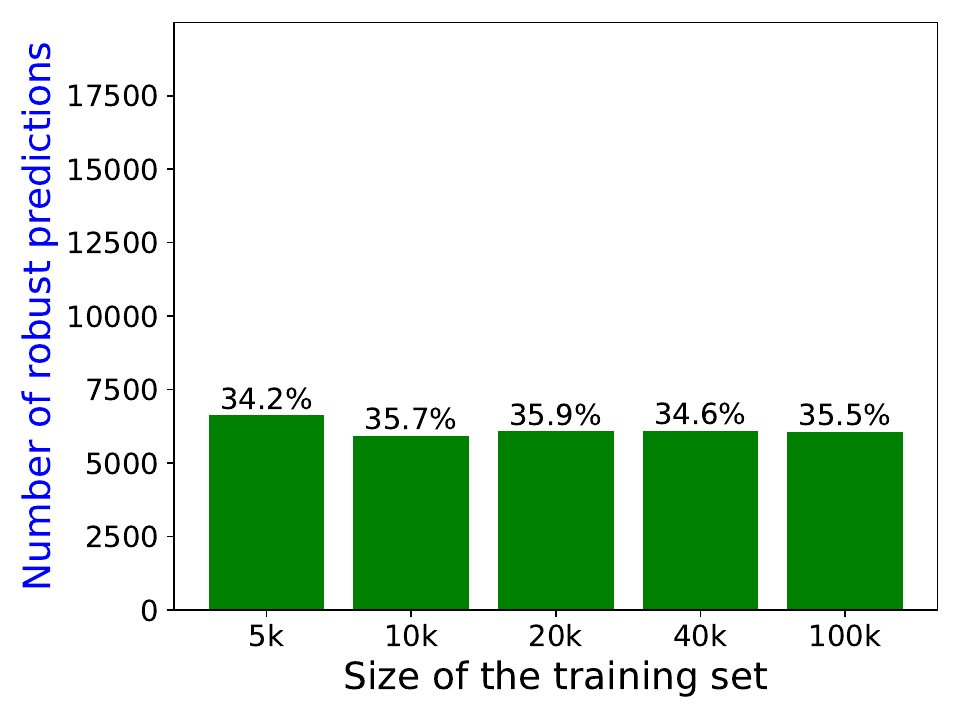}
    \caption{Online Medical Record}
    \end{subfigure}
    \begin{subfigure}{0.32\linewidth}
    \includegraphics[width=\linewidth]{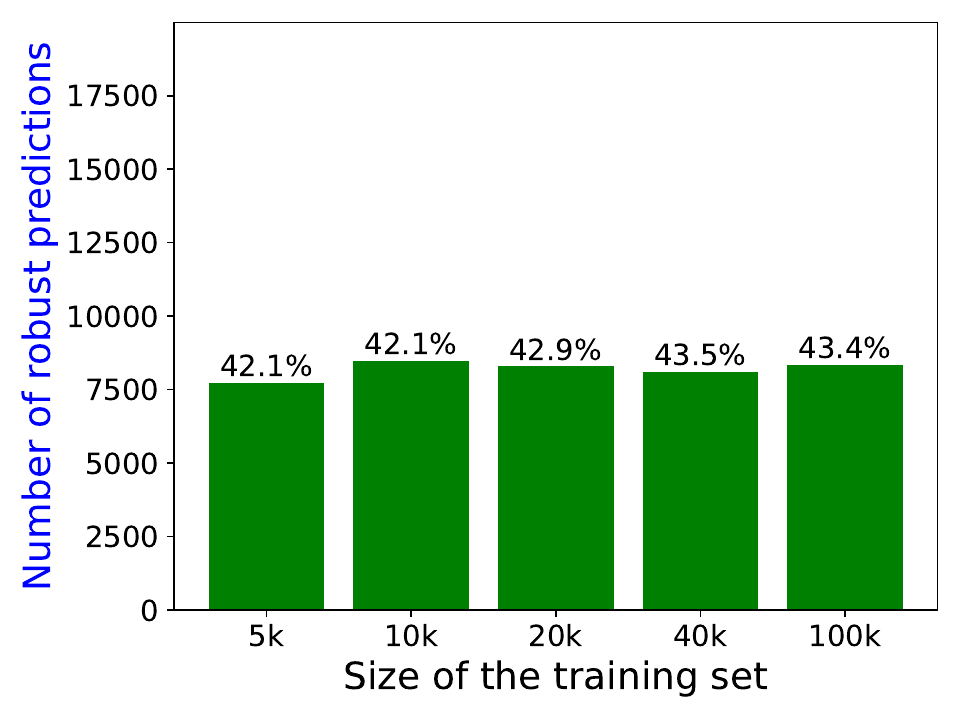}
    \caption{OMR + Lab. tests}
    \end{subfigure}
    \begin{subfigure}{0.32\linewidth}
    \includegraphics[width=\linewidth]{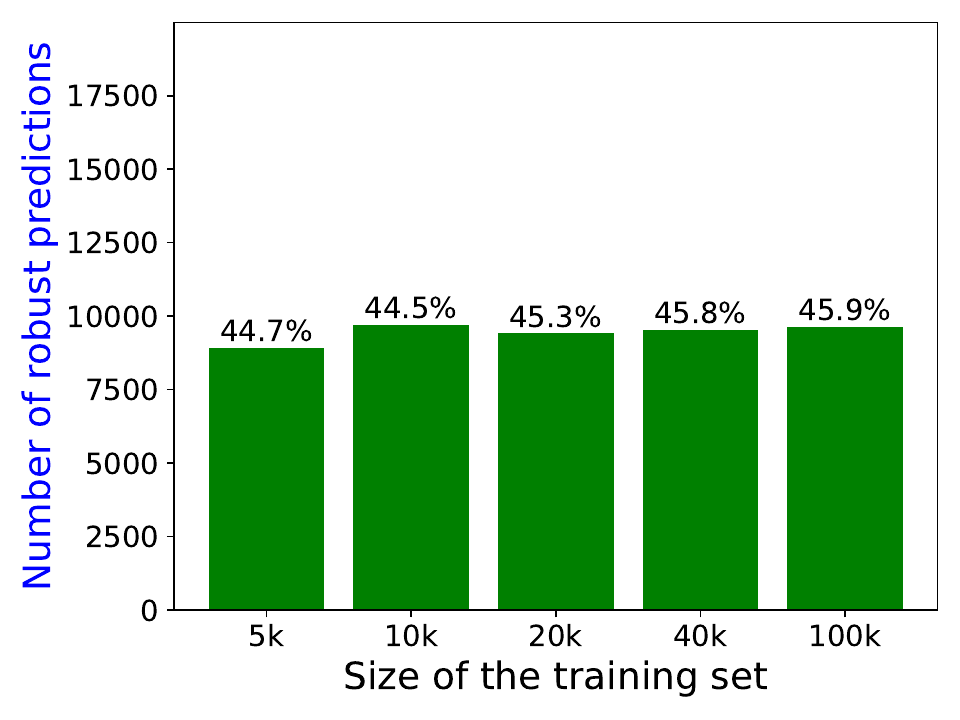}
    \caption{All + Microb. culture}
    \end{subfigure}
    \caption{Global test performance (top) and reliable predictions with performance on reliable predictions above each bar (bottom) for each model \emph{vs.} size of the training set on MIMIC-IV dataset with Neighbors-based decomposition.}\label{fig:mimic-eknn}
\end{figure}

\section{Summary of the experimental results}\label{app:summary}

In this section, we present the results of our ALMA framework. The findings reveal that the proposed method is effective across different scenarios, though the number of reliable predictions depends heavily on the underlying model and its performance. When the base model is weak, the method cannot yield satisfactory results. Additionally, the experiments demonstrate that ALMA's reliability varies with the complexity of the model, and it works best when the model has sufficient training data and complexity. Furthermore, the rejection strategy, which compares epistemic uncertainty (EU) and aleatoric uncertainty (AU), is crucial for guiding the model toward more reliable predictions.

The overall results, including the number of rejected predictions and the model performance across various methods, are summarized in two tables: Table~\ref{tab:bioscan_reject} for the BIOSCAN-5M dataset and Table~\ref{tab:mimic_reject} for the MIMIC-IV dataset. 
The presented values correspond to the rejected test instances based solely on the fully trained model, at the very end of the active learning process.
These tables provide a model comparison for the eight methods tested, offering insights into the effectiveness of ALMA across different model types and dataset characteristics. The difference between the entropy-based and variance-based decompositions does not appear to have a particularly strong impact. The centroid-based approach is the only one performing very poorly with our ALMA method, while Deep Ensemble seems to be an overall very good fit, whether using entropy-based or variance-based decomposition.

\begin{table}[p]
    \centering
    \small
    \begin{tabular}{@{}l p{0.7cm} p{1.2cm} p{0.7cm} p{1.2cm} p{0.7cm} p{1cm}@{}}
    \toprule
    Method & \multicolumn{2}{c}{Image} & \multicolumn{2}{c}{Image + Geo.} & \multicolumn{2}{c}{DNA} \\
    \midrule
    & EU & AU & EU & AU & EU & AU \\
    \midrule
    Deep-Ensemble (Ent.) & 1,143 & 2,576 & 328 & 3,345 & 0 & 0\\
    Deep-Ensemble (Var.) & 1,136 & 2,610 & 327 & 3,434 & 0 & 0\\
    Bayesian Deep (Ent.) & 305 & 3,539 & 1,093 & 2,568 & 122 & 475\\
    Bayesian Deep (Var.) & 20 & 3,794 & 21 & 3,651 & 23 & 432 \\
    Random Forest (Ent.) & 871 & 3,079 & 879 & 3,433 & 217 & 807 \\
    Random Forest (Var.) & 1,577 & 2,376 & 312 & 3,252 & 220 & 937\\
    Centroid-based & 6 & 3,964 & 1,757 & 2,242 & 3,574 & 138\\
    Neighbors-based & 41 & 3,672 & 2 & 3,133 & 2 & 591\\
    \bottomrule\\
    \end{tabular}
    \caption{Number and type of rejected predictions at the end of the learning (20k train instances) \emph{vs.} model complexity for each of the eight studied methods on BIOSCAN-5M. 4k test instances.}
    \label{tab:bioscan_reject}
\end{table}
\begin{table}[p]
    \centering
    \small
    \begin{tabular}{@{}l p{0.7cm} p{1.2cm} p{0.7cm} p{1.2cm} p{0.7cm} p{1cm}@{}}
    \toprule
    Method & \multicolumn{2}{c}{OMR} & \multicolumn{2}{c}{+ Lab tests} & \multicolumn{2}{c}{+ Microb. culture} \\
    \midrule
    & EU & AU & EU & AU & EU & AU \\
    \midrule
    Deep-Ensemble (Ent.) & 4 & 13,270 & 164 & 6,089 & 423 & 4,132 \\
    Deep-Ensemble (Var.) & 3 & 13,668 & 105 & 7,264 & 389 & 4,714 \\
    Bayesian Deep (Ent.) & 3 & 13,255 & 0 & 5,172 & 0 & 4,491 \\
    Bayesian Deep (Var.) & 0 & 12,328 & 0 & 5,312 & 0 & 3,048 \\
    Random Forest (Ent.) & 485 & 13,302 & 543 & 9,689 & 105 & 9,313 \\
    Random Forest (Var.) & 584 & 12,875 & 633 & 9,721 & 206 & 9,219 \\
    Centroid-based & 1 & 17,999 & 1 & 17,956 & 1 & 17,128 \\
    Neighbors-based & 7 & 11,918 & 31 & 9,621 & 201 & 8,157 \\
    \bottomrule\\
    \end{tabular}
    \caption{Number and type of rejected predictions at the end of the learning (100k train instances) \emph{vs.} model complexity for each of the eight studied methods on MIMIC-IV. 18k test instances.}
    \label{tab:mimic_reject}
\end{table}

\end{document}